\documentclass[10pt,graphicx,twoside]{article}
\usepackage{graphicx}
\usepackage{amsmath}

\usepackage{fancyheadings}

\usepackage{epsfig}
\usepackage{amsmath}
\usepackage{amssymb}
\usepackage{color}
\usepackage{url}

\usepackage{algorithm}
\usepackage{algorithmic}
\usepackage{mathrsfs}

\def\micron{$\mu$m}
\newcommand{\commentout}[1]{}

\begin{document}

\title{Data Mining for Faster, Interpretable Solutions to Inverse Problems: A Case Study Using Additive Manufacturing}
\author{Chandrika Kamath, Juliette Franzman, Ravi Ponmalai \\
Lawrence Livermore National Laboratory, Livermore, CA, USA}

\date{January 26, 2021}
\maketitle

\begin{abstract}
  Solving inverse problems, where we find the input values that result
  in desired values of outputs, can be challenging. The solution
  process is often computationally expensive and it can be difficult
  to interpret the solution in high-dimensional input spaces. In this
  paper, we use a problem from additive manufacturing to address these
  two issues with the intent of making it easier to solve inverse
  problems and exploit their results. First, focusing on Gaussian
  process surrogates that are used to solve inverse problems, we
  describe how a simple modification to the idea of tapering can
  substantially speed up the surrogate without losing accuracy in
  prediction.  Second, we demonstrate that Kohonen self-organizing
  maps can be used to visualize and interpret the solution to the
  inverse problem in the high-dimensional input space. For our data
  set, as not all input dimensions are equally important, we show that
  using weighted distances results in a better organized map that
  makes the relationships among the inputs obvious.
\end{abstract}

%
\section{Introduction}
\label{sec:intro}
%

Many applications require the solution of an inverse problem, where,
given the values of desired outputs, we seek the input values that
result in these outputs. In our previous work in additive
manufacturing using laser powder-bed fusion, we addressed the problem
of finding input parameters that resulted in a certain depth of molten
powder into the substrate. While we were able to solve the problem
using a Gaussian process model as a code surrogate, we encountered two
challenging issues: the first was the high computational cost of
building the surrogate and the second was our inability to understand
where the solution lay in the high-dimensional input space.

In this paper, we investigate several ideas to address these
challenges. We first introduce the application of additive
manufacturing in Section~\ref{sec:background} and describe the data
used in our study (Section~\ref{sec:data_desc}). Next, in
Section~\ref{sec:gp_speedup}, we discuss how we can potentially speed
up the Gaussian process (GP) surrogate by using approximations to the
linear solver, which is the computationally-expensive step in GP. We
consider both iterative solvers as well as a simple algorithm, which
we call the independent-block GP, that was inspired by the use of
tapering to approximate the covariance matrix.  In
Section~\ref{sec:understand_soln}, we describe the use of parallel
coordinate plots and Kohonen self-organizing maps to understand the
solution to the inverse problem in the input space. We present our
results in Section~\ref{sec:results} and conclude in
Section~\ref{sec:conclusions} with a brief summary.
Though our investigations were conducted using a data set from
additive manufacturing, the solutions we present are broadly
applicable to other inverse problems as well.

The notation used in this paper is as follows: Greek and lower case Roman
letters are scalars, bold lower case Roman letters are vectors, and
bold upper case Roman letters are matrices.

%
\section{Background and motivation}
\label{sec:background}
%

\begin{figure}[htb]
\centering
\begin{tabular}{c}
\includegraphics[trim = 0cm 2cm 0cm 2cm, clip = true,width=0.49\textwidth]{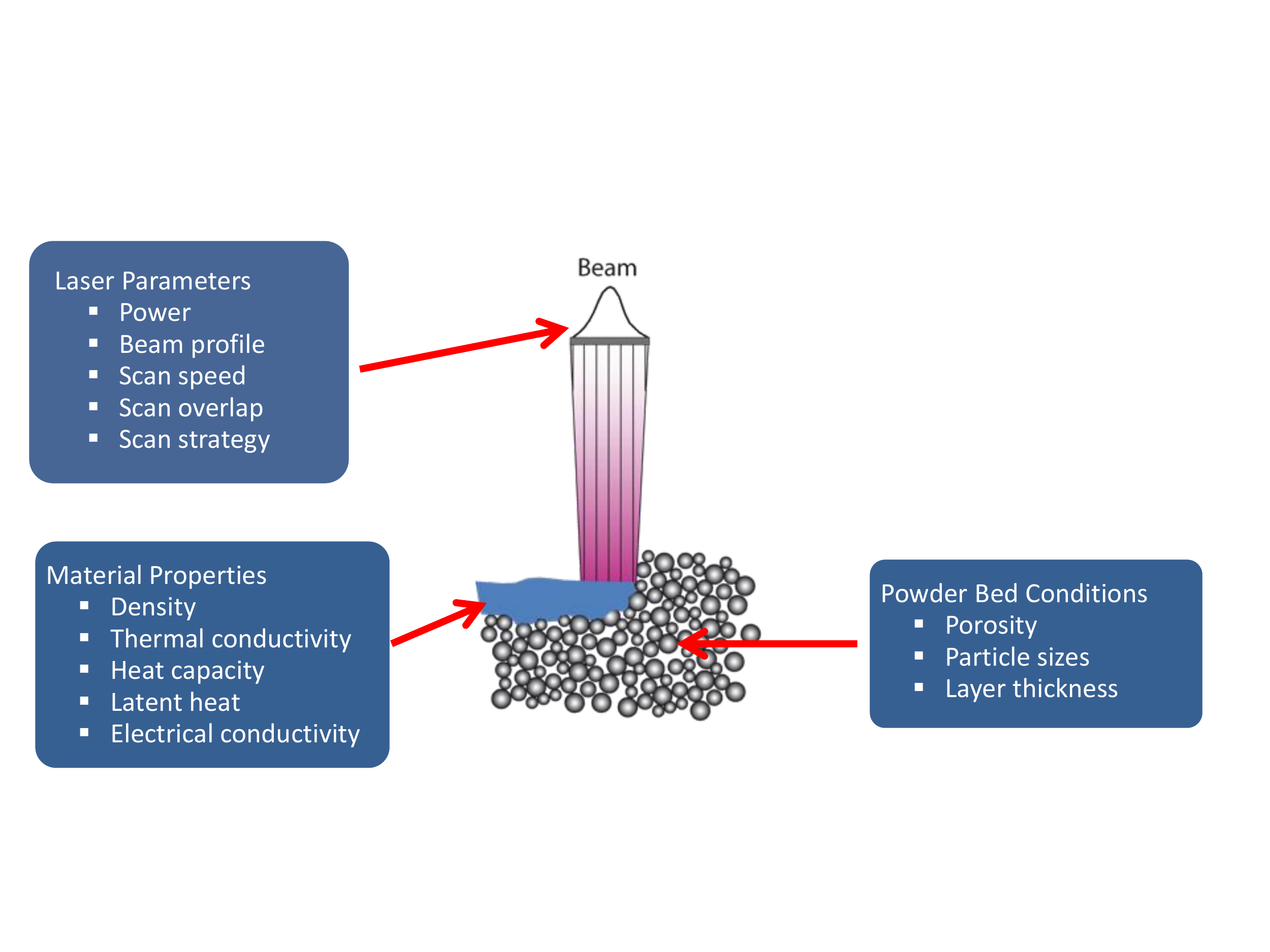} \\
\end{tabular}
\vspace{-0.2cm}
\caption{A schematic of the additive manufacturing process using laser
  powder-bed fusion illustrating the different variables that
  influence the properties of the part that is produced.}
\label{fig:am_process}
\end{figure}

Additive manufacturing (AM), or 3-D printing, is a process for
fabricating parts, layer-by-layer. In laser powder-bed fusion, each
layer in a part is created by spreading a thin layer of powder and
using a laser beam to selectively melt the powder in specific
locations so that it blends into the layers below.  A specific
challenge in AM, especially as we work with new materials and new
additive manufacturing machines, is to understand how the many
parameters that control the process affect the properties and quality
of a part. These parameters range from the settings of the laser power
and speed, to the characteristics of the powder, such as the particle
sizes and layer thickness, and the properties of the material, such as
thermal conductivity (Figure~\ref{fig:am_process}). While some
parameters are under user control, such as the power and speed of the
laser, other parameters, such as the material properties, may have
values that are known only approximately.

In previous work on additive manufacturing, we considered the problem
of determining process parameters for building dense parts. We
proposed an iterative approach~\cite{kamath14:density,kamath16:statinf}
that combined physics-based computer simulations with simple
experiments to generate data that informed the choice of parameters
likely to result in high-density parts. Intuitively, we
wanted to find parameters that fully melted each layer of powder into
the substrate below. Too little energy would result in un-melted
powder, which would lead to lower density due to the voids between the
powder particles. Too much energy would be wasteful, or worse, could
result in keyhole-mode melting~\cite{king2014:keyhole}, with very deep
melt pools and the formation of pores that would reduce the density.

In our approach, we started with a very simple, but approximate,
physics model to identify the range of viable parameters that resulted
in moderate-depth melt pools. We then identified parameters in this
range for single-track experiments where we used specific laser power
and speed values to melt a fixed thickness of powder spread on a
plate. The plate was cut perpendicular to each track to determine
how far into the substrate (the plate, in this case) the material had
melted.  Using these results, we could propose possible power-speed
pairs for use in building a dense part.

Our initial physics model, though fast enough to allow us to explore
the design space fully, was very simple and therefore, very
approximate.  As the single track experiments are expensive, we had to
carefully select the power-speed combinations so we could gain maximum
insight into the process from a limited number of tracks.  To enable
this selection, we proposed an intermediate step that used a more complex, and
accurate, physics model that was run at a small number of 
points identified by the simpler model. We could then use the resulting
data to build a code surrogate that allowed us
to predict the output at any point in the input parameter space.

These code surrogates~\cite{kamath16:statinf,kamath2018:small} are
essentially machine learning models that relate the inputs (the
parameters used in AM) to the outputs (the characteristics, such as
depth, of the melt-pool).  We can use them to solve ``inverse
problems'', where we identify input parameters that give melt-pool
depths in a given range for use in single-track experiments or in
improved control of the process through process
maps~\cite{beuth13:processmap} that identify process parameters that
result in a fixed value of the melt-pool depths.  This map would allow
us to maintain constant processing of the powder, for example, when
the laser slows down to make a turn.

\pagebreak

In our work on using code surrogates to solve inverse problems, we
encountered two data-related challenges:

\begin{itemize}

\item {\bf The computational cost of building a good surrogate:} Our
  prior work~\cite{kamath2018:small} showed that two methods ---
  locally weighted kernel regression~\cite{atkeson97:regression} and
  Gaussian process (GP)~\cite{rasmussen06:book} --- provide accurate
  predictions when the data sets are small, which is the case when
  both experiments and simulations are expensive. GPs are preferable
  as they provide both the prediction and the uncertainty on the
  prediction.  However, as we perform more experiments and
  simulations, the size of the data increases, and GP can become too
  expensive to use, even for moderate sized problems.

\item {\bf Understanding the solution of the inverse problem:} When the
  number of input parameters is greater than two, it can be challenging
  to understand where the solution to the inverse problem lies in the
  input space. The ability to explore the solution in input space is
  required to gain scientific insight and to understand which input
  parameters to vary and how to vary them to get the desired output.

\end{itemize}
We discuss ways to address these challenges in the context of additive
manufacturing but observe that they are common to the solution of
inverse problems in any application domain.

Our contributions in this paper are as follows: first, inspired by the
concept of tapering that introduces sparsity in the covariance matrix
to reduce the computational cost, we propose a simple modification,
which we call the independent-block GP, and provide guidelines for
selecting the blocks such that we improve the speed of GP without
sacrificing accuracy.  Second, we investigate ways in which we can
gain insight into the high-dimensional solutions to inverse problems
using parallel coordinate plots and Kohonen self-organizing maps. We
show that using weighted distances can lead to a better organized map,
providing greater insight.

%
\section{Description of the data}
\label{sec:data_desc}
%

We conduct our investigations using data from the Eagar-Tsai
model~\cite{eagar83:temp}, which is the simple model we used to
identify parameters for the single track experiments and sample points
for the more complex model. This model considers a Gaussian laser beam
moving on a flat plate.  The resulting temperature distribution on a
three-dimensional spatial grid representing the plate is used to
compute the melt-pool depth as a function of four input parameters:
laser power, laser speed, laser beam size, and laser absorptivity of
the powder, which indicates how much of the laser energy is absorbed
by the powder.  We generate sample points with stratified random
sampling --- using the parameter ranges and the number of levels per
input specified in Table~\ref{tab:et}, we place a sample point
randomly in each cell, resulting in a 462-sample data set, which we
refer to as ET462~\cite{kamath14:density}. This data set is used in
our work on reducing the computational cost of the GP surrogate - it
is small enough to permit investigations into different ideas, yet
large enough to help us to understand why we need to speed up the
surrogate.

\begin{table}[htb]
  \begin{center}
    \begin{tabular}{|l| l| l | l |} \hline
      Parameter & Minimum & Maximum  & Levels    \\
      \hline
      Power (W) & 50 & 400 & 7  \\
      \hline
      Speed (mm/s)& 50 & 2250 & 10  \\
      \hline
      Beam size ($\mu$m) & 50 & 68 & 3   \\
      \hline
      Absorptivity $\eta$ & 0.3 & 0.5 & 2   \\
      \hline
    \end{tabular}
  \end{center}
  \vspace{-0.2cm}
  \caption{Ranges and number of levels for the stratified random
    sampling in the four-dimensional input parameter space of the
    Eagar-Tsai model.}
  \label{tab:et}
\end{table}

The second data set we use is derived from the ET462 data set and was
generated to understand how to display the solution to the inverse
problem. We first build a GP model using the ET462 data set and use it
to predict output values at 5000 samples in the space defined by the
domain in Table~\ref{tab:et}. We use the best-candidate sampling
method~\cite{mitchell91:sampling,kamath16:statinf} to place samples
randomly as far apart from each other as possible and use the GP model
to generate the melt-pool depth at these points. This data set is
referred to as the ET5000 data set. Then, to solve the inverse problem
of finding input parameters that result in $60$\micron~ depth, we simply
identify the sample points with depth values in the range $60 \pm
\Delta$\micron, where the small interval is used to account for the
fact that few inputs give a depth of exactly $60$\micron. We
considered two values of $\Delta = 2 \ \text{and} \ 5$, consisting of 271
and 700 samples, respectively.

%
\section{Speeding up the Gaussian process surrogate}
\label{sec:gp_speedup}
%

The Gaussian process model (GP)~\cite{rasmussen06:book}, though
relatively expensive, is often a good choice as a surrogate model as
it provides accurate predictions, as well as an uncertainty on the
prediction.  We describe GP~\cite{ebden08:gp} using a $d$-dimensional
problem with a training set of $N$ samples, $\{ {\bf x}_1, \ldots,
{\bf x}_N \}$ and the associated scalar output values ${\bf y} = \{
y_1, \ldots, y_N \}$. Each ${\bf x}_i$ is of dimension $d$.  The pair
$({\bf x}_i, y_i)$ is referred to as an instance in the data set. Two
samples ${\bf x}_i$ and ${\bf x}_j$ in the training data are related
to each other through the covariance function $k(D)$ where $D$ is the
distance between ${\bf x}_i$ and ${\bf x}_j$. We use the
squared-exponential function:
\begin{equation}
   k({\bf x}_i,{\bf x}_j) = k(D) = \sigma_f^2 \exp \bigg( \frac{-D^2}{2l^2} \bigg) + \delta_{ij}\sigma_n^2  
\label{eqn:covar}
\end{equation}
where $ \sigma_f^2 $ is the maximum allowable covariance and $l$ is a
length parameter that determines the extent of influence of each
sample point and thus controls the smoothness of the
interpolation.
The term $ \sigma_n^2$ accounts for possible noise in the data, with
$\delta_{ij} = 1$ when $i = j$ and 0 otherwise.  We use a weighted
Euclidean distance $D$, with weights $w_m$ in dimension $m$, defined as
\begin{equation}
  D({\bf x}_i,{\bf x}_j) = \sqrt{\sum_{m=1}^d w_m ({x}_{im}-{x}_{jm})^2} . 
\label{eqn:dist}
\end{equation}
as we have found that it gives more accurate predictions by weighting
each input dimension based on its importance in predicting the output.
These weights, combined with the length parameter, are included in
the hyperparameters for GP and determined as described later in this
section.

Suppose we want to use the training data to predict the output at a
new sample point ${\bf x}_{\ast}$. As the data can be represented as a
sample from a multivariate Gaussian distribution, we have
\begin{equation}
  \begin{bmatrix} 
    {\bf y} \\ y_{\ast} 
  \end{bmatrix} 
  = \mathcal{N} \Bigg(  {\bf 0}, 
  \begin{bmatrix} 
    {\bf C} & {\bf c}_{\ast}^T \\ 
    {\bf c}_{\ast} & c_{\ast \ast}
  \end{bmatrix}  \Bigg) \ ,  \notag
\end{equation}
where $\bf y$ is the output variable corresponding to the $N$ training
instances, $y_{\ast}$ is the prediction of the output at point ${\bf x}_{\ast}$,
and the sub-matrices are defined as follows:
\begin{equation}
  {\bf C} = 
  \begin{bmatrix}
    k({\bf x}_1, {\bf x}_1) & k({\bf x}_1, {\bf x}_2) & \ldots & k({\bf x}_1, {\bf x}_N) \\
    k({\bf x}_2, {\bf x}_1) & k({\bf x}_2, {\bf x}_2) & \ldots & k({\bf x}_2, {\bf x}_N) \\
    \vdots & \vdots & \vdots & \vdots \\
    k({\bf x}_N, {\bf x}_1) & k({\bf x}_N, {\bf x}_2) & \ldots & k({\bf x}_N, {\bf x}_N) \\
  \end{bmatrix}\ , 
\label{eqn:covar_matrix}
\end{equation}
\begin{equation}
  {\bf c}_{\ast}^T = 
  \begin{bmatrix}
    k({\bf x}_{\ast}, {\bf x}_1) & k({\bf x}_{\ast}, {\bf x}_2) & \ldots & k({\bf x}_{\ast}, {\bf x}_N) \\
  \end{bmatrix} \ , \notag
\end{equation}
and
\begin{equation}
  c_{\ast \ast} = k({\bf x}_{\ast}, {\bf x}_{\ast}).  \notag
\end{equation}

The probability of $y_{\ast}$, that is, the output at the new sample point,
is then given by
\begin{equation}
  \overline{y}_{\ast} = {\bf c}_{\ast}^T {\bf C}^{-1} {\bf y}  
\label{eqn:prediction}
\end{equation}
and the uncertainty in the estimate is given by the variance
\begin{equation}
  \text{var}(y_{\ast}) = c_{\ast \ast} - {\bf c}_{\ast}^T {\bf C}^{-1} {\bf c}_{\ast} . 
\label{eqn:variance}
\end{equation}
The most expensive step in these equations is solving a system of the
form ${\bf C} {\bf u} = {\bf v}$. If we are interested only in
predictions at new sample points, this system has to be solved only
once for all new sample points with ${\bf v} = {\bf y}$.  But if we
are also interested in the variance in the prediction, then the system
has to be solved for each new sample point as, from
Equation~\ref{eqn:variance}, the right-hand side ${\bf v}$ is ${\bf
  c}_{\ast}$, which depends on the new sample point.

Equations~\ref{eqn:covar} and~\ref{eqn:dist}, indicate that the GP
model is described by $d+2$ hyperparameters: $ \sigma_f$, $\sigma_n $,
and the $d$ weighted length scales $l/w_i$ that are used in the distance
calculation.  To obtain the optimal values for these $d+2$
hyperparameters for our training data set, we use a random
search~\cite{bergstra12:hyperparam} through the $(d+2)$-dimensional
space of parameters. To evaluate the quality of each set of
hyperparameters, we use the mean absolute error (MAE) for a leave-one-out
evaluation, that is, for each sample, we build the GP model using the
remaining $(N-1)$ samples, calculate the prediction on the sample held
out, and use the mean of the absolute error in prediction across all
samples as the quality metric. If $R$ sets of hyperparameters are used
to find the optimum, this results in building $R * N$ GP models. This
can be a large number when $N$, the number of instances in the data
set, is large. Even for moderate $N$, when the dimension $d$ of the
problem is large, exploring the high-dimensional hyperparameter space
to find the optimal set would require a large value for $R$.

Therefore, to reduce the cost of hyperparameter optimization in
building a surrogate model, we can use one or more of the following ideas:

\begin{itemize}

\item reduce the number $R$ of sets of hyperparameters considered in
  the optimization, which could result in identifying a less than
  optimal set, or

\item partially evaluate each set of hyperparameters if possible, and
  fully evaluate only those found to be
  promising~\cite{li2018:hyperband}, or

\item use a metric that considers only some of the instances in the
  data set~\cite{rasmussen06:book}, which could result in a less than
  optimal set of hyperparameters depending on which training instances
  were used in the evaluations, or

\item use an approximate instead of an exact method to solve the linear
  system in Equations~\ref{eqn:prediction} and~\ref{eqn:variance}.

\end{itemize}

In this paper, we focus on the last of these ideas, where we reduce
the cost of hyperparameter optimization, not by a less-than-thorough
search of the parameter space, but by reducing the cost of evaluating
each parameter set through an approximate solution of the linear
system. We consider two options: first, the use of a direct solver
with an approximate covariance matrix obtained using a simple idea
derived from the concept of tapering, and second, an iterative solver,
where we control the approximation using the number of iterations.

%
\subsection{Using approximations in a direct solver}
\label{sec:tapering}
%

\begin{figure}[htb]
\centering
\begin{tabular}{c}
\includegraphics[width=0.45\textwidth]{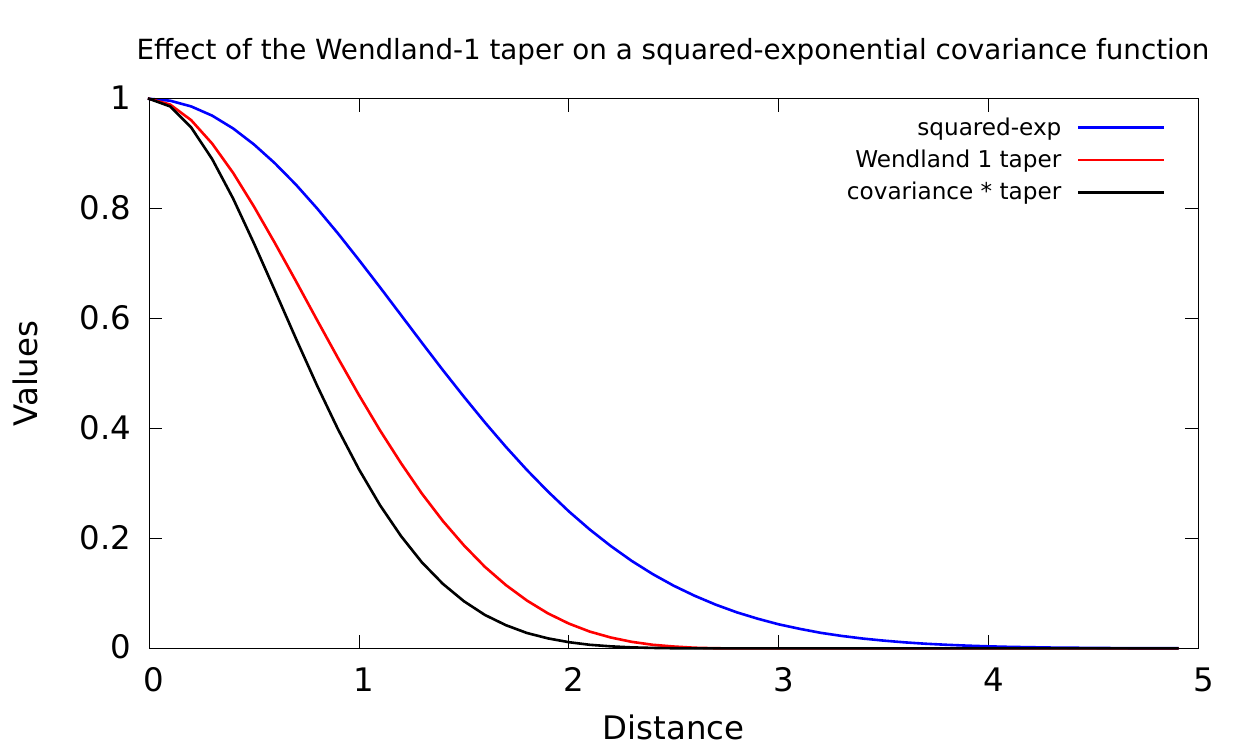} \\
\end{tabular}
\vspace{-0.2cm}
\caption{The effect of a Wendland-1 taper, with $\theta = 3.0$ applied
  to a squared-exponential covariance function
  (Equation~\ref{eqn:covar}) with $\sigma_f = 1.0$, $\sigma_n = 0.0$
  and $l = 1.2$. Note that tapering reduces all values of the
  covariance function.}
\label{fig:taper}
\end{figure}

To solve the system ${\bf C} {\bf u} = {\bf v}$, with the covariance
matrix ${\bf C}$ defined in Equation~\ref{eqn:covar_matrix}, using a direct
solver, we first obtain the Cholesky factorization of ${\bf C} = {\bf
  L} {\bf L}^T$ and then solve for ${\bf u}$ using two triangular
solves ${\bf u} = {\bf L}^{-T} ( {\bf L}^{-1} {\bf v} ) $.  One way to
approximate the covariance matrix is by tapering, where we introduce
sparsity into the matrix and exploit the sparsity to reduce the
computational cost. While we could explicitly
set values in the covariance matrix less than a certain threshold, $\theta$,
to zero, the resulting matrix could become
non-positive-definite~\cite{furrer2006:taper}. Instead, tapering is
applied by multiplying the covariance function by another function
$k_{\theta}$
\begin{equation}
  k_{taper}(D) = k(D) * k_{\theta}(D)
  \label{eqn:covar_taper}
\end{equation}
where $k_{\theta}(D)$ is
defined to be 0 if $D \ge \theta$ and a value between 0 and 1 if $D <
\theta$. This introduces sparsity while retaining the
positive-definiteness of the matrix.  Several taper functions have
been suggested, for example, the Wendland-1
taper~\cite{furrer2006:taper}:
\begin{equation}
  k_{\theta}(D) = \bigg( 1-\frac{D}{\theta} \bigg)_{+}^{4} \bigg( 1-4 \frac{D}{\theta} \bigg)
  \label{eqn:wend1_taper}
\end{equation}
where $(x)_{+} = max(0,x)$. Figure~\ref{fig:taper} shows the effect of
the Wendland-1 taper on a squared-exponential covariance function
based on the distance from a point.

In our previous work~\cite{franzman2019:taper}, we found that tapering
benefited only problems with a specific characteristic.  When we
introduce sparsity into the covariance matrix, the amount of sparsity
must be such that it can be exploited to reduce the computational
cost. At the same time, the values being dropped must be very small so
that the solution with the tapered covariance matrix is close to that
of the original covariance matrix.  Both these conditions are
satisfied when the function being approximated varies relatively
rapidly so that the influence of each sample point is small, i.e., the
optimal length scale $l$ in the covariance function is small.

This finding led to two observations. First, tapering would be
inappropriate if the function we want to model using GP varies
smoothly, not rapidly. This is the case with our additive
manufacturing data set, and as the true range of the covariance
function is much larger than the tapering range, we cannot introduce
sparsity without introducing a large error in the tapered covariance
matrix~\cite{bolin2013:taper}.  Second, in the search for optimal
hyperparameters we cannot artificially restrict the length scale to be
small, as we do not know apriori the true length scale of our data
set.  So, even if our data set could benefit from tapering, there will
be some parameter sets in the hyperparameter optimization that will
not benefit from tapering, and the computational costs may not reduce
to the extent desired to make the use of GP tractable.

In the course of understanding the conditions under which we could use
tapering, we came across a very simple idea in the literature on
covariance tapers that had been found to work better than traditional
tapers. In a paper discussing the theory behind the choice of a
taper~\cite{stein13:taper}, Stein found that ``for purposes of
parameter estimation, covariance tapering often does not work as well
as the simple alternative of breaking the observations into blocks and
ignoring dependence across blocks.''. As he observes, this idea can be
considered as a specific form of a taper, with
\begin{equation}
  k_{\theta} ({\bf x}_i, {\bf x}_j) =
  \begin{cases}
    1, & \text{if}\ {\bf x}_i\ \text{and}\ {\bf x}_j\ \text{are in same block} \\
    0, & \text{otherwise.}
  \end{cases} \notag
\end{equation}
A later paper by Bolin and Wallin on adaptive tapers for data sampled
at irregular locations~\cite{bolin2015:taper} also observed that while
adaptive tapers outperform regular tapers, block tapering was
often a better method for parameter estimation.
Unfortunately, neither paper discussed how the data were to be
partitioned into blocks; specifically, the dimensions along which we
should block the data or the block sizes to use.  To partition the
observations into blocks such that nearby observations are in the same
block, we have two options:

\begin{itemize}
\item {\bf Block tapering:} This is the approach
  in~\cite{stein13:taper,bolin2015:taper} where the covariance matrix
  is created as a block diagonal matrix, with each diagonal block
  corresponding to a block of the data.  Though the Cholesky
  decomposition of each diagonal block is calculated separately, the
  mean absolute error across all blocks is considered in evaluating a
  set of hyperparameter values.  As a result, a single set of optimal
  hyperparameters is identified for the data set, though this set may
  be suboptimal for each block considered independently.

\item {\bf Independent-block GP:} Instead of blocking the covariance
  matrix, we propose a simple alternative where we apply the blocking
  at the level of the GP.  We split the data set into several smaller
  non-overlapping blocks along one or more input dimensions and
  build a separate GP model for each block, which would now have its
  own set of optimal hyperparameters. To predict the value at a new
  point, we would find which block it belonged to and use the model
  for that block. This approach has several benefits: it is easier to
  implement as the software requires minimal changes, and it can give more
  accurate results as the hyperparameters selected are optimal for
  each block as they can now be different across blocks. Further, it
  allows the possibility of overlapping blocks to build the GP model, which
  can help in reducing the prediction error at the interior block
  boundaries. We refer to this algorithm as the independent-block GP
  method and show how we can exploit the characteristics of the data
  to select the blocks.

\end{itemize}

%
\subsection{Using an iterative solver}
\label{sec:iterative}
%

An alternate way of introducing approximation in the linear solver
in an attempt to reduce its computational cost is to use an iterative
solver as we can stop the iterations before an accurate solution has
been found.  Since the covariance matrix
(Equation~\ref{eqn:covar_matrix}) is symmetric positive definite, an
obvious choice is the conjugate gradient method~\cite{saad03:book},
described in Algorithm~\ref{algo:cg}.

\begin{algorithm*}
\caption{The Conjugate Gradient algorithm to solve ${\bf C} {\bf u} = {\bf v}$}
\label{algo:cg}
\begin{algorithmic}[1]

\STATE Set the threshold $\epsilon$ for the two norm of the residual and $max_{iter}$, the maximum number of iterations.

\STATE Set the initial guess for the solution, ${\bf u}_0$.

\STATE Set initial residual ${\bf r}_0 = {\bf v} - {\bf C} {\bf u}_0$ and initial direction vector ${\bf p}_0 = {\bf r}_0$.

\FOR{$k = 0$ to $iter_{max}$}

\STATE Calculate $\alpha_k = \frac{{\bf r}_k^T {\bf r}_k}{{\bf p}_k^T {\bf C} {\bf p}_k}$

\STATE Update solution vector: ${\bf u}_{k+1} = {\bf u}_k + \alpha_k {\bf p}_k$

\STATE Update residual vector: ${\bf r}_{k+1} = {\bf r}_k - \alpha_k {\bf C} {\bf p}_k$

\STATE Calculate $\beta_k = \frac{{\bf r}_{k+1}^T {\bf r}_{k+1}}{{\bf r}_k^T {\bf r}_k}$

\STATE If $\sqrt{ {\bf r}_{k+1}^T {\bf r}_{k+1} } < \epsilon$ or $(k+1) > max_{iter}$, exit.

\STATE Update direction vector: ${\bf p}_{k+1} = {\bf r}_{k+1} + \beta_k {\bf p}_k$

\ENDFOR
\end{algorithmic}
\end{algorithm*}

Though the conjugate gradient method is guaranteed to converge in $N$
iterations for a matrix of size $N$, this is rarely the case due to
round-off error, especially if the condition number of the matrix is
large.  Running the algorithm for $N$ iterations would also make it as
expensive as a direct solver as each iteration involves a
matrix-vector multiply, which is an $\mathcal O (N^2)$ operation. So,
we investigated how the solution accuracy was affected when we
stopped the iterations i) when the two-norm of the residual became
smaller than a threshold, and ii) after a fixed number of iterations
smaller than $N$.

%
\section{Understanding the solution of an inverse problem}
\label{sec:understand_soln}
%

\begin{figure*}[htb]
\centering
\begin{tabular}{cc}
\includegraphics[width=0.49\textwidth]{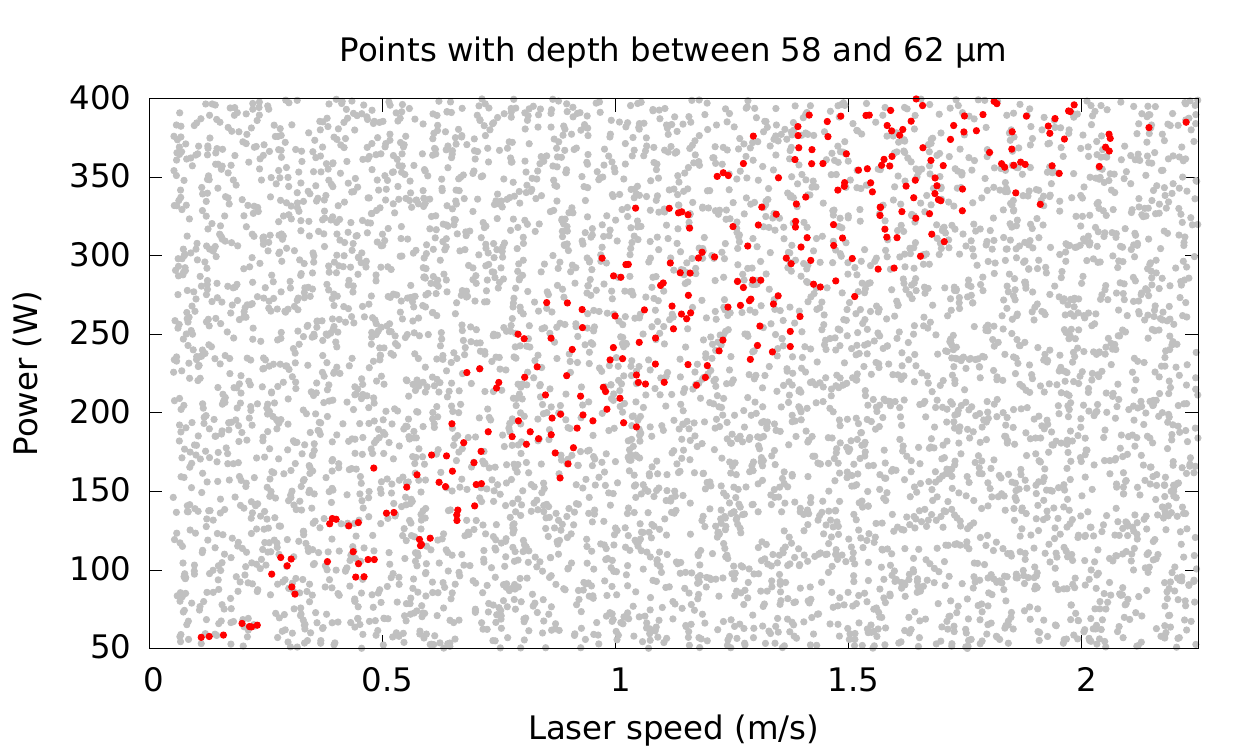} &
\includegraphics[width=0.49\textwidth]{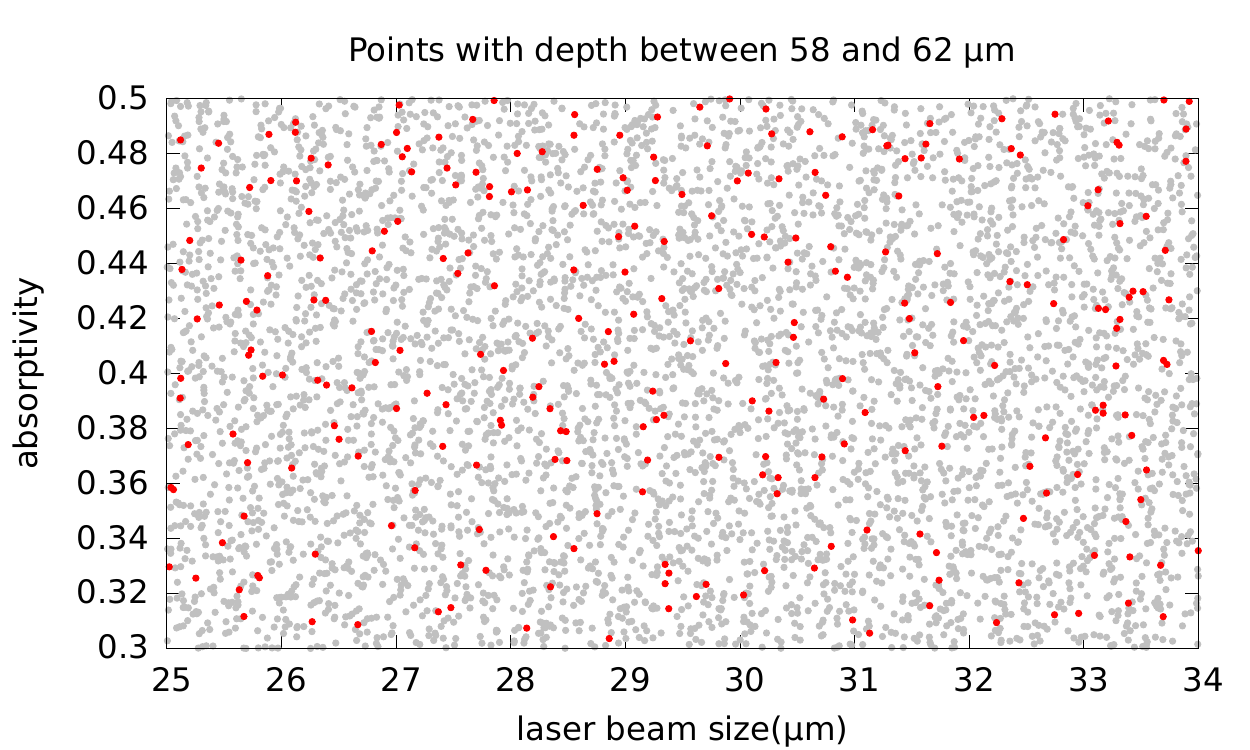} \\
\end{tabular}
\vspace{-0.2cm}
\caption{The sample points in the ET-5000 data set that have melt-pool
  depths in the range [58,62]\micron, shown in red. Left: in the laser
  power - laser speed space and right: in the beam size - absorptivity
  space, indicating a pattern in the former, but not in the latter. }
\label{fig:two_space_projection}
\end{figure*}

Once we have built a surrogate model for the forward problem in the
form of a Gaussian process, we can use the model to predict the values
of the outputs at a relatively dense set of sample points in the input
space. The solution to the inverse problem would then be the sample
points that meet the target range, as described in
Section~\ref{sec:data_desc}.  When the input space is high-dimensional
(greater than 2), it is a challenge to understand and interpret the
solution points in this space as they cannot be visualized easily. For
example, Figure~\ref{fig:two_space_projection} shows the locations of
the samples from the ET5000 data set with depth in the range
[58,62]\micron~ using two input dimensions. For problems with a
moderate to large number of input dimensions, such an approach can
soon become impractical.

In this paper, we consider two methods for the display of
high-dimensional data - parallel coordinate
plots~\cite{inselberg09:book} and Kohonen self-organizing
maps~\cite{kohonen00:book}, which we describe next.

%
\subsection{Parallel coordinate plots}
\label{sec:parplots}
%

Parallel coordinate plots, or parallel plots, are a visual technique
useful for displaying high-dimensional
data~\cite{inselberg09:book}. Unlike a traditional Cartesian
coordinate plot, where we plot two or three dimensions using axes that
are orthogonal to each other, in a parallel plot, the axes for all
dimensions are placed parallel to each other and perpendicular to a
horizontal line. This allows us to visualize more than three
dimensions at a time in one plot.

To create a parallel plot for a data set, we first scale the values in
each dimension to lie between 0 and 1, or any other minimum and
maximum value. Then, an instance in the data set, which would be a
point in a $d$-dimensional space, will be represented by a poly line,
that is, a sequence of line segments connecting the scaled values of
each dimension for that instance on each of the parallel axes.

In our earlier work~\cite{kamath14:density}, we used a parallel plot
for the ET462 data set to gain insight into the data. We found that
laser speed and power were important variables as they were correlated
with the depth of the melt pool. As one would intuitively expect,
higher laser power and lower laser speed resulted in deep melt pools,
while lower laser power and higher laser speed resulted in shallow
melt pools. This allowed us to easily identify input parameters for
the moderate-sized melt pools of interest. The other two inputs did
not appear to be correlated to the depth.

%
\subsection{Kohonen self-organizing maps}
\label{sec:soms}
%

A self-organizing map (SOM)~\cite{kohonen00:book} is a form of
unsupervised neural network that is used to visualize and understand a
high-dimensional data set by mapping it into a two-dimensional array
using a non-linear projection. While extensive research in SOMs has
been conducted over the years, in this paper, we consider a simple,
parameter-free, fast version called the Batch Map
algorithm~\cite{kohonen93:things}. Our choice is motivated by our
previous work~\cite{ponmalai2019:som}, where we investigated the
effects of various parameters in the more complex variants of SOMs and
found that the simpler versions worked just as well as the more complex
ones, but required fewer parameters.

\begin{figure}[htb]
\centering
\begin{tabular}{c}
\includegraphics[width=0.45\textwidth]{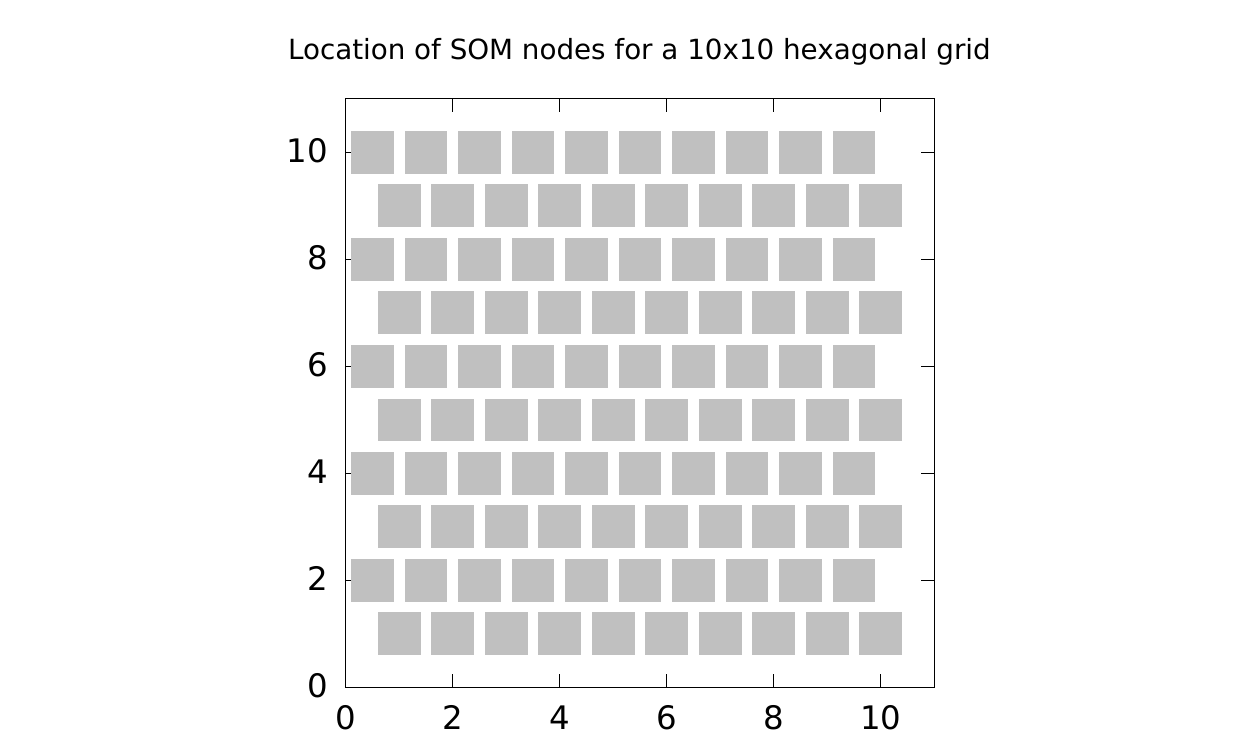} 
\end{tabular}
\vspace{-0.2cm}
\caption{The locations of grid nodes for a $10 \times 10$ hexagonal
  grid. Each node has six immediate neighbors that are equidistant
  from it, unlike a square grid, where a node has four
  neighbors at a unit distance and four neighbors $\sqrt{2}$ away. }
\label{fig:hex_som}
\end{figure}

A two-dimensional SOM model is associated with a grid or lattice of
nodes in two dimensions. The grid is defined by an architecture which
includes the number of nodes, $NX$ and $NY$ in the $x$ and $y$
direction, respectively, and a grid layout, square or hexagonal, which
defines the relative locations of the nodes on the grid. For example,
Figure~\ref{fig:hex_som} shows the layout of a $10\times10$ SOM on a
hexagonal grid, so named as each interior node has six neighboring nodes that
are equidistant from it. We use only hexagonal grids in our work as
they have a more uniform neighborhood structure than square grids.
Associated with each of the $NX \times NY$ grid nodes are two vectors:
a two-dimensional coordinate vector with the ($x$,$y$) coordinates of
the node in the grid and a $d$-dimensional weight vector, sometimes
called a model vector, where the $d$ dimensions correspond to the
features in the input data set.

The SOM algorithm starts by initializing the weight vector at each
node of the grid. There are many ways in which this can be done; in
our work, we initialize each dimension of the weight vector at a node
randomly with a value between the minimum and maximum of the
corresponding feature in the data set. Then, the weight vectors are
updated using the data set so that they become spatially, globally
ordered. As we iterate many times through the instances in the data
set, the weight vectors of nodes that are closer to each other in the
grid become similar, while the weight vectors of nodes that are far
from from each other become less
similar~\cite{kohonen2013:essentials}.  

In the online version of the SOM algorithm, the weight vectors are
updated by considering an instance at a time, as follows: starting
with a random instance, ${\bf x}_j$, in the data set, we first find
the node $c$, whose weight vector ${\bf m}_c$ is the closest to the
instance:
\begin{equation}
c = \underset{i \in {1,\ldots,NX \times NY} }{\arg\min} \| {\bf x}_j - {\bf m}_i \|
\label{eqn:ref_node}
\end{equation} 
where the distance is calculated in the feature space. The node $c$ is
called the reference node for the instance ${\bf x}_j$.  The weight
${\bf m}_c$ is then updated to be closer to the instance ${\bf
  x}_j$. In addition, the weights in the nodes in a neighborhood
around node $c$ are also updated to be closer to ${\bf x}_j$. The
neighborhood is defined using the coordinate vectors at the nodes and
reduces in size as the iterations progress. The process is repeated
with another randomly-selected instance, until we have processed $N$
instances, completing one iteration. As the instances are selected
randomly, we may process some instances more than once and others none
at all during an iteration. The iterations repeat until the node
weights stabilize.

A faster and simpler version of the SOM algorithm is the batch version
which finds the closest reference nodes to all instances first. It
then updates the weight vector in each reference node with the nearest
instances. This is followed by updates to the nodes in the
neighborhood of the reference node of each instance.  Typically, the
weights in the neighborhood nodes are updated based on how far these
nodes are from the reference node, with farther away nodes being
updated by a smaller amount. In the Batch Map
algorithm~\cite{kohonen93:things}, which is described in
Algorithm~\ref{algo:batch_map}, this condition is simplified even
further and all nodes in the neighborhood are updated to the same
extent regardless of their distance to the reference node.

Both the online and batch versions of the self-organizing map operate
on the features of the instances and organize the instances in this
input feature space, without taking into account the output associated
with each instance. The batch version can be considered as
deterministic compared to the stochastic online version.  Both
algorithms start with a random assignment of initial weight
vectors. But, as there is no randomness in the processing of the
instances in the batch algorithm, the weight vectors stabilize and,
unlike the online version, give the same results when the algorithm is
run repeatedly.

\begin{algorithm*}
\caption{The Batch Map algorithm}
\label{algo:batch_map}
\begin{algorithmic}[1]

\STATE Set $NX$ and $NY$, the number of grid nodes in
the $x$ and $y$ direction, respectively.

\STATE Generate the grid coordinates for each node.

\STATE Set $r_{max}$ and $r_{min}$, the initial and final radii of the
neighborhood, and $r_{iter}$, the number of iterations over which the
radius is decreased from $r_{max}$ to $r_{min}$.

\STATE Set the current neighborhood radius to $r_{max}$ and define $r_{delta} = (r_{max} - r_{min}/r_{iter})$.

\STATE Set $iter_{max}$, the maximum number of iterations.

\STATE Initialize the weight vectors at each node.

\FOR{$i=1$ to $iter_{max}$}

\FOR{$inode=1$ to $NX \times NY$}
\STATE List the instances nearest to the weight vector of $inode$ or nodes in its neighborhood.
\ENDFOR

\FOR{$inode=1$ to $NX \times NY$}
\STATE Replace the weight vector at $inode$ by the mean of the instances in its list. 
\ENDFOR

\STATE Exit if SOM has converged. 

\STATE Reduce the radius of the neighborhood by $r_{delta}$, without going below $r_{min}$.
\ENDFOR
\end{algorithmic}
\end{algorithm*}

The Batch Map algorithm is summarized in
Algorithm~\ref{algo:batch_map}. This algorithm has six parameters:
$NX$ and $NY$, the number of grid nodes in the $x$ and $y$ dimension,
respectively; $r_{max}$ and $r_{min}$, the initial and final radii of
the neighborhood, respectively; $r_{iter}$, the number of iterations
over which the radius is decreased from $r_{max}$ to $r_{min}$; and
$iter_{max}$, the maximum number of iterations of the SOM algorithm,
where one iteration involves a pass through all instances in the data
set. Instead of stopping after a fixed number of iterations, we can
also monitor a convergence metric such as
\begin{equation}
\frac{1}{N} \sum_{i=1}^{N} \| {\bf x}_i - {\bf m}_c \|
\end{equation} 
which is the average distance of each instance to the nearest node. A
SOM has stabilized when this metric does not change over iterations or
the maximum number of iterations has been reached. Since the Batch Map
is fast enough for our data sets, we just let it run for $iter_{max}$
iterations, which is chosen to be suitably large so that the metric
above stabilizes.

For our work with SOMs, we use two grids, one of size $30 \times 30$
and the other of $10 \times 10$. For the larger grid, we use $r_{max}
= 20$, $r_{min} = 1$, $r_{iter} = 150$ and $iter_{max} = 200$. The
corresponding values for the smaller grid are 8, 1, 100, and
120. $r_{max}$ was set as a large fraction of the grid size $NX$,
while $r_{iter}$ was set by experimentation to be large enough to
allow the map to self organize. We found that the ierations stabilized
soon after $r_{iter}$, which led to the choice of $iter_{max}$. Batch
Map is a very fast algorithm, and as it required no more than a few
seconds for our data sets, we were able to experiment with various
parameter values. We also found that the results were remarkably
robust to the parameter settings.

%
\section{Experimental results and discussion}
\label{sec:results}
%

We next present the results of our experiments with approximate
solvers for GP using the ET462 data set and high-dimensional data
exploration methods with the ET5000 data set.

All times reported in this paper are for codes run on an Intel Xeon
W-2155 workstation with ten cores, two threads per code, and 62GB of
memory. The codes are run on a single core. 

%
\subsection{Evaluating independent-block GP}
\label{sec:results_taper}
%

In independent-block GP, we divide the input domain into 
blocks of sample points and ignore the dependence between the blocks. 
To accomplish this, we need to determine:

\begin{itemize}

\item {\bf The number of blocks to use:} Suppose we divide the data
  into $B$ blocks. Then, with $N$ data points in the data set, each
  block will have $(N/B)$ data points if $N$ is divisible
  by $B$; if not, we can divide the data so the blocks have roughly
  equal number of points. Solving a linear system with $N$ equations
  using a direct method requires $\mathcal{O} (N^3)$ floating point
  operations. For $B$ blocks, this reduces to $B * \mathcal{O} (
  (N/B)^3 )$, a $B^2$ reduction in the computations. This
  makes blocking a very appealing idea if it does not result in lower
  predictive accuracy. To maintain the accuracy, we need to ensure
  that each block has a sufficient number of data points, which limits
  how many blocks we can create. In our work we use $B = 2 \
  \text{and} \ 4$.

\item {\bf The dimension(s) along which to block the data:} We tried
  two options - blocking along just one dimension and blocking using
  two of the four input dimensions. Blocking along additional
  dimensions, though an option for large datasets, would have resulted
  in too many blocks and too few samples in each block for our dataset.

\end{itemize}

To generate results for hyper-parameter optimization using independent-block
GP, we considered 100 samples in the six-dimensional GP
hyper-parameter space: $ \sigma_f$, $\sigma_n $, and the four weighted length scales
 $l/w_i, i = 1, \ldots, d$ that are used in the distance calculation. As with
the generation of samples for the ET5000 data set, these
six-dimensional samples were also generated using the best candidate
method~\cite{mitchell91:sampling}, which, by placing the samples
randomly, but far apart from each other, allowed us to explore the
hyperparameter space fully.  For a fair comparison, we used the same
100 hyperparameter samples across all our experiments.  We compared
the quality of the prediction using the predicted vs. actual value for
the instances in each block, where the predicted value is obtained
using a GP model built with the remaining instances in the block.

For the direct solver, we use the optimized BLAS and LAPACK solvers
for the symmetric matrix stored in the packed storage format, while
the conjugate gradient method uses the BLAS 2 matrix vector multiply
algorithm, with the matrix stored in the symmetric packed
storage~\cite{anderson99:lapack}. All computations are in double
precision. We did invesitgate the use of single precision Cholesky
solvers, but found that the system had to be very large for a
substantial reduction in time.  In addition, depending on the
hyperparameters, the covariance matrix could become singular in single
precision.

In the tables with the results for independent-block GP, we also report the
total time for hyperparameter optimization to find the best of 100
samples for a data set with $M$ instances, where $M$ is the size of a
block. For each hyperparameter sample, we perform a leave-one-out
(LOO) analysis that involves the creation and solution of $M$ systems
of equations, each of size $(M-1)$. In addition, we perform two
leave-one-out analyses, one with a default set of hyperparameters and
the other with the optimal set. For these two cases, we also generate
the variance on the prediction of the instance held out from the data.

\begin{figure*}[htb]
\centering
\begin{tabular}{ccc}
\includegraphics[width=0.35\textwidth]{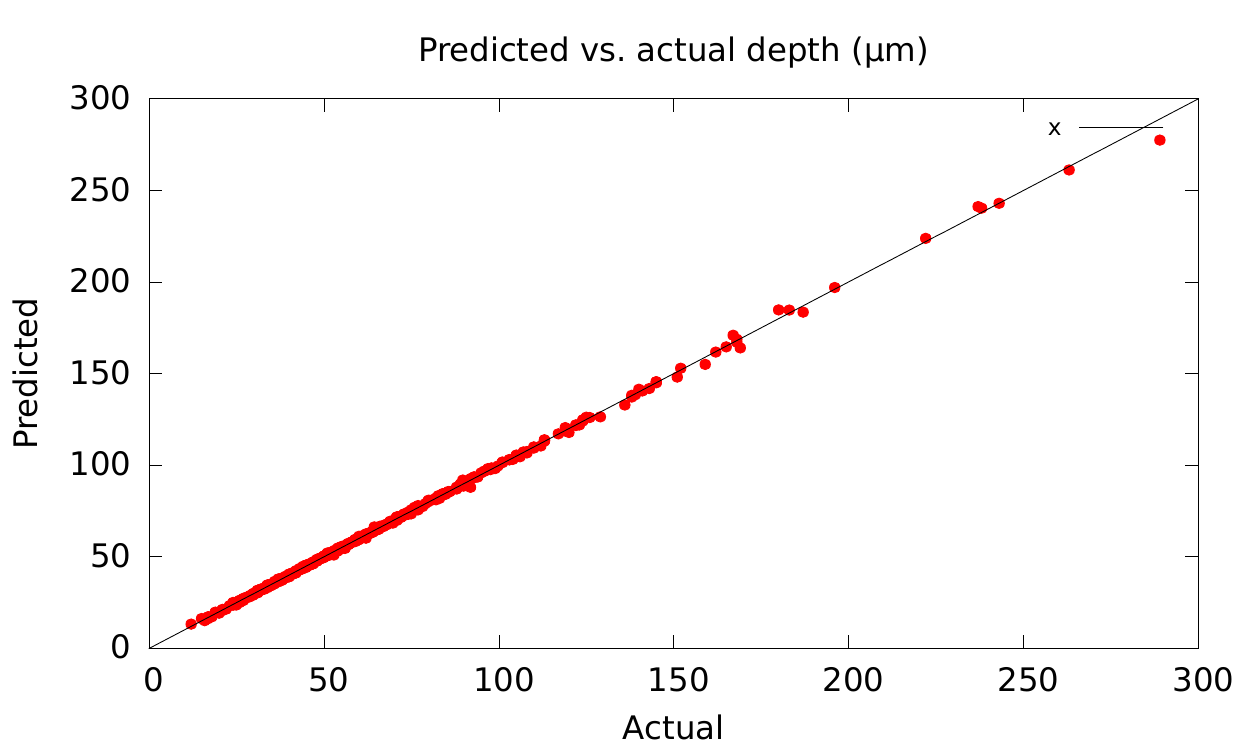} \\
\end{tabular}
\vspace{-0.2cm}
\begin{center}
  \begin{tabular}{| c| c | c | c | c | c | c |} \hline
    $ \sigma_f$ & $\sigma_n $ & $l/w_1$ & $l/w_2$ & $l/w_3$ & $l/w_4$ & MAE (opt)\\
    \hline
    4.015e-05 & 2.080e-07 &  0.157 & 0.327 & 1.451 & 1.735 & 5.27e-07\\
    \hline
  \end{tabular} \\
\end{center}
\vspace{-0.3cm}
\caption{Predicted vs. actual values of the melt pool depth for the
  full ET462 data set (no blocking), obtained using the LOO method and
  the optimal hyperparameters listed in the table. MAE is the mean
  absolute error at the optimal parameters. Determining the optimal
  hyperparameters for the original data took 1979s.}
\label{fig:gp_orig}
\end{figure*}

\begin{figure*}[!htb]
\centering
\begin{tabular}{cccc}
\includegraphics[width=0.23\textwidth]{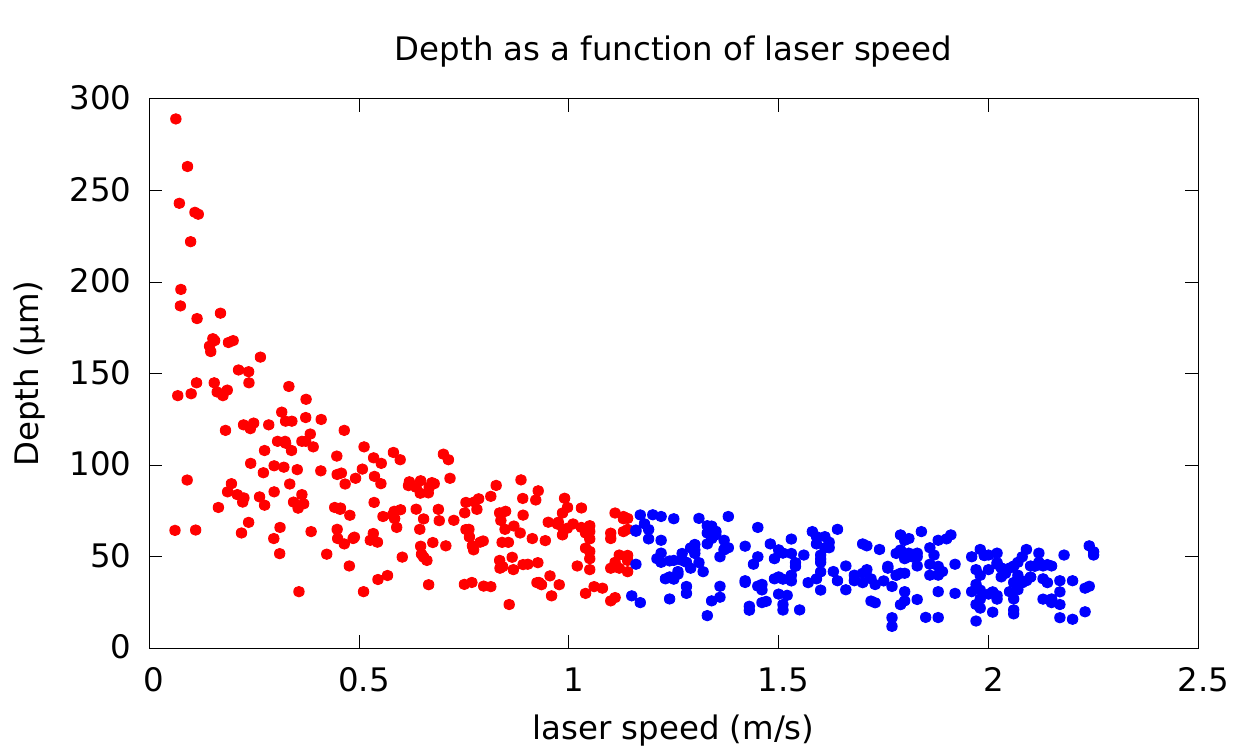} &
\includegraphics[width=0.23\textwidth]{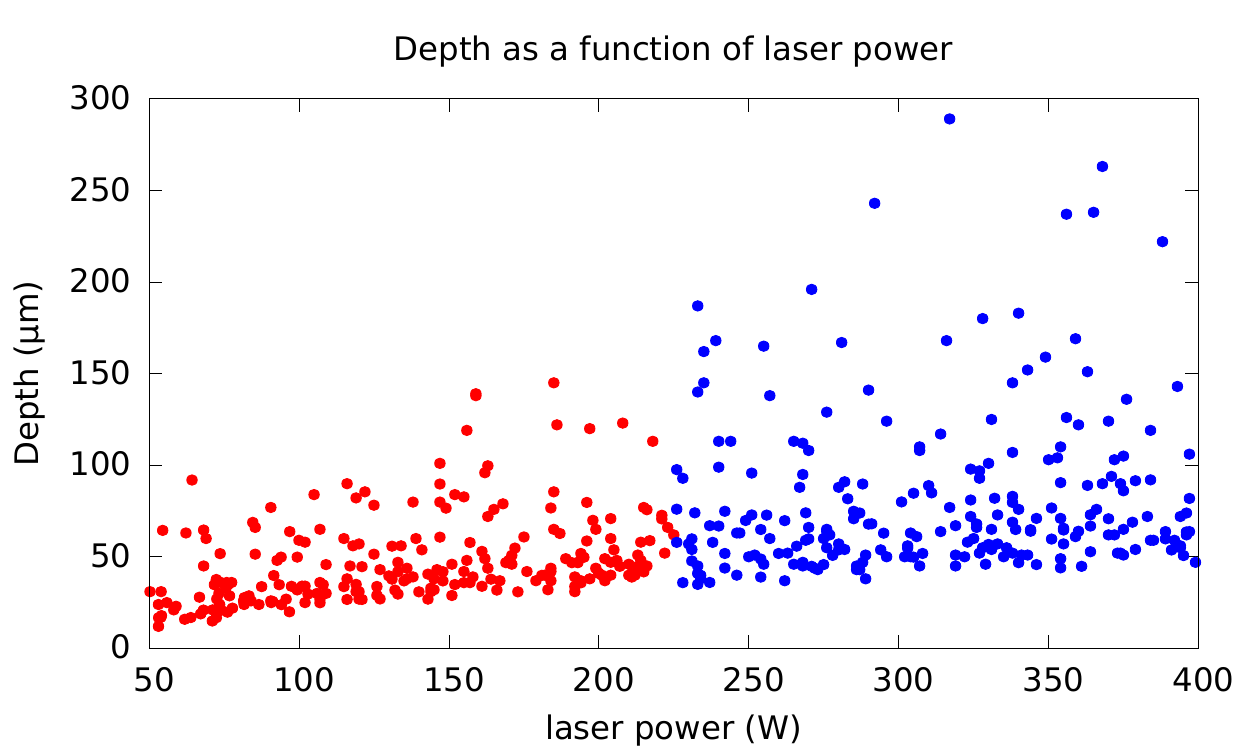} &
\includegraphics[width=0.23\textwidth]{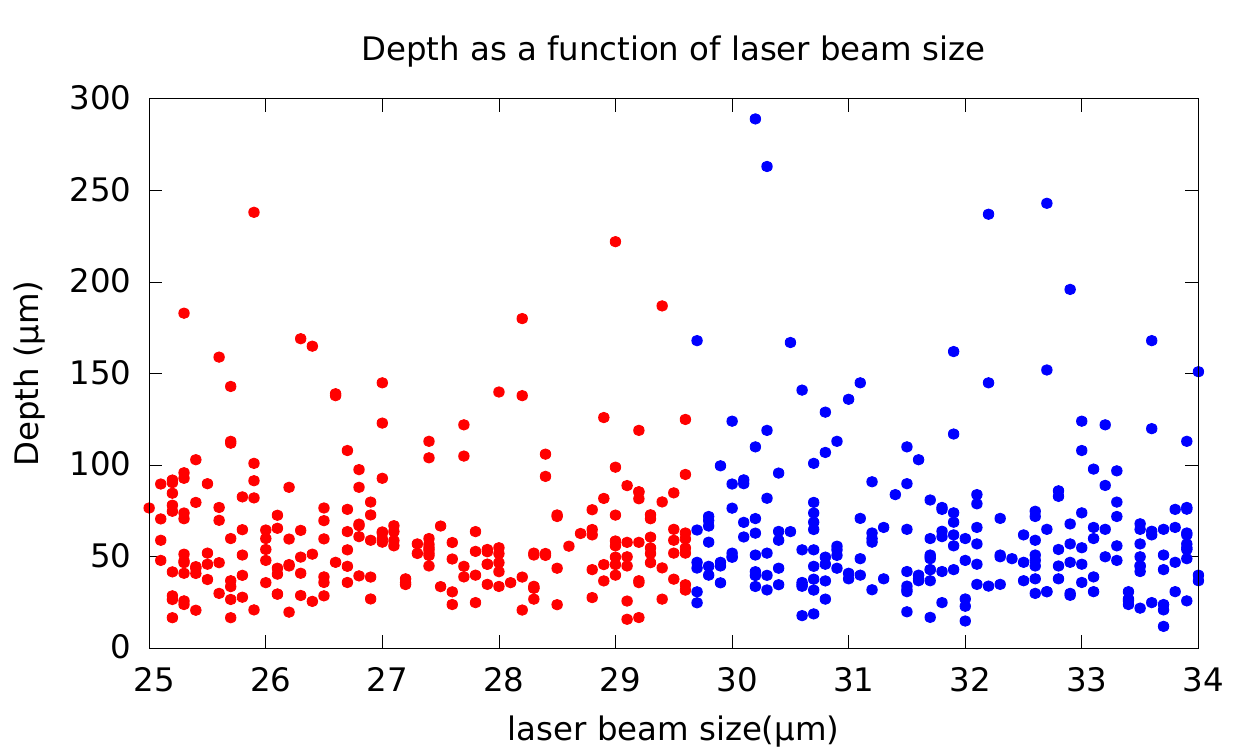} &
\includegraphics[width=0.23\textwidth]{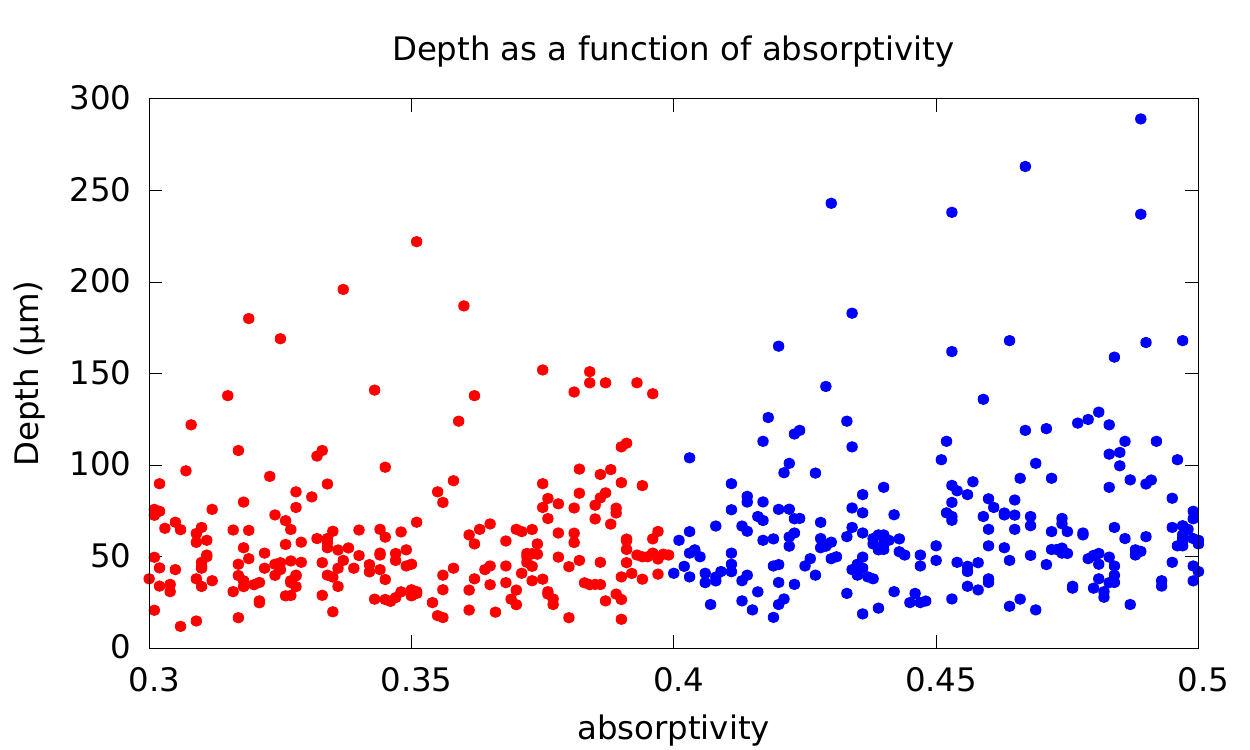} \\
\includegraphics[width=0.23\textwidth]{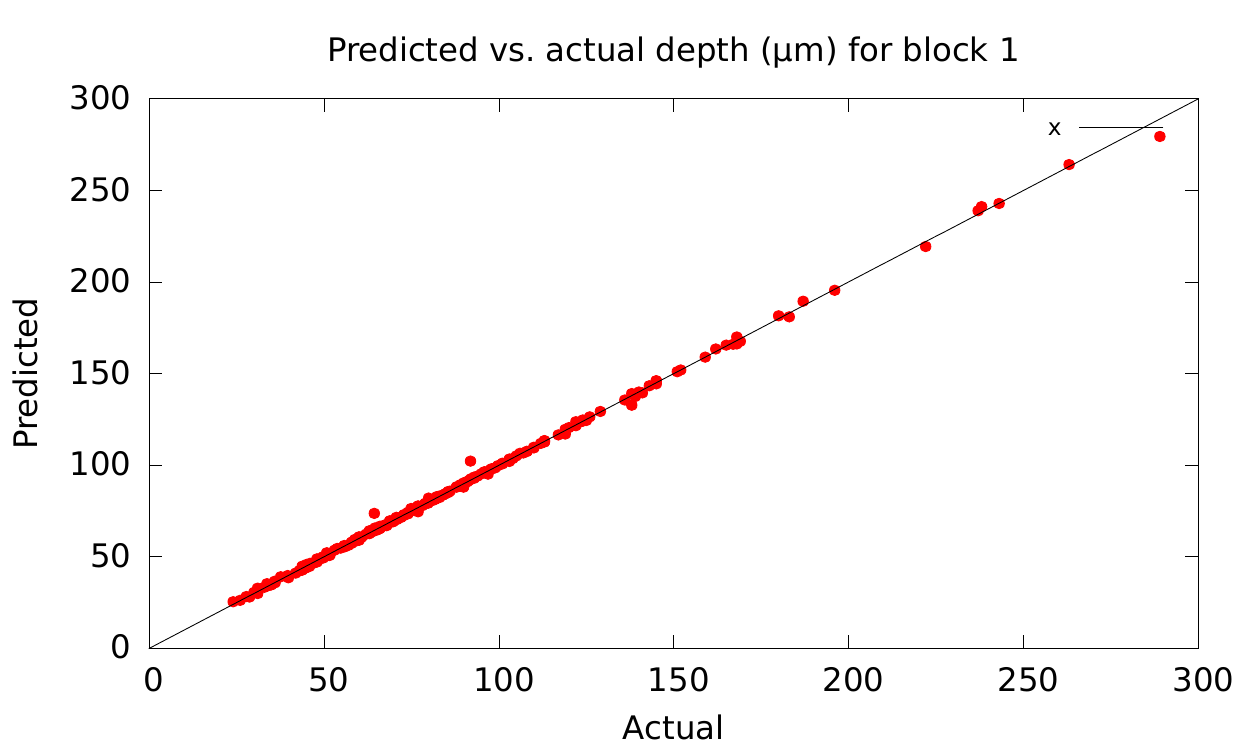} &
\includegraphics[width=0.23\textwidth]{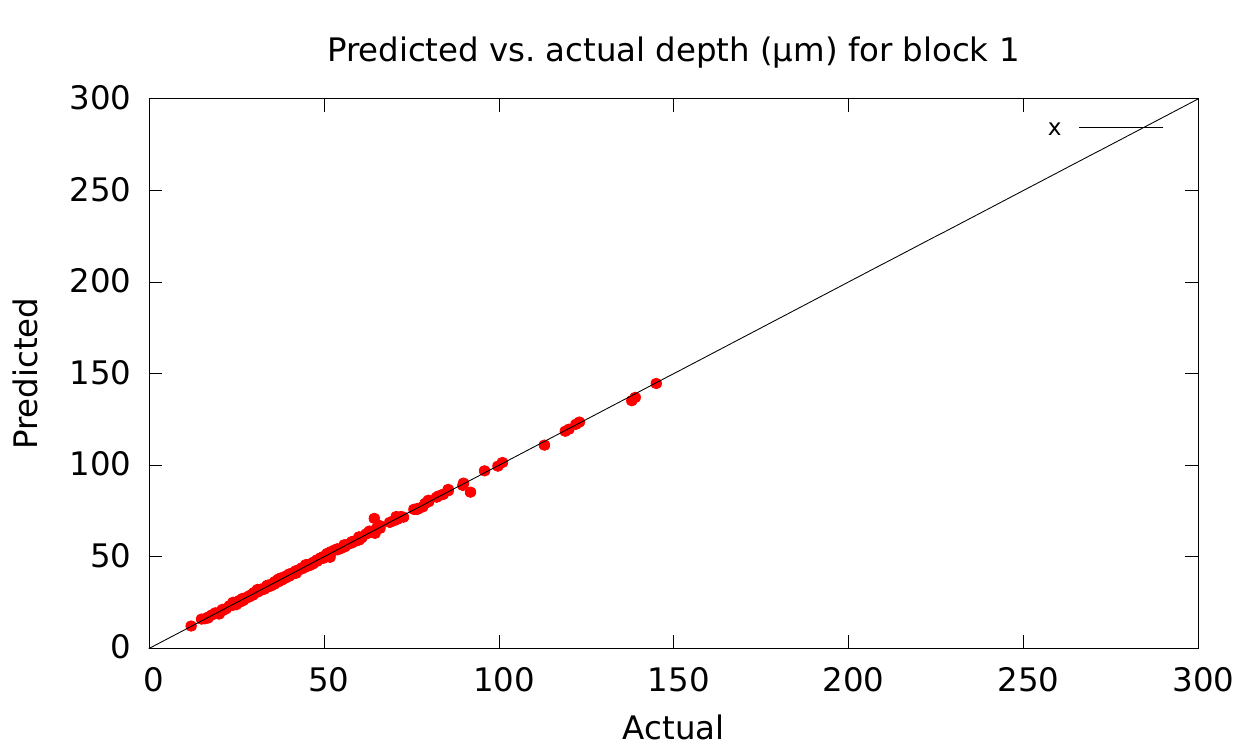} &
\includegraphics[width=0.23\textwidth]{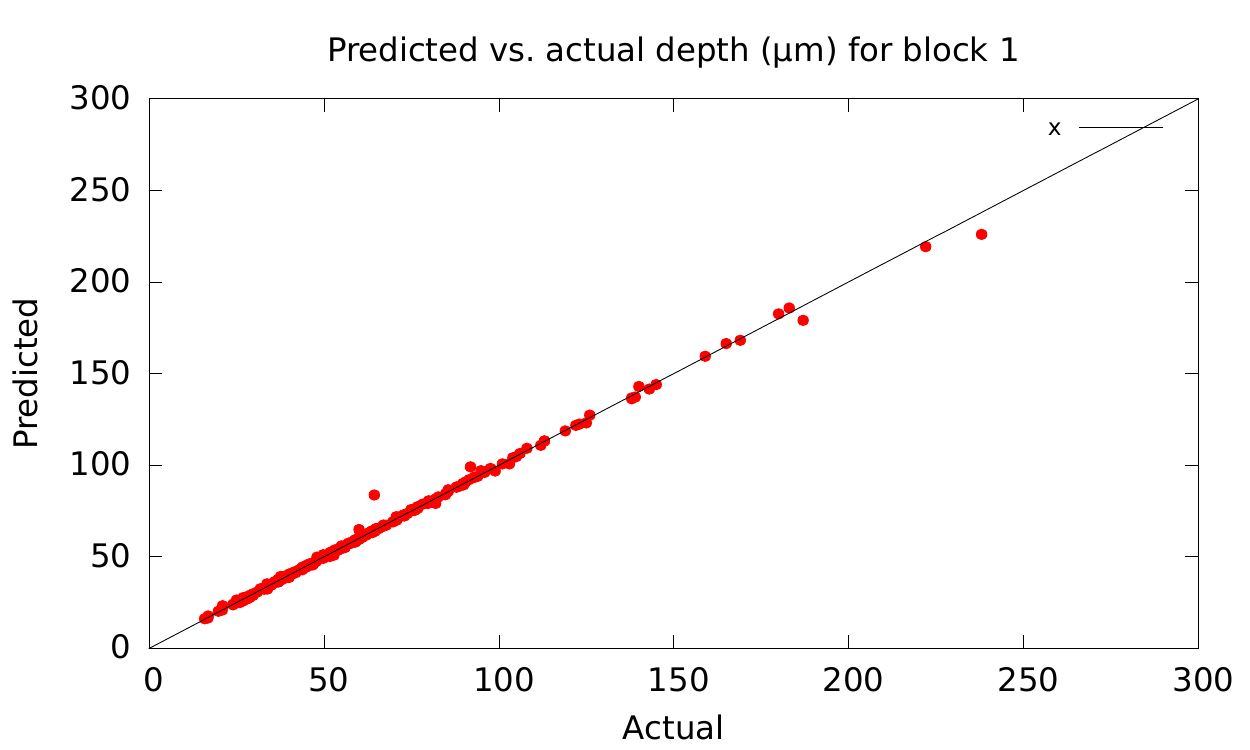} &
\includegraphics[width=0.23\textwidth]{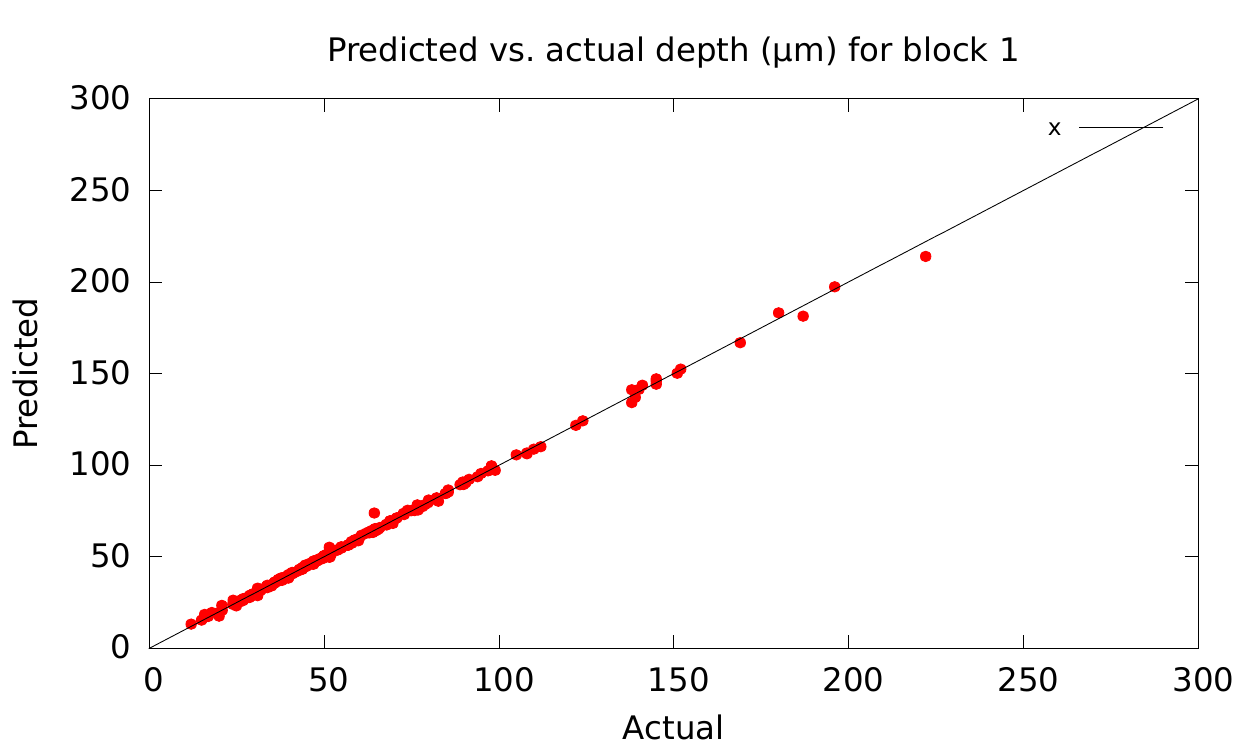} \\
\includegraphics[width=0.23\textwidth]{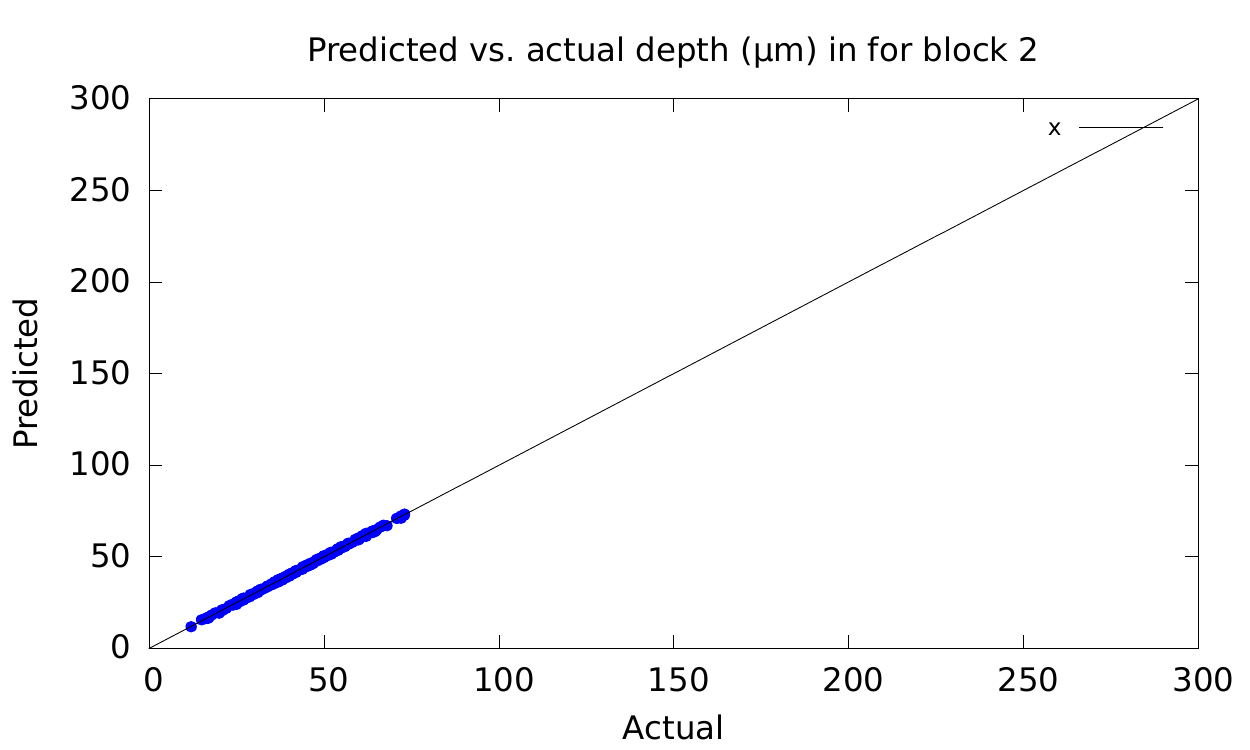} &
\includegraphics[width=0.23\textwidth]{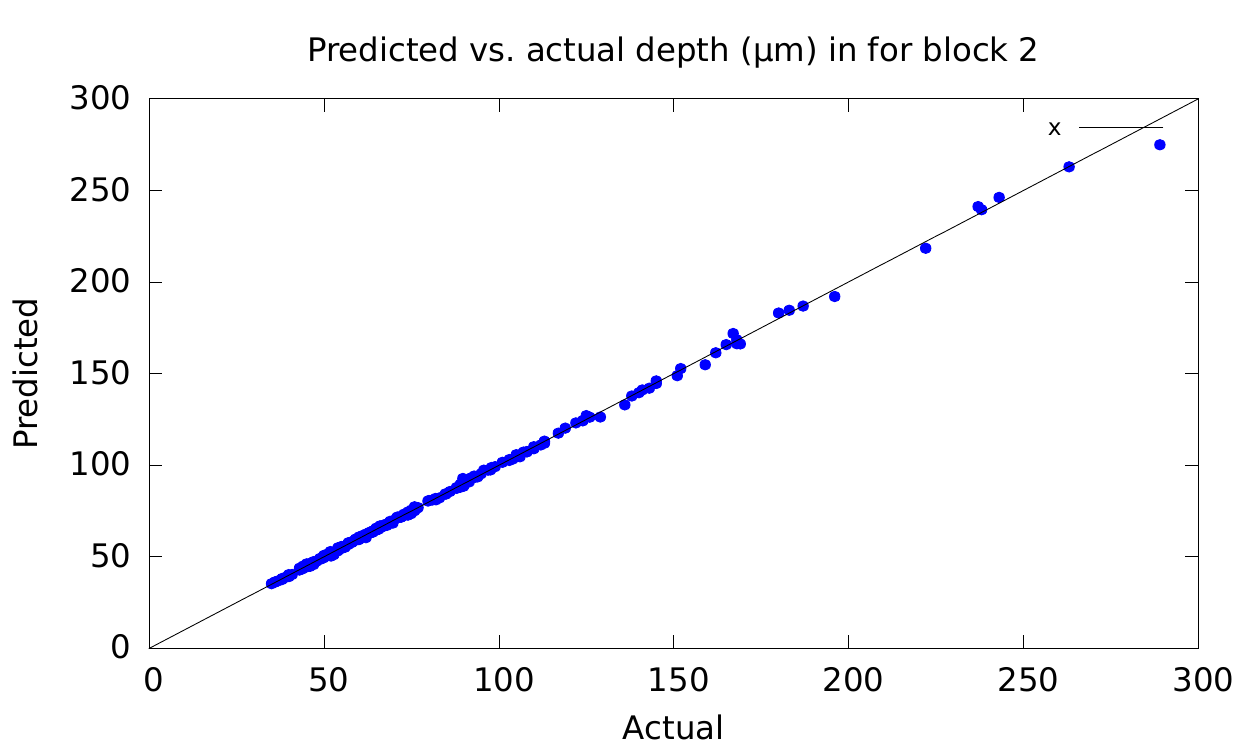} &
\includegraphics[width=0.23\textwidth]{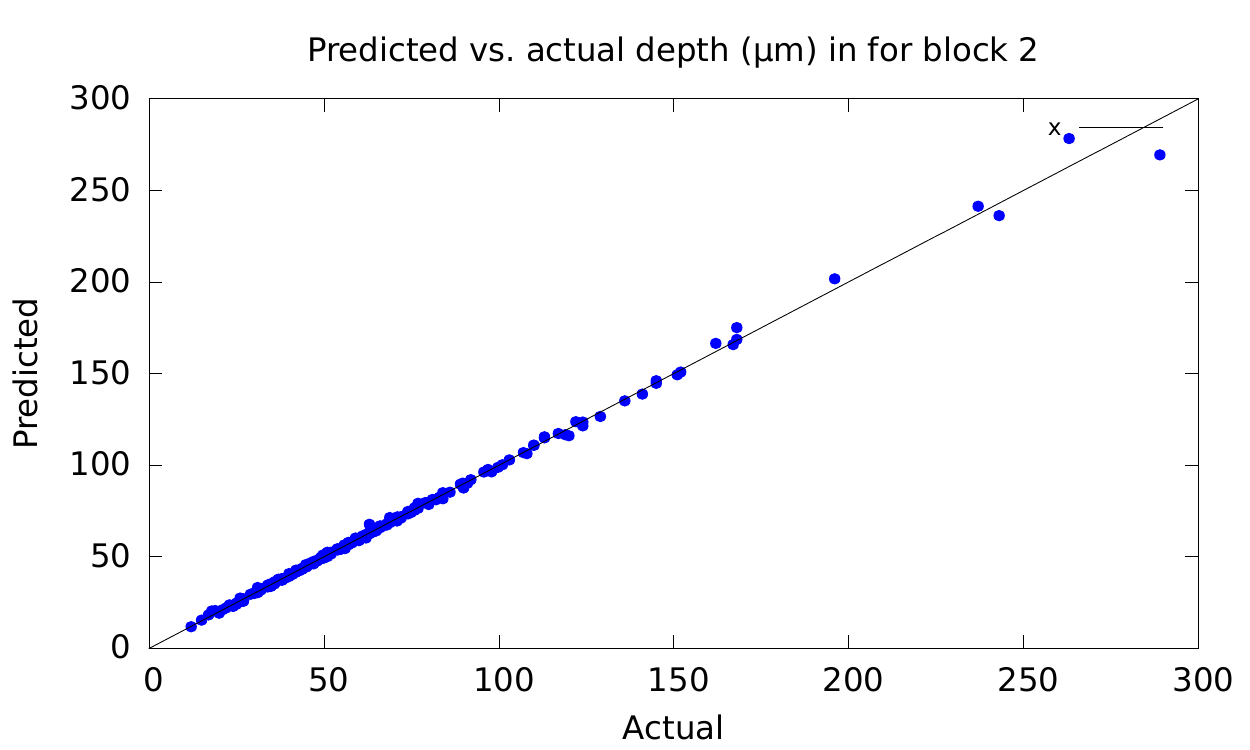} &
\includegraphics[width=0.23\textwidth]{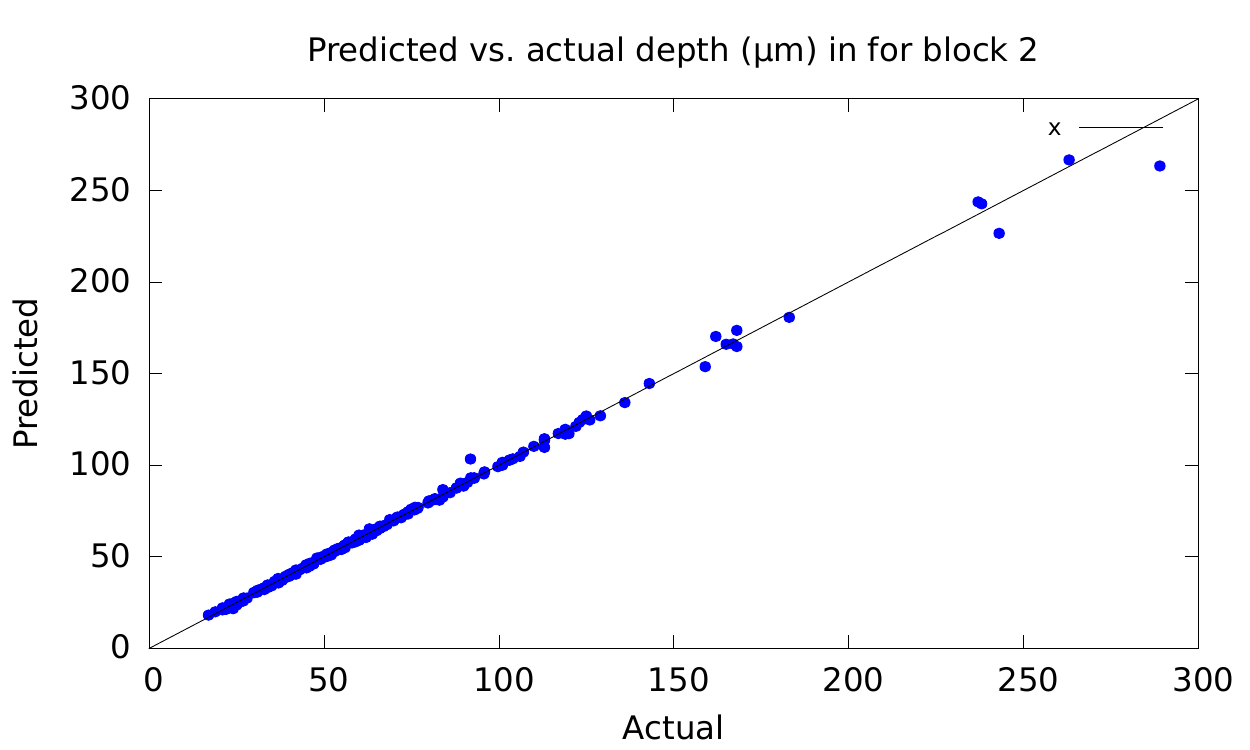} \\
\end{tabular} 
\vspace{-0.3cm}
\caption{GP predictions with two blocks along each dimension.  From
  left to right: blocking on the laser speed, laser power, laser beam
  size, and absorptivity. Top row: values of melt-pool depth as a
  function of the values of each input; red is low values and blue is
  high values. The predicted vs. actual values for the depth for
  points with lower values of the input (in red, middle row) and
  higher values of the input (in blue, bottom row).}
\label{fig:taper_oned_twob}
\end{figure*}

\begin{figure*}[!htb]
\centering
\begin{tabular}{cccc}
\includegraphics[width=0.23\textwidth]{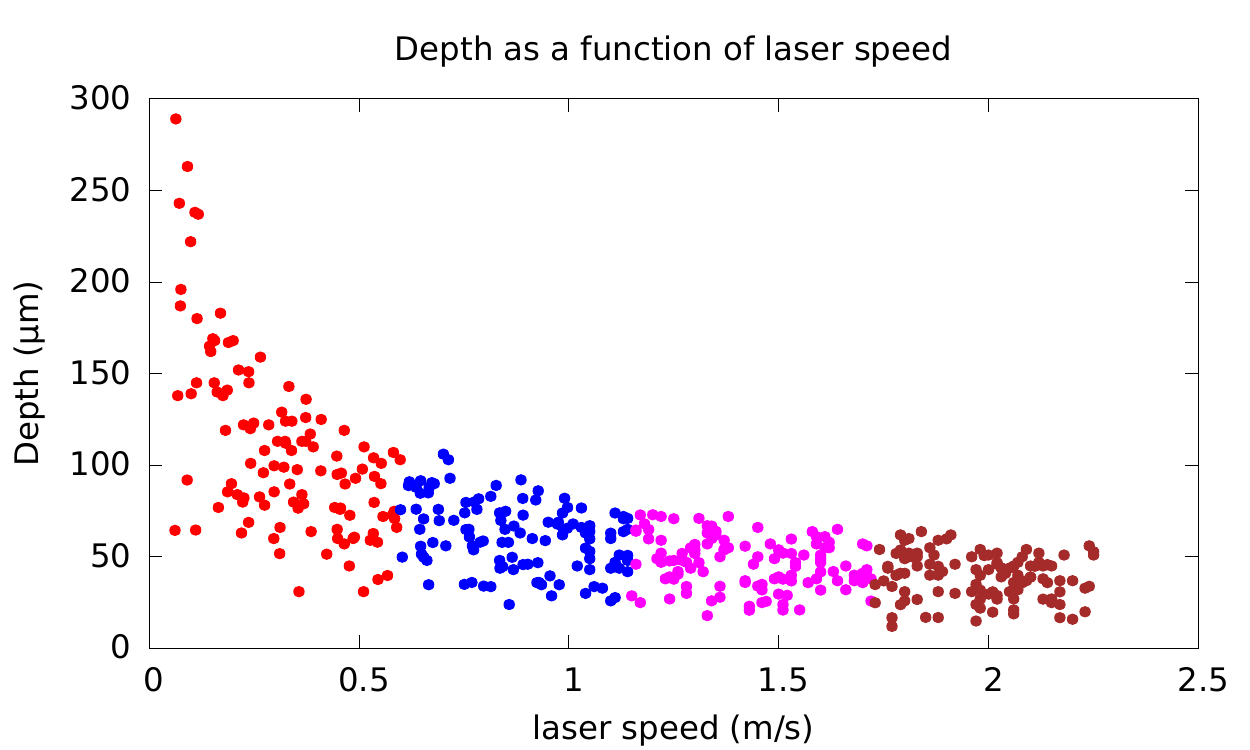} &
\includegraphics[width=0.23\textwidth]{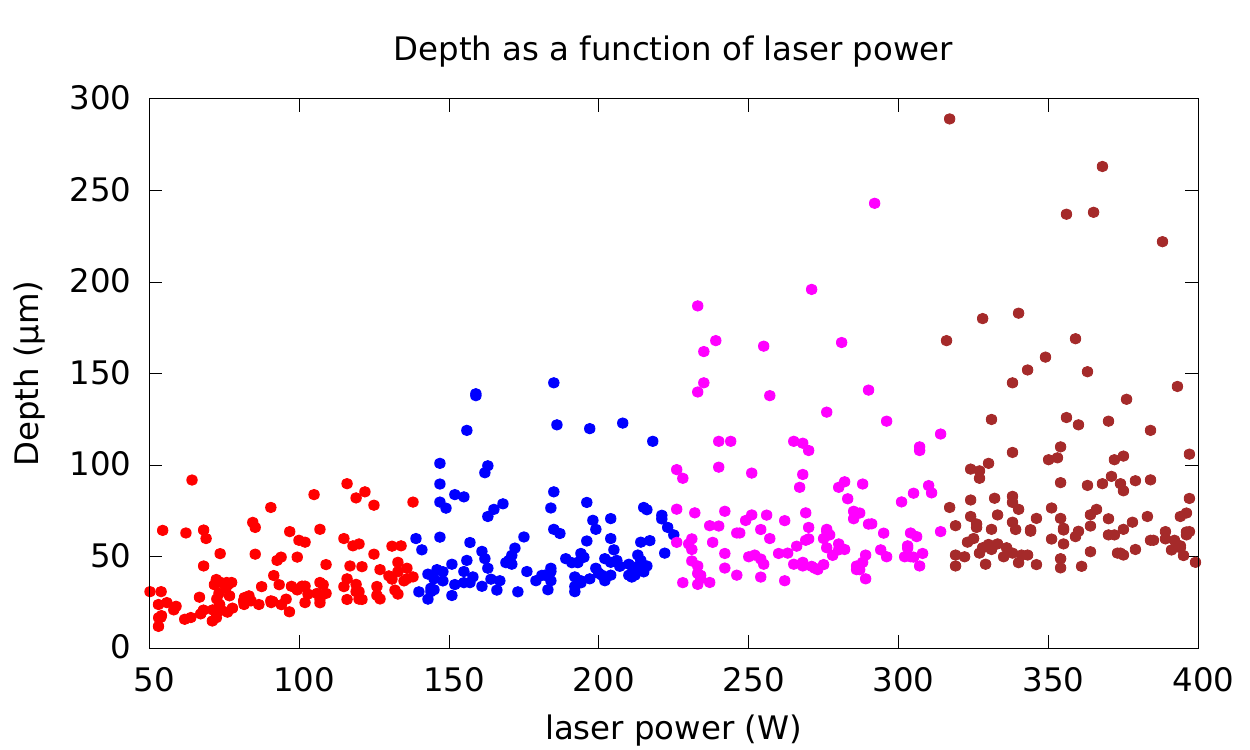} &
\includegraphics[width=0.23\textwidth]{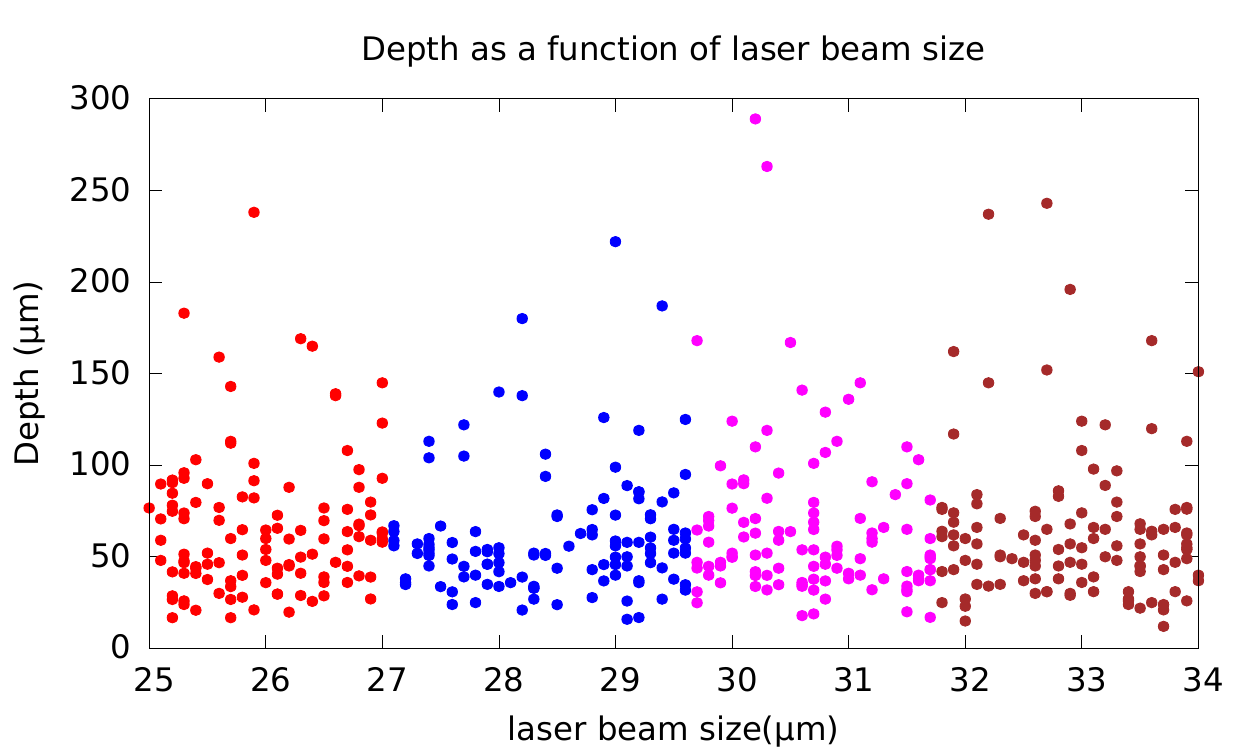} &
\includegraphics[width=0.23\textwidth]{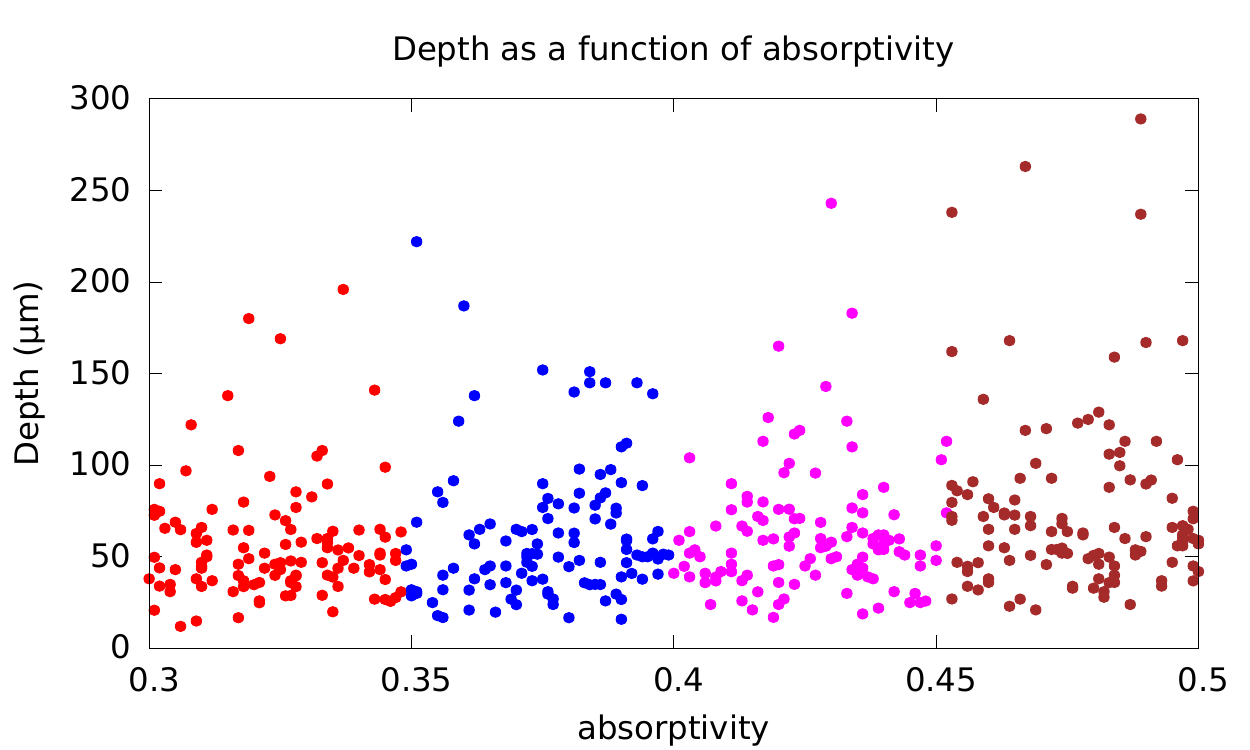} \\
\includegraphics[width=0.23\textwidth]{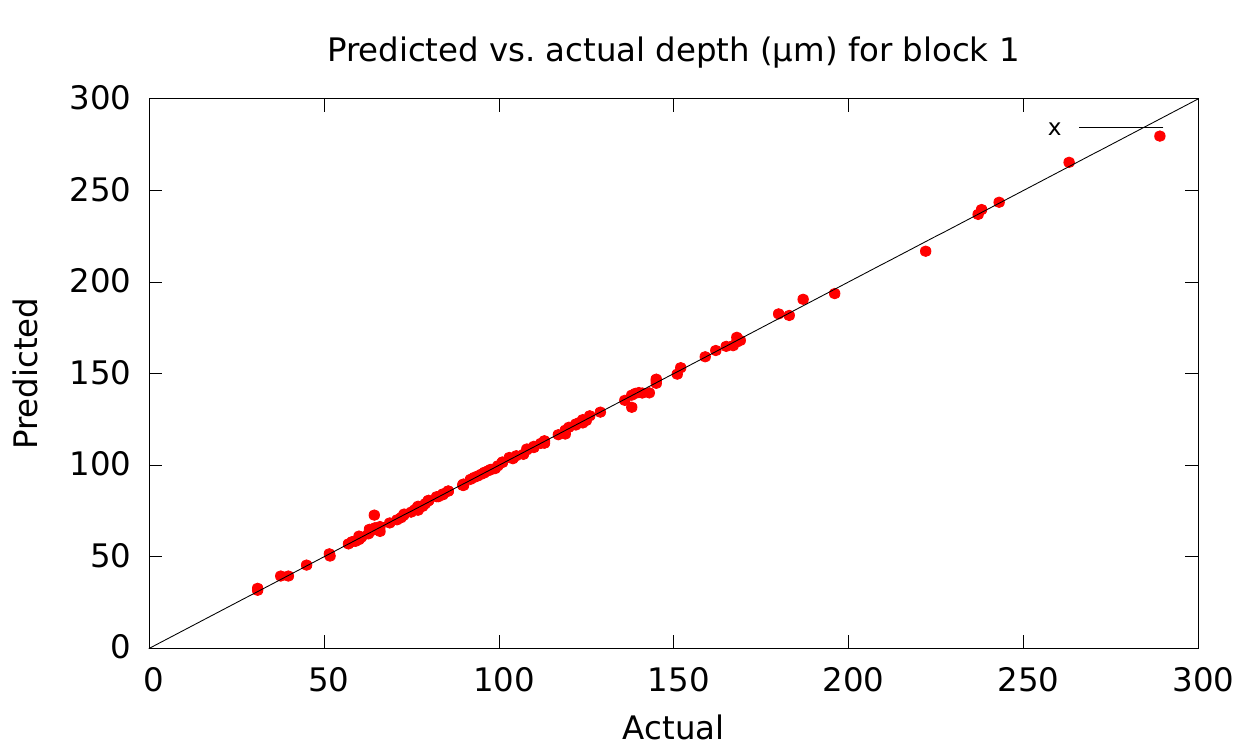} &
\includegraphics[width=0.23\textwidth]{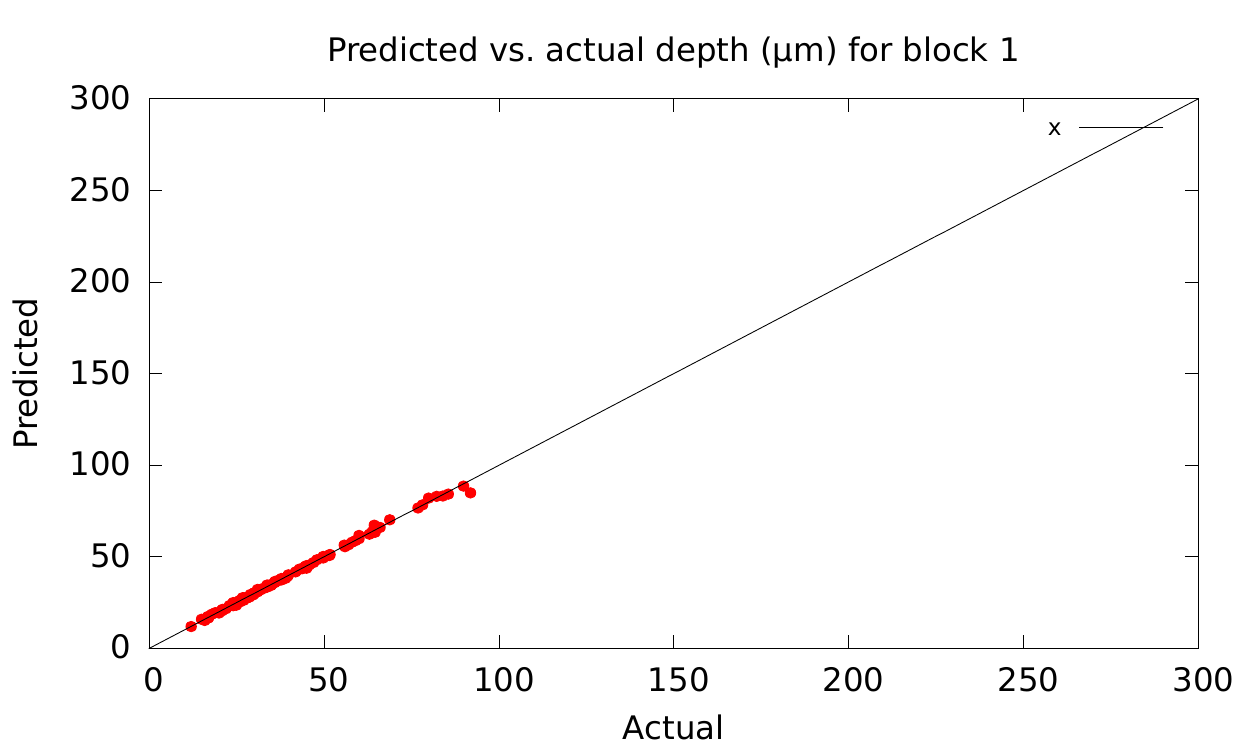} &
\includegraphics[width=0.23\textwidth]{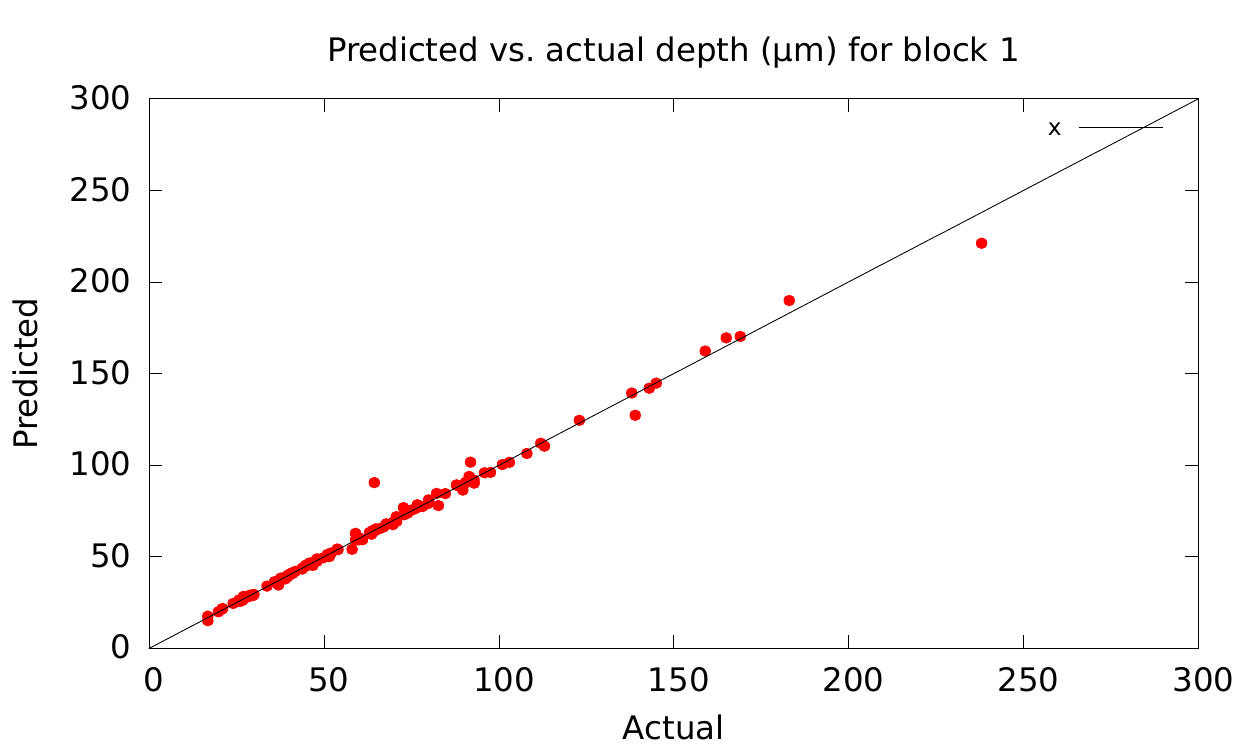} &
\includegraphics[width=0.23\textwidth]{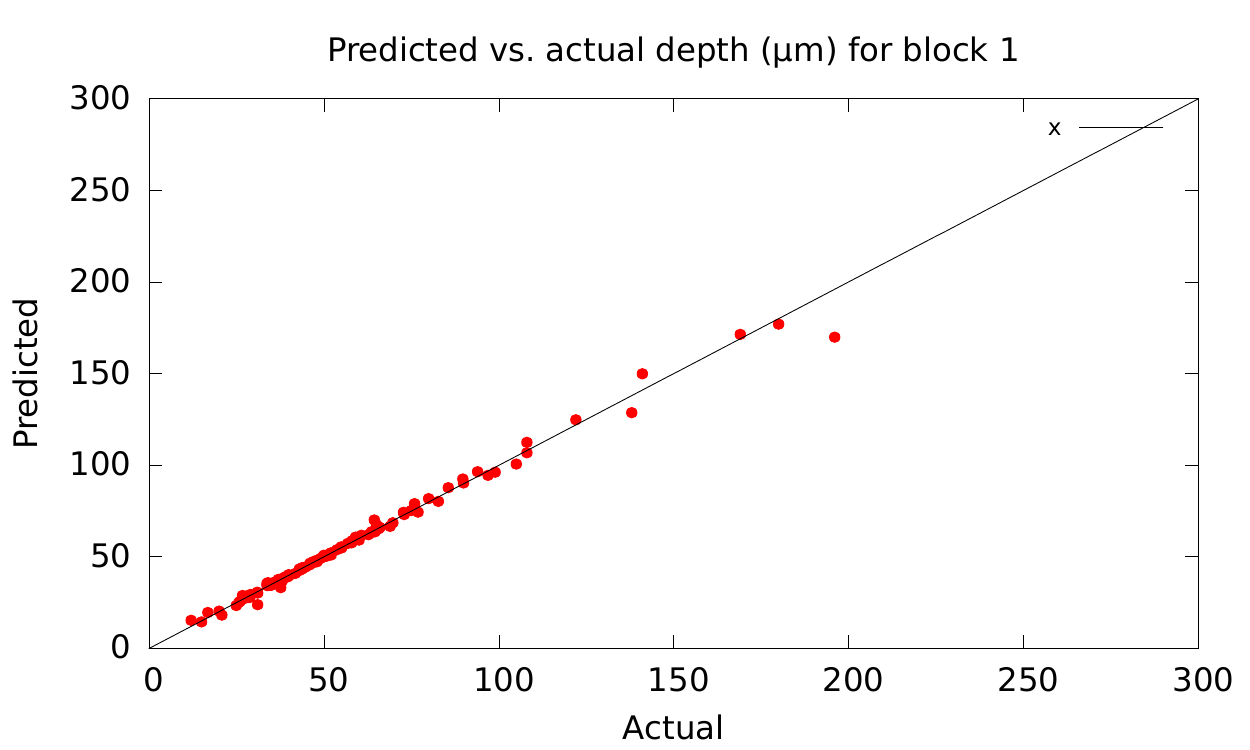} \\
\includegraphics[width=0.23\textwidth]{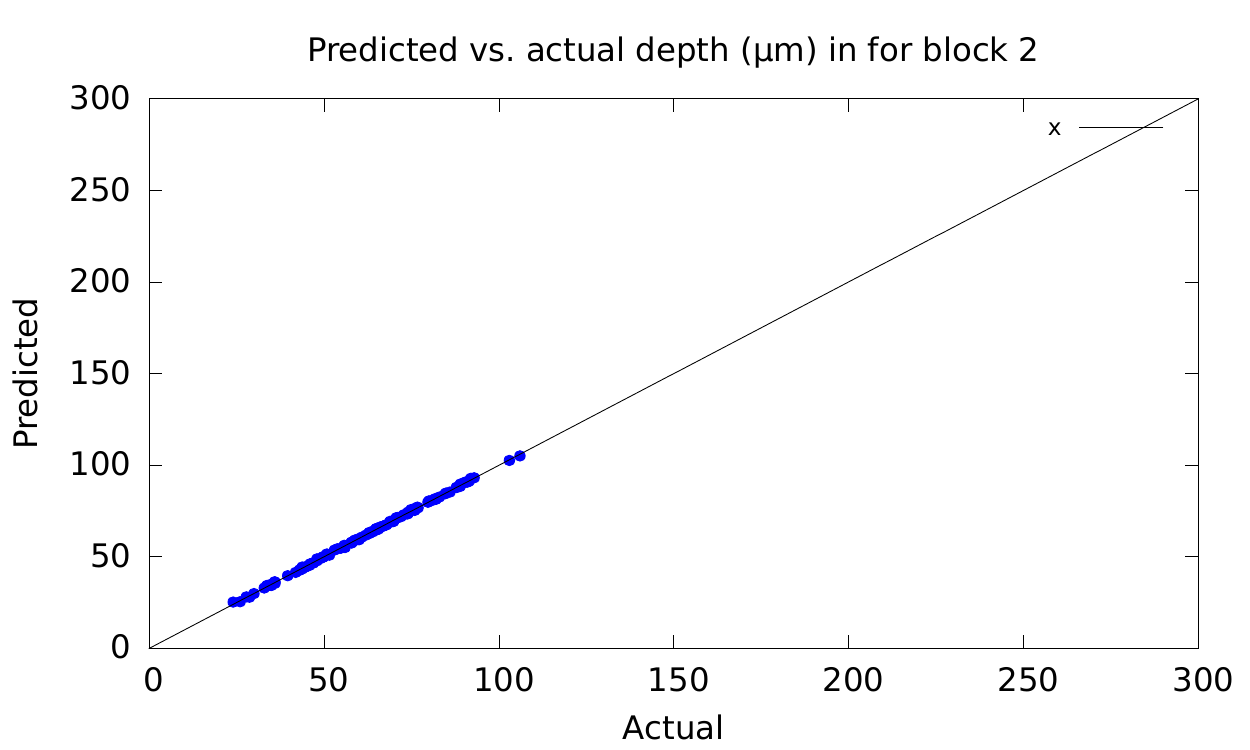} &
\includegraphics[width=0.23\textwidth]{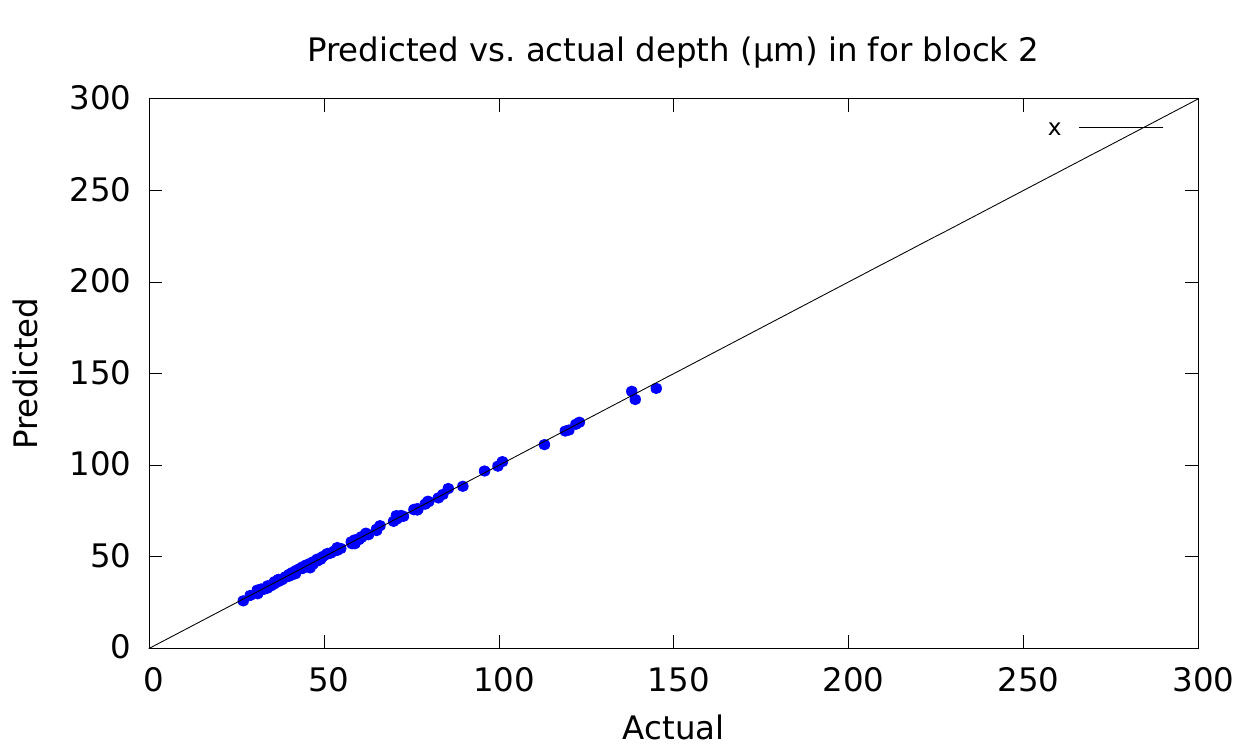} &
\includegraphics[width=0.23\textwidth]{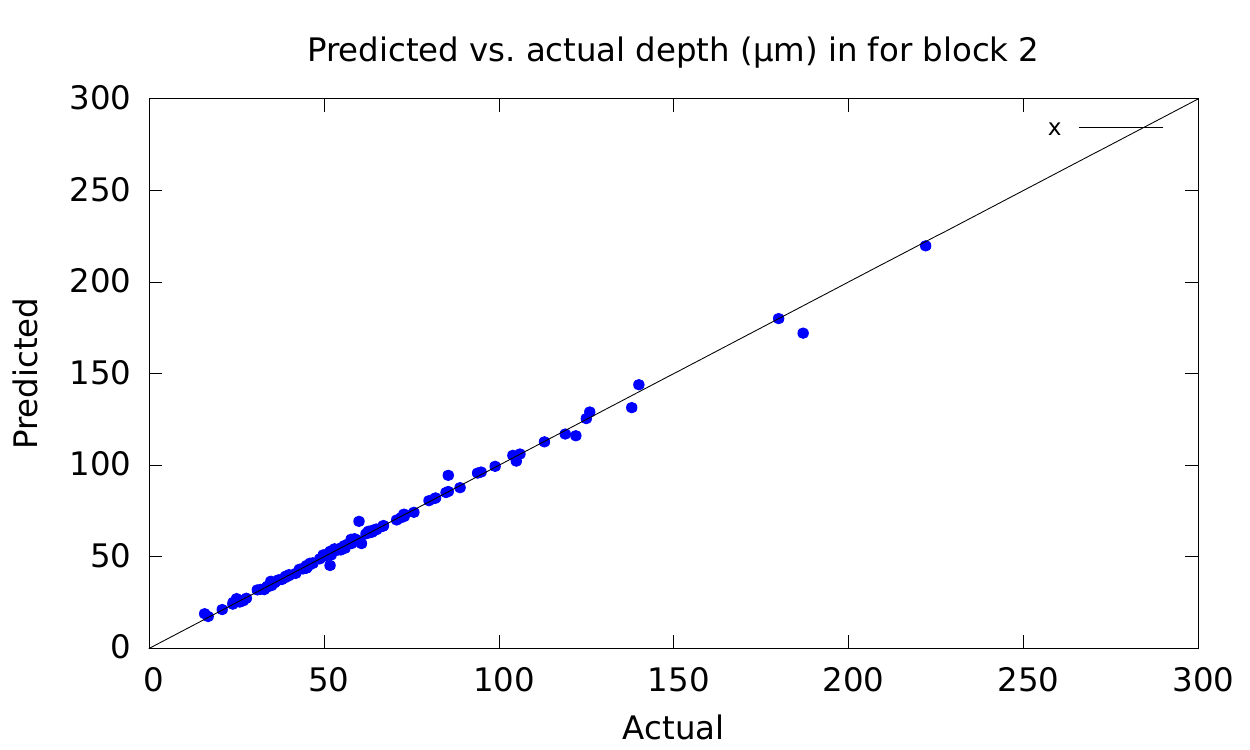} &
\includegraphics[width=0.23\textwidth]{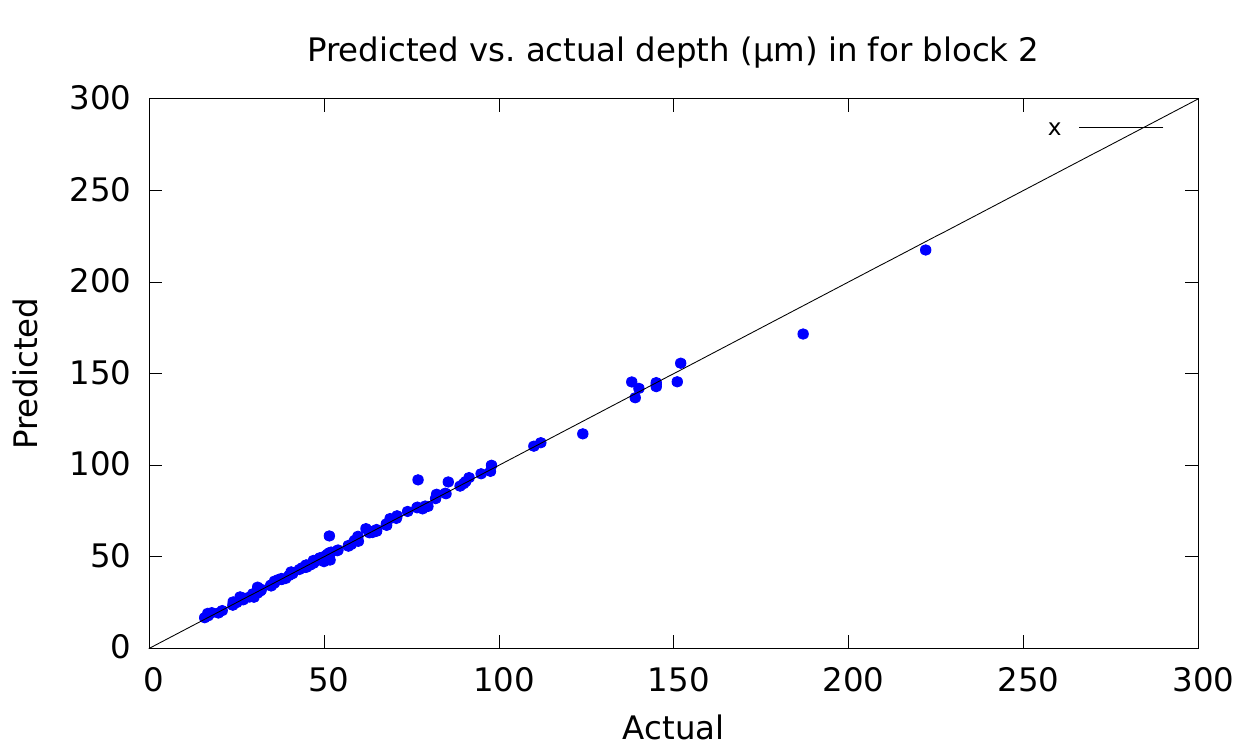} \\
\includegraphics[width=0.23\textwidth]{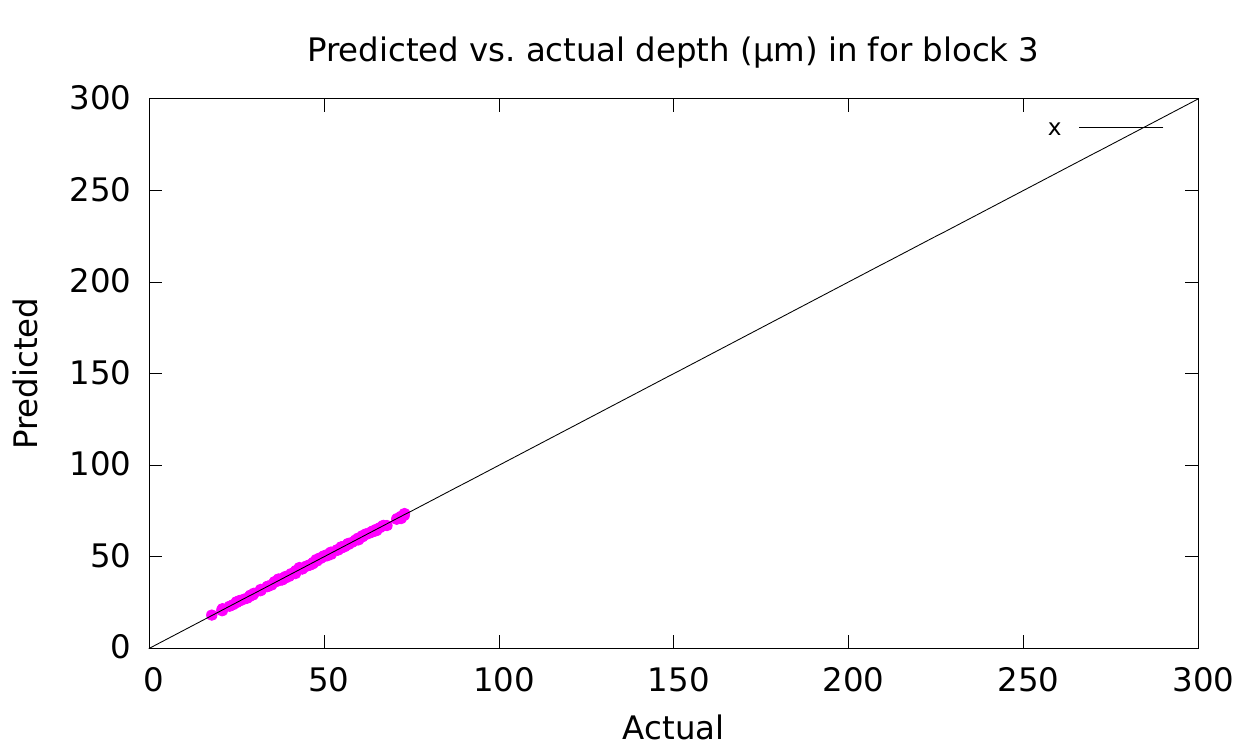} &
\includegraphics[width=0.23\textwidth]{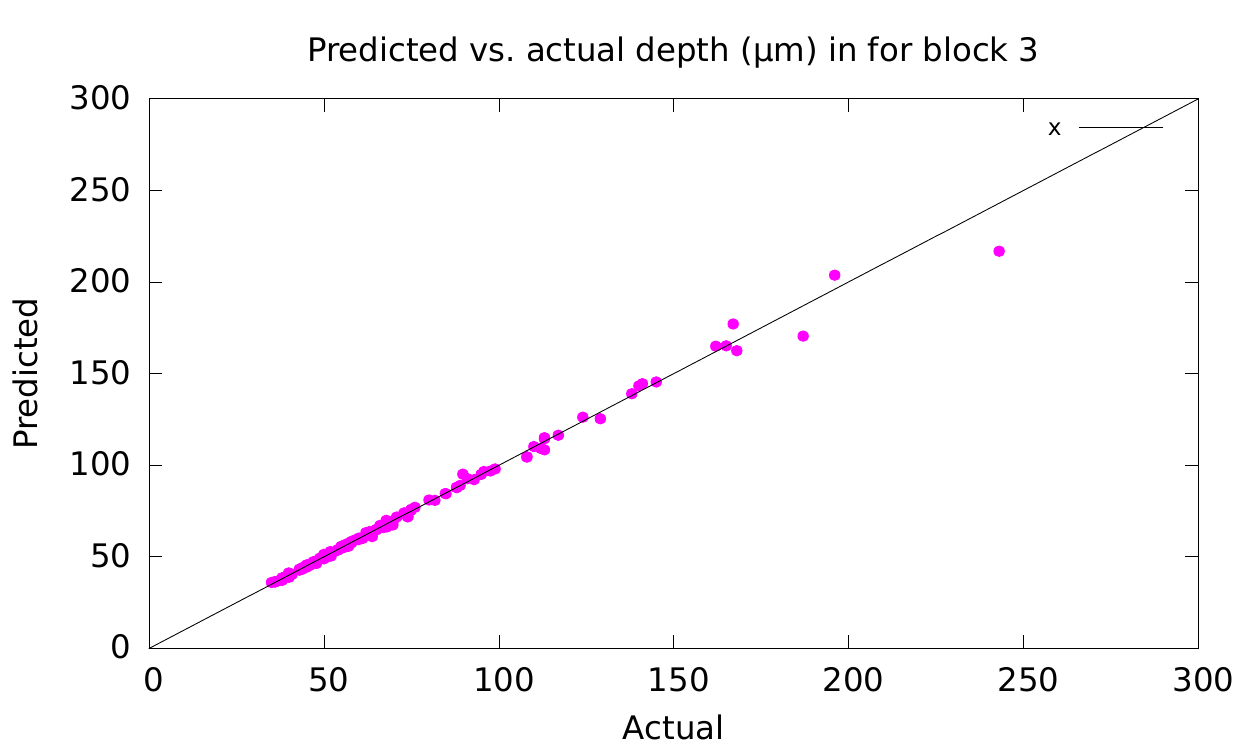} &
\includegraphics[width=0.23\textwidth]{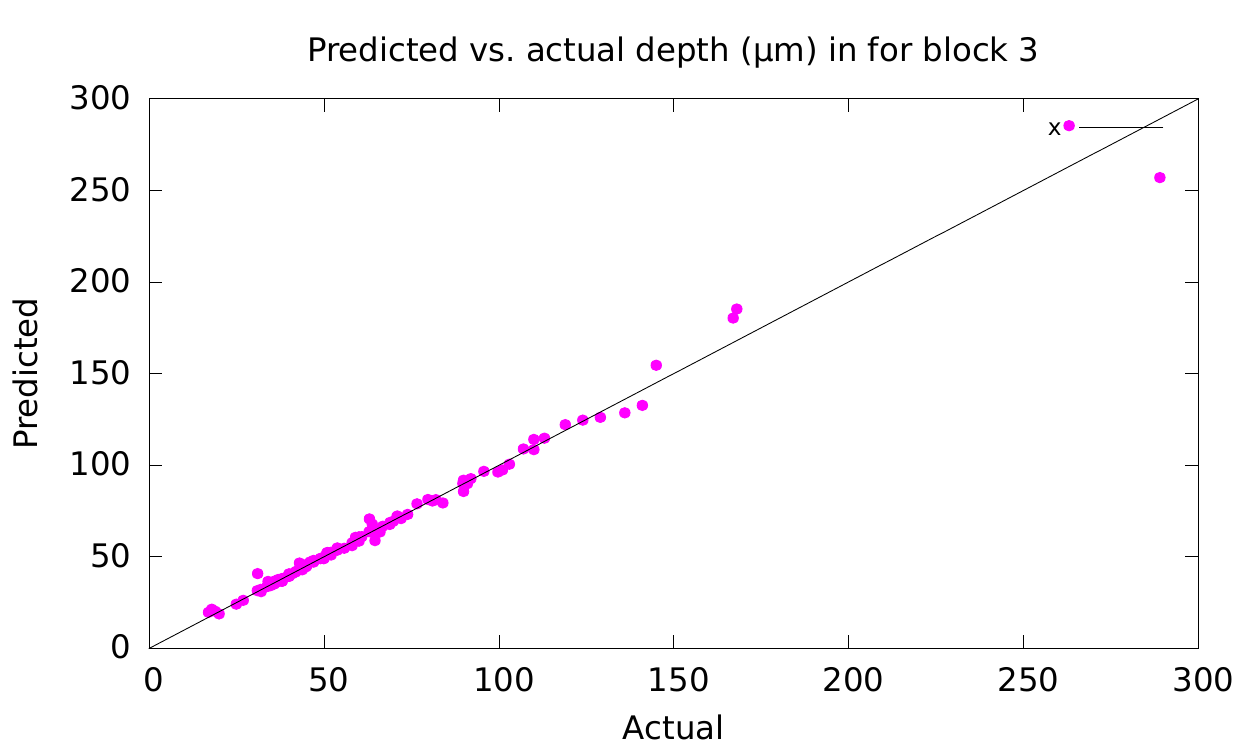} &
\includegraphics[width=0.23\textwidth]{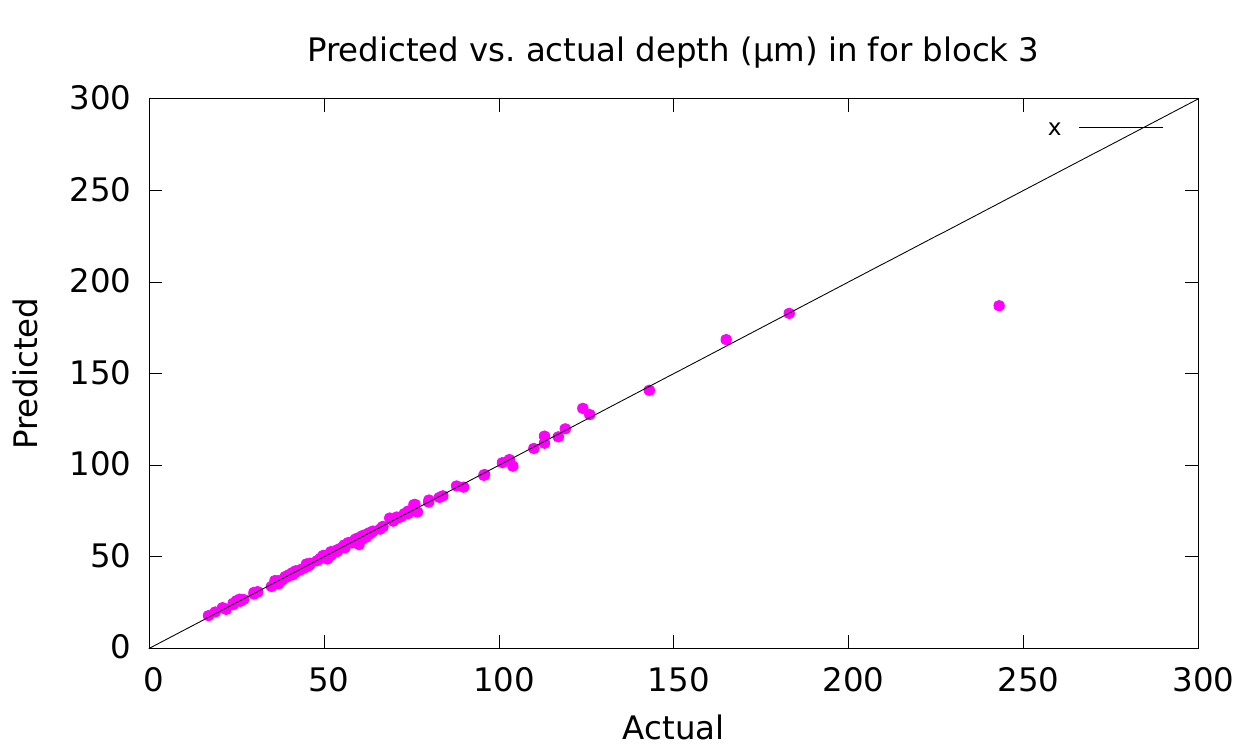} \\
\includegraphics[width=0.23\textwidth]{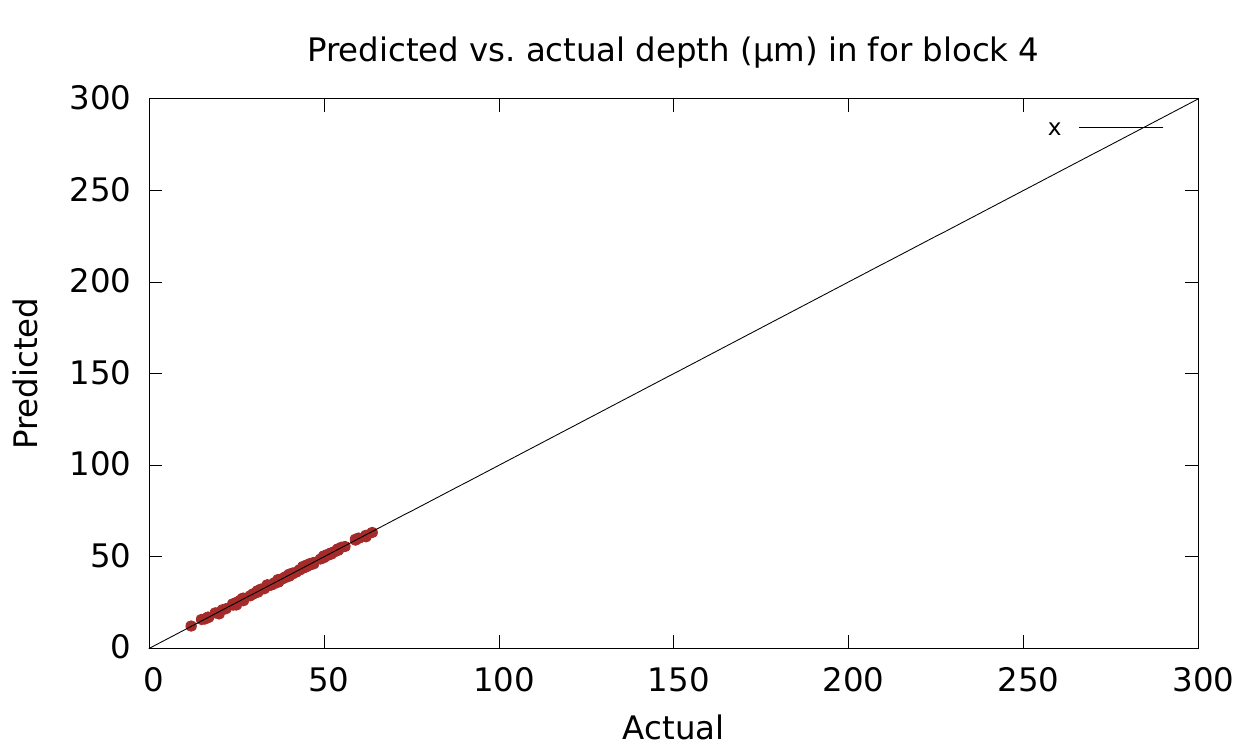} &
\includegraphics[width=0.23\textwidth]{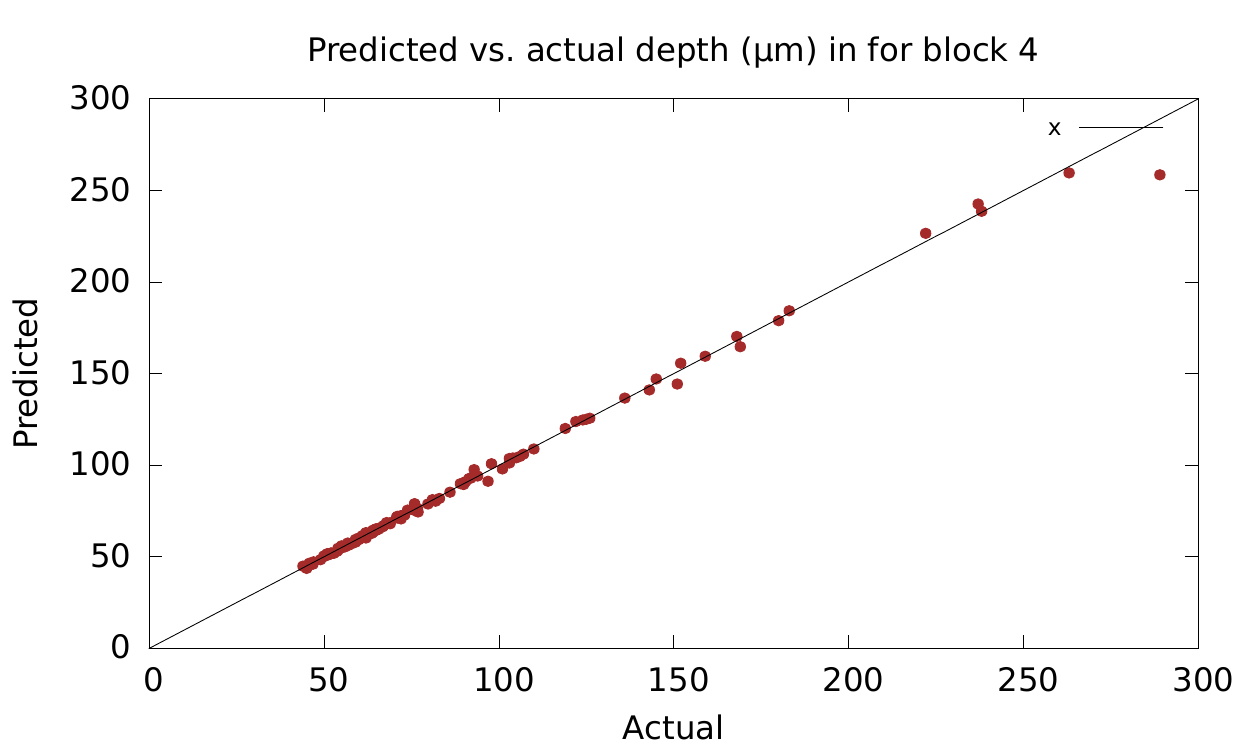} &
\includegraphics[width=0.23\textwidth]{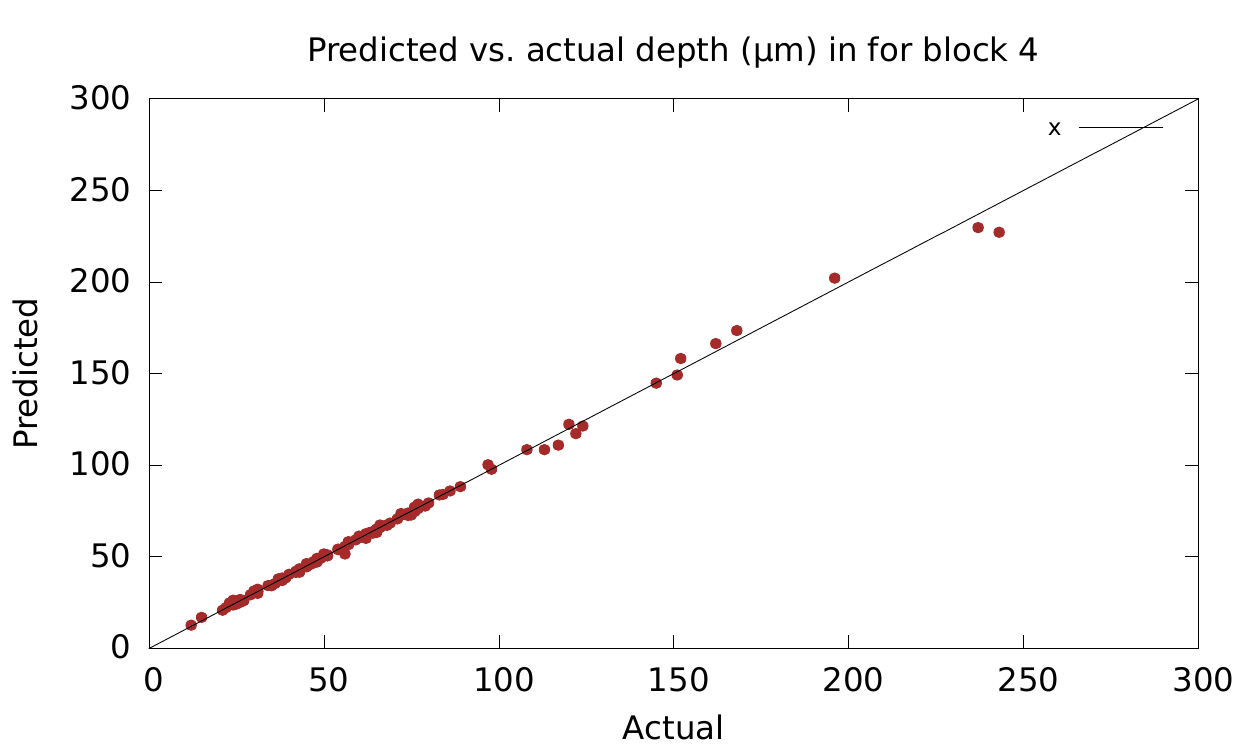} &
\includegraphics[width=0.23\textwidth]{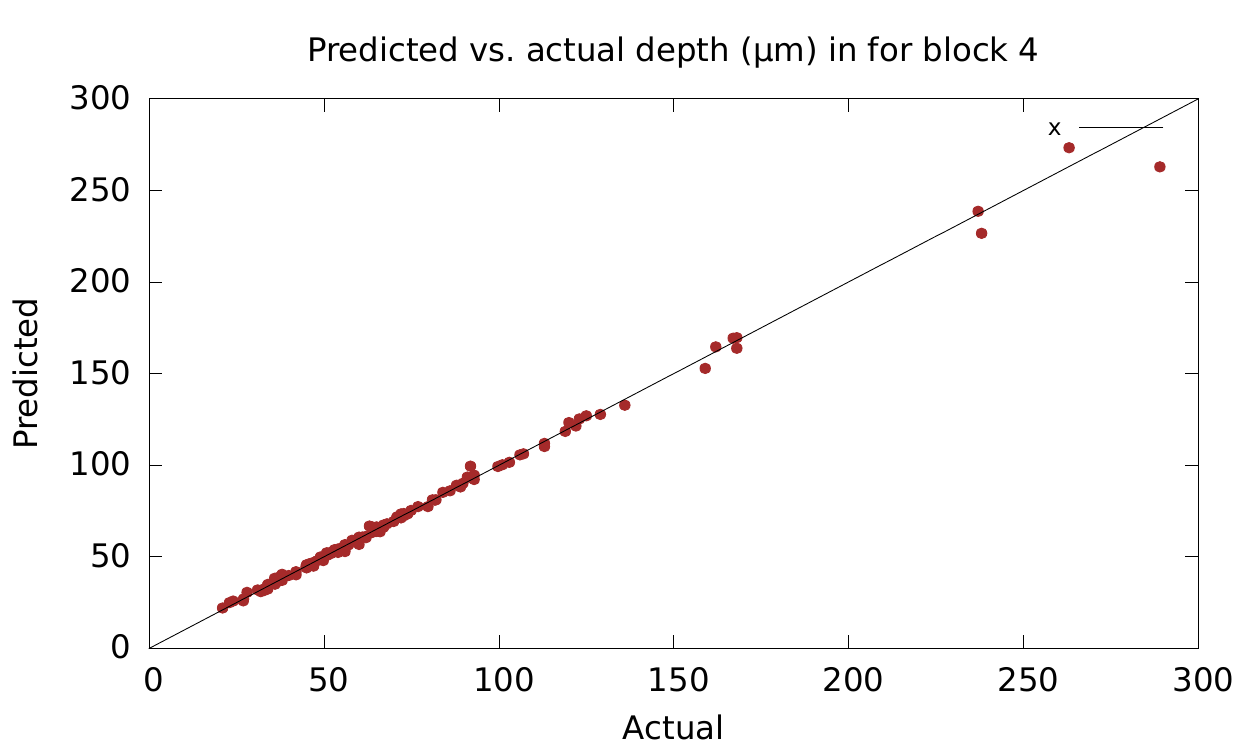} \\
\end{tabular} 
\vspace{0.1cm}
\caption{GP predictions with four blocks along each dimension.  From
  left to right: blocking on the laser speed, power, beam size, and
  absorptivity. Top row: the values of melt-pool depth as a function
  of each of the inputs. Rows two through five: the LOO predicted
  vs. actual values for the depth corresponding to points with
  increasingly larger values of the input (coded by color). }
\label{fig:taper_oned_fourb}
\end{figure*}

\begin{figure*}[!htb]
\centering
\begin{tabular}{cccc}
\multicolumn{2}{c}{\includegraphics[width=0.37\textwidth]{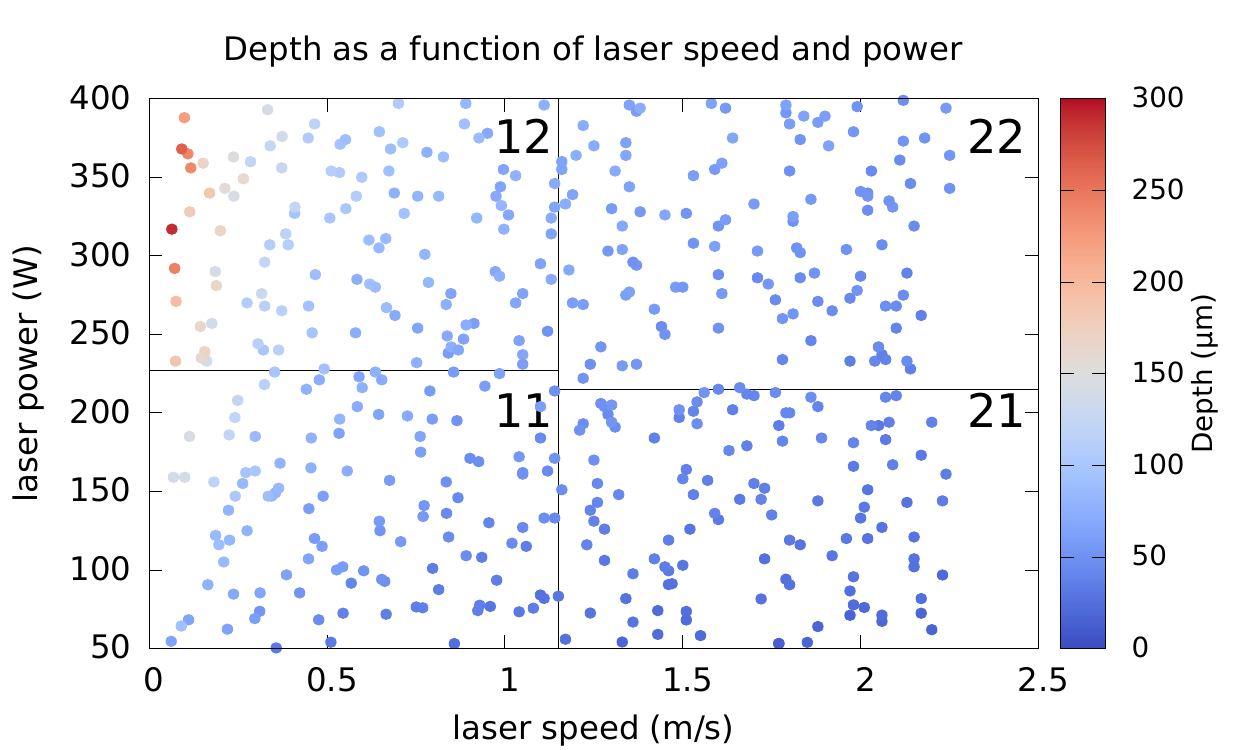}} &
\multicolumn{2}{c}{\includegraphics[width=0.37\textwidth]{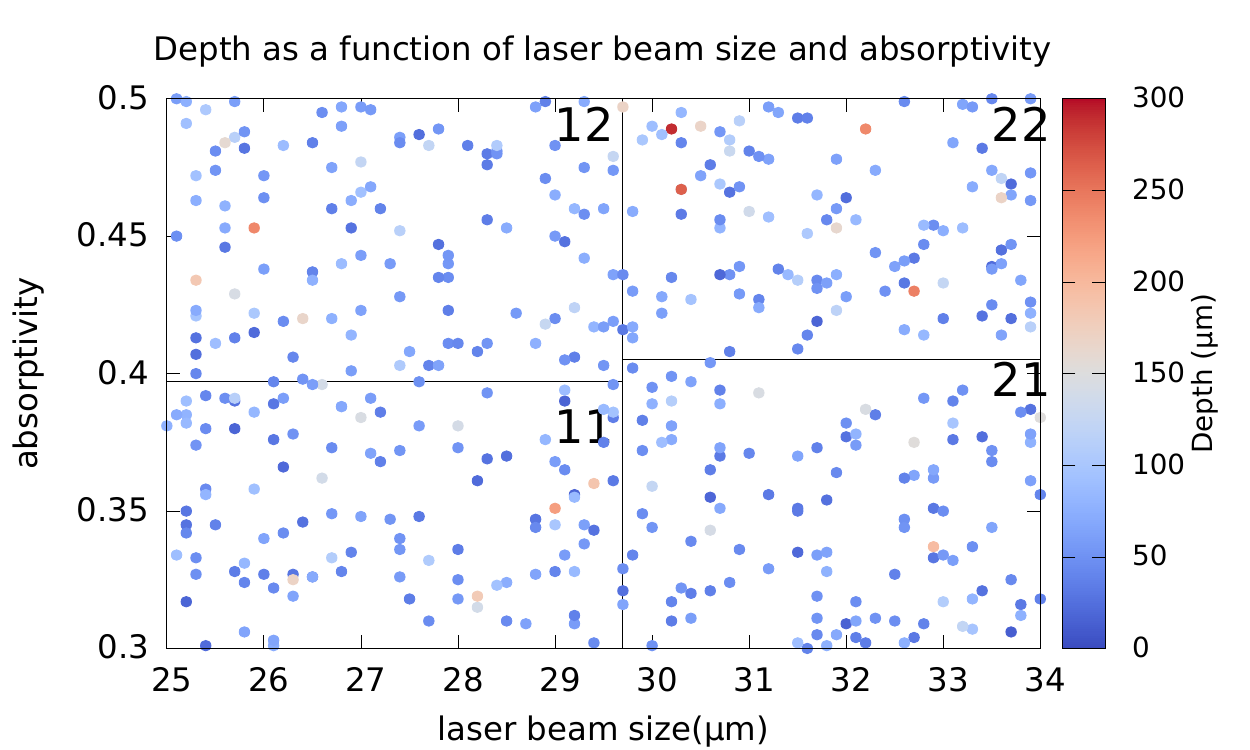}} \\
\multicolumn{2}{c}{(a)} & \multicolumn{2}{c}{(b)} \\
\includegraphics[width=0.23\textwidth]{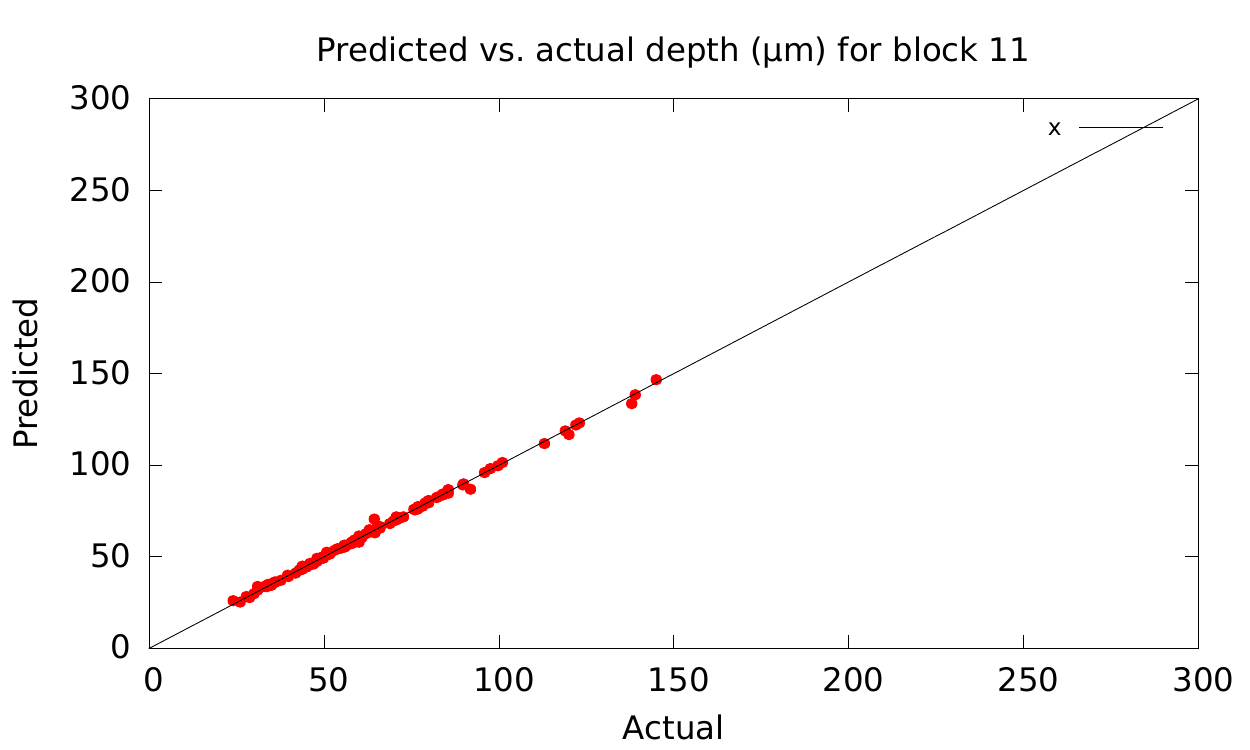} &
\includegraphics[width=0.23\textwidth]{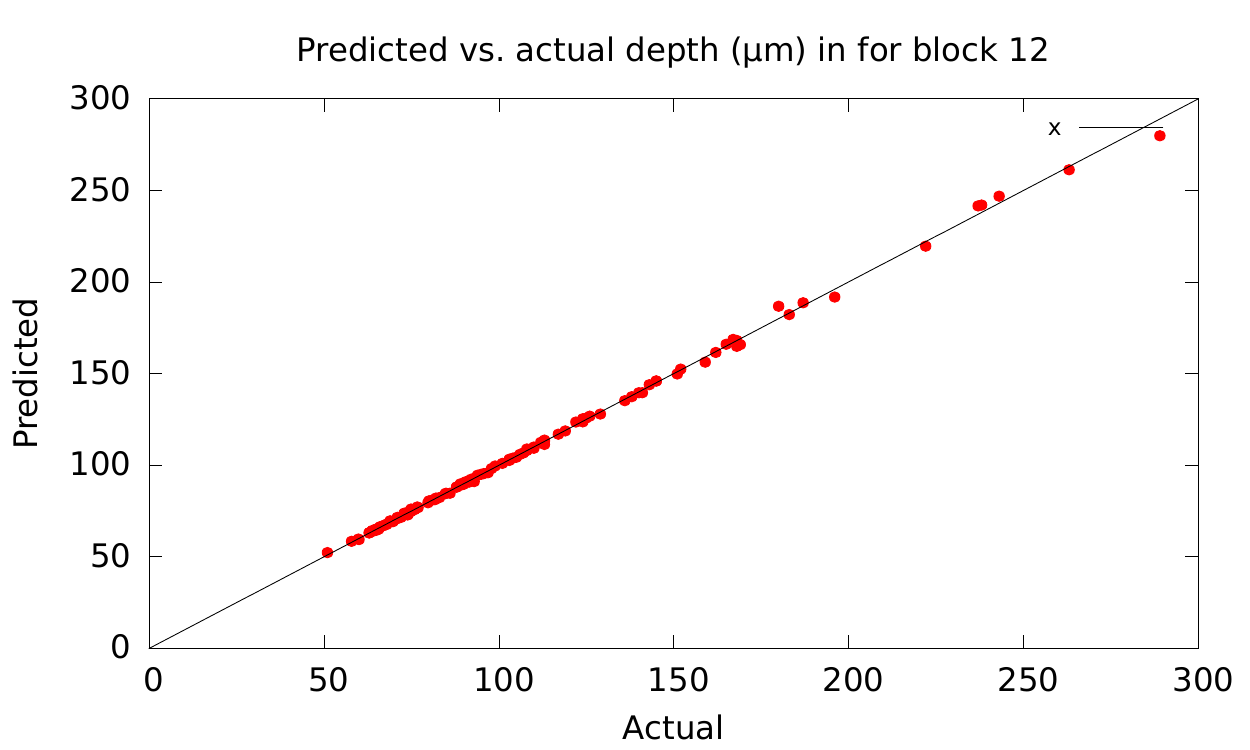} &
\includegraphics[width=0.23\textwidth]{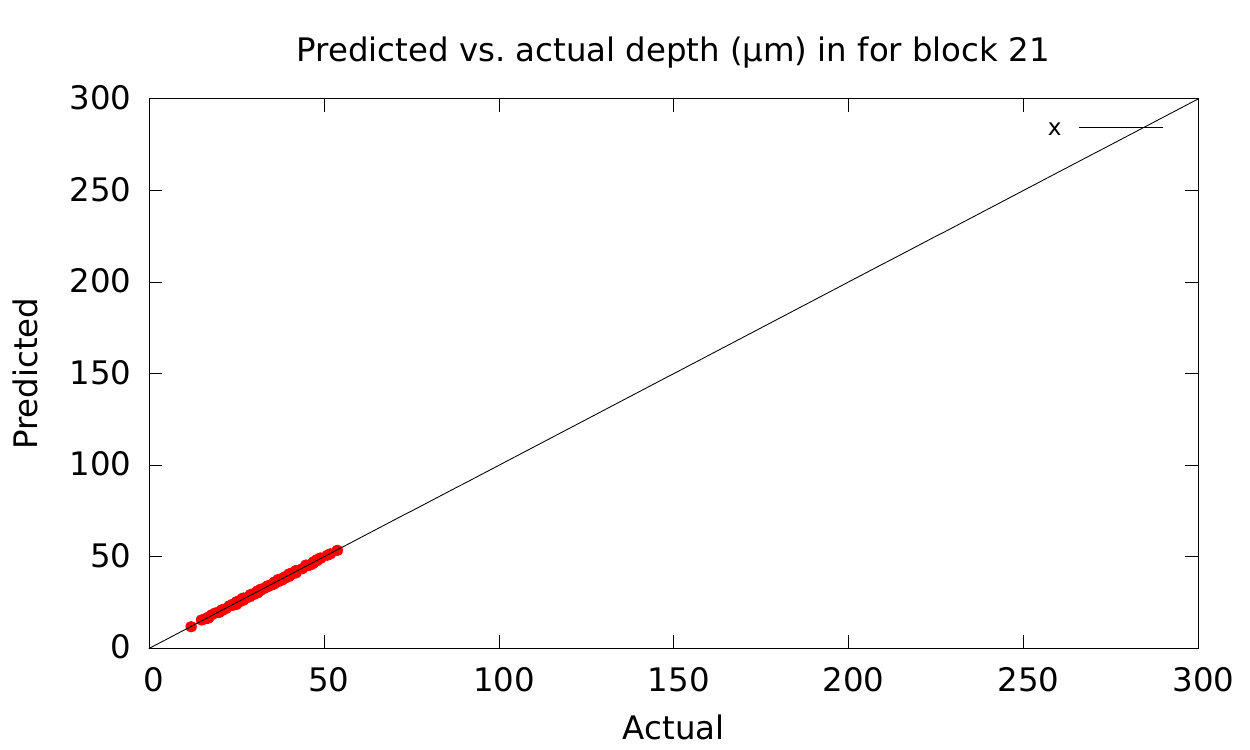} &
\includegraphics[width=0.23\textwidth]{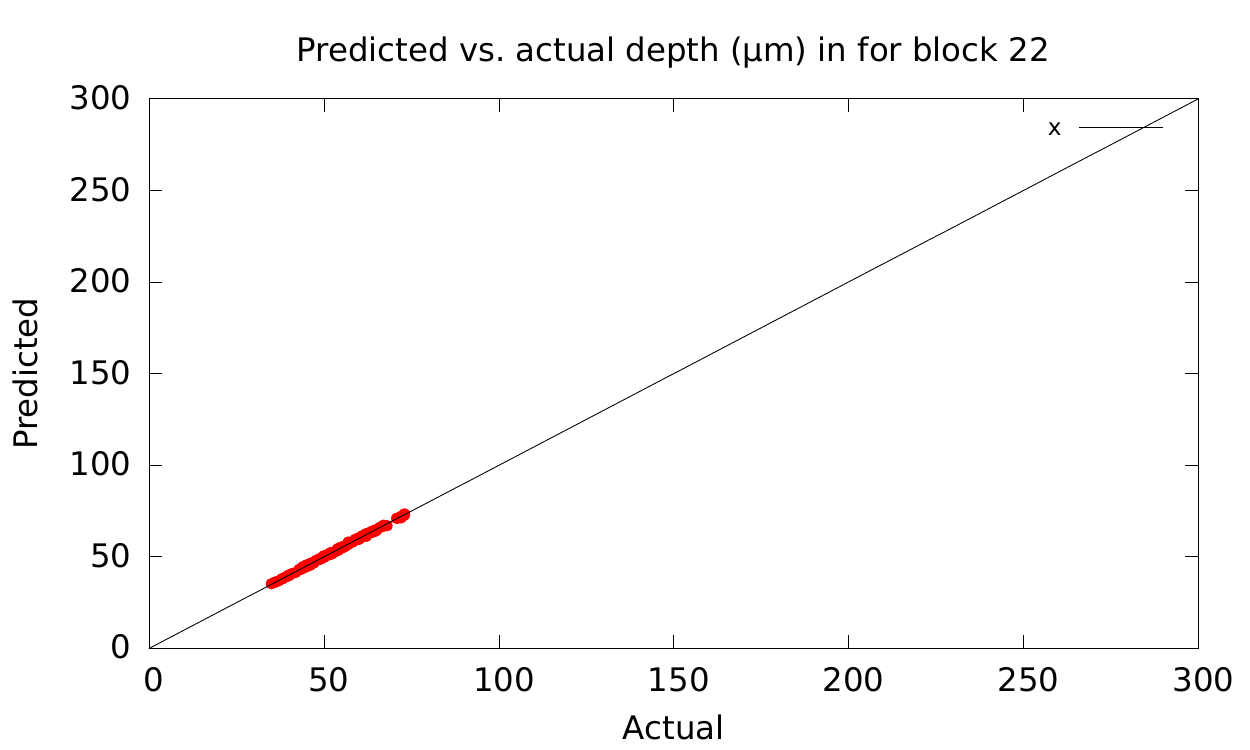} \\
\includegraphics[width=0.23\textwidth]{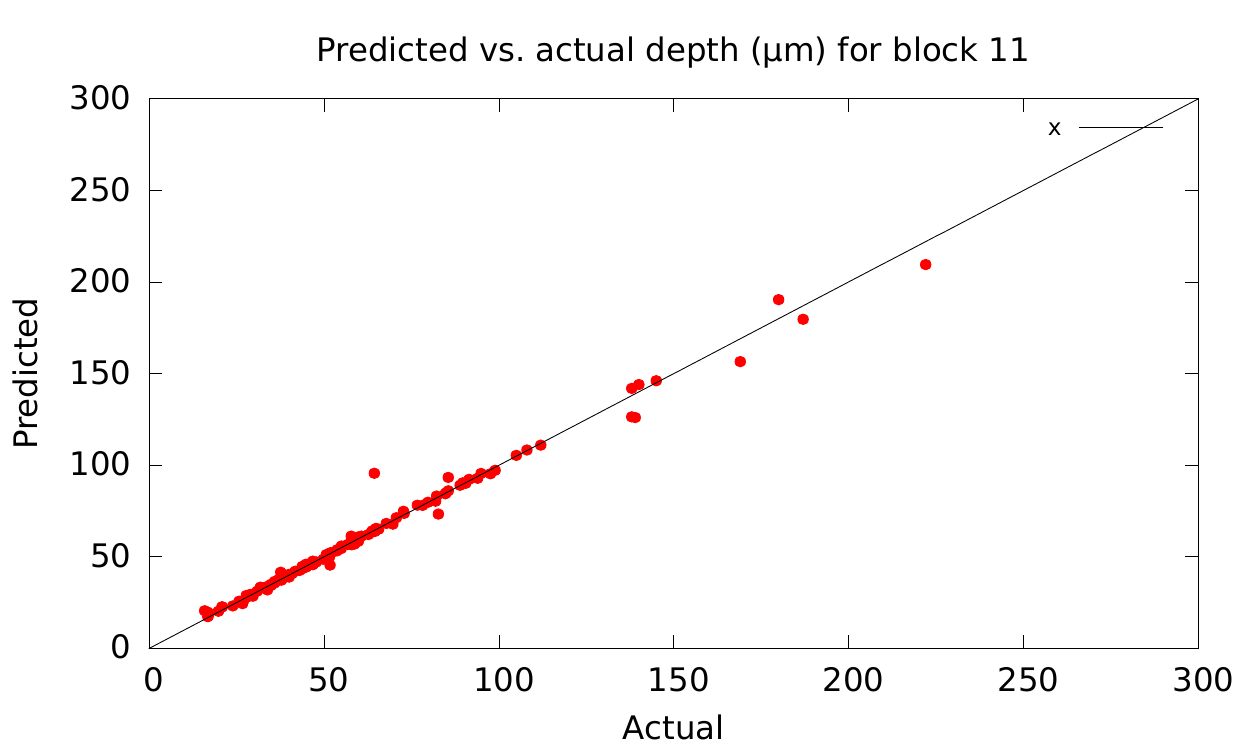} &
\includegraphics[width=0.23\textwidth]{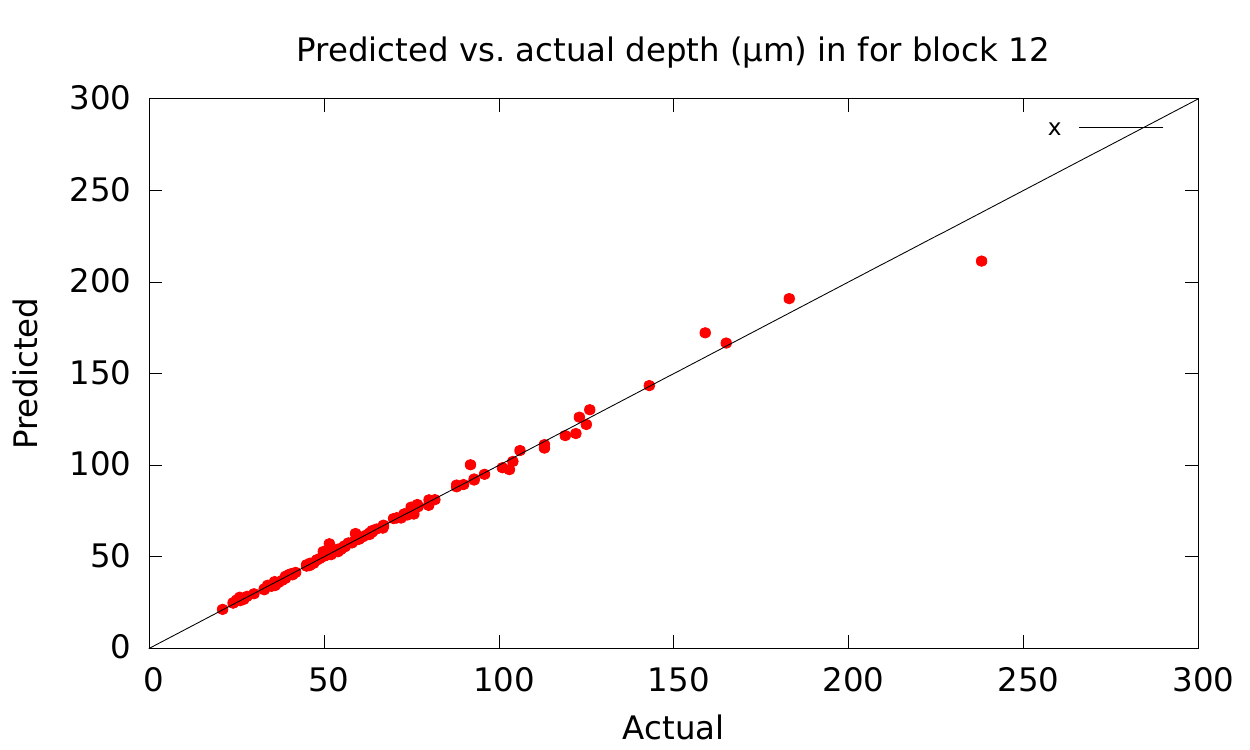} &
\includegraphics[width=0.23\textwidth]{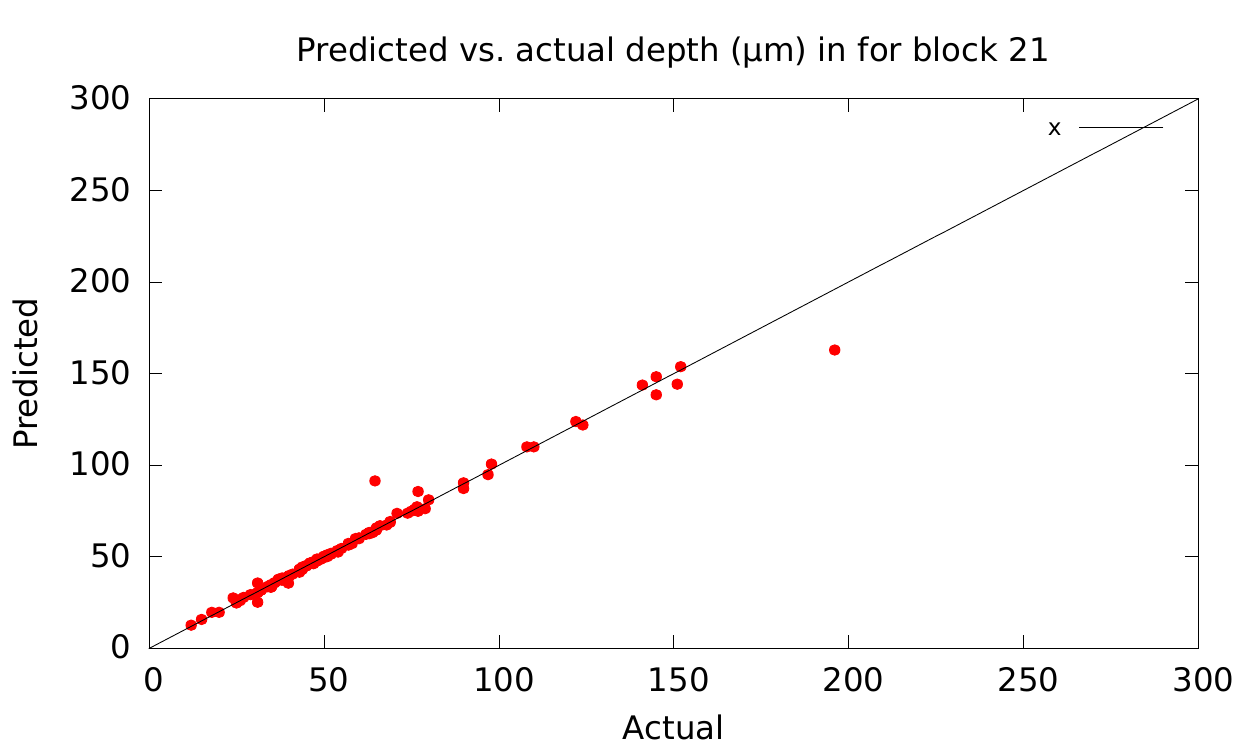} &
\includegraphics[width=0.23\textwidth]{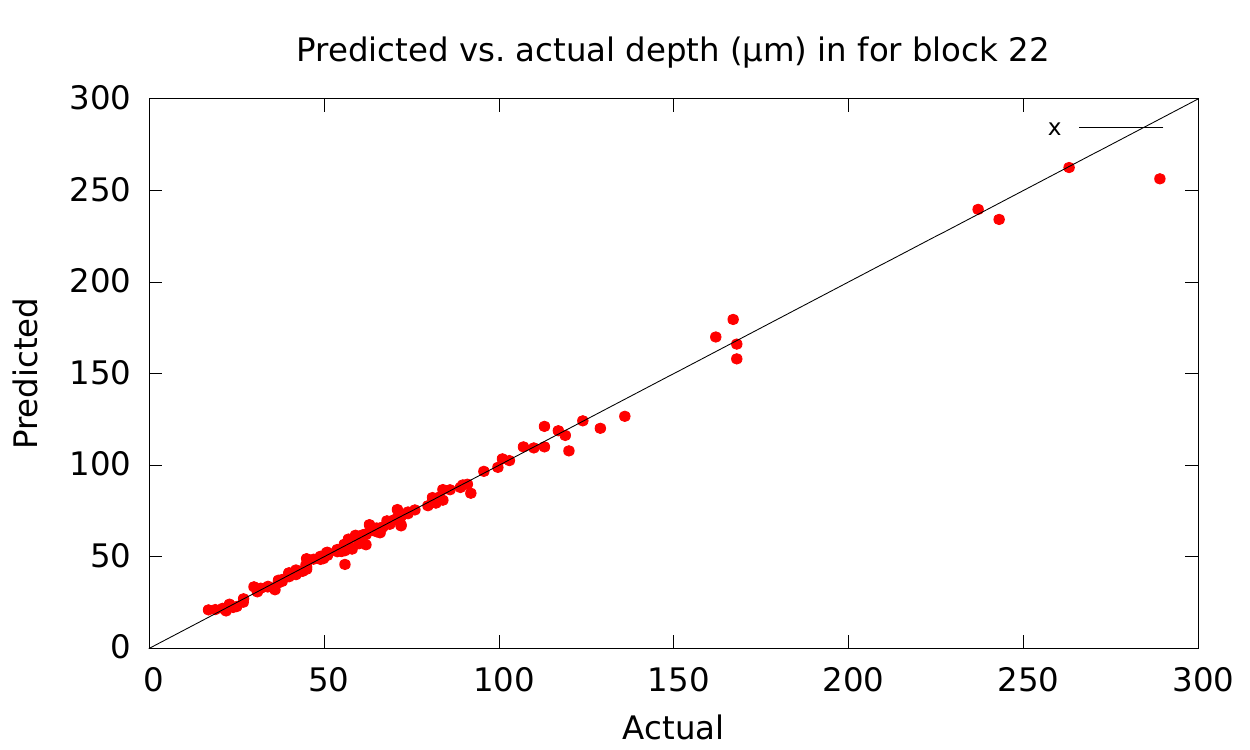} \\
\end{tabular}
\caption{GP predictions with four blocks along two dimensions. (a)
  blocking in the laser speed and laser power inputs. (b) blocking in
  the laser beam size and absorptivity inputs. Each block is numbered
  in its top right corner. Middle and bottom rows correspond to (a)
  and (b), respectively, and show the LOO predicted vs. actual values
  for the depth for each block, from left to right: block 11, 12, 21,
  and 22.}
\label{fig:taper_twod_fourb}
\end{figure*}

First, in Figure~\ref{fig:gp_orig}, we show the LOO accuracy of GP on
the full ET462 data set with the optimal hyperparameters used to
generate these results.  For independent-block GP, we created the blocks by
splitting the points in an input dimension based on their values; this
may result in slightly unequal blocks if there is more than one point
with the same input value.  We then ran the GP on each individual
block and found the optimal set of hyperparameters from the 100
samples.
Figures~\ref{fig:taper_oned_twob},~\ref{fig:taper_oned_fourb},
and~\ref{fig:taper_twod_fourb} present the results with two blocks
along each input dimension, four blocks along each input dimension,
and four blocks split $2 \times 2$ across two input dimensions,
respectively. In each figure, we first show the values of the
melt-pool depth variable as a function of the input variable used in
the split, followed by the performance of GP on each block using the
optimal parameters identified for that block. The corresponding
optimal hyperparameters for each block for the different blocking
schemes are in
Tables~\ref{tab:opt_param_oned_twob},~\ref{tab:opt_param_oned_fourb},
and~\ref{tab:opt_param_twod_fourb}. The tables also include the time
for processing each block, the speedup relative to processing the full
data set, and the mean absolute error in prediction for each block at
the optimal values of the hyperparameters.

\begin{table*}[!htb]
  \begin{center}
    \begin{tabular}{|c |c |c| c| c | c | c | c | c | c |} \hline
      Block & Block & Time & $ \sigma_f$ & $\sigma_n $ & $l/w_1$ & $l/w_2$ & $l/w_3$ & $l/w_4$ & MAE\\
      variable & number & (s) & & & & & & & (opt) \\ 
      \hline
       Speed & 1000 & 226 & 3.855e-05 & 2.348e-07 & 0.185 & 1.107  & 1.781 & 1.973 & 7.08e-07\\
             & 2000 & 226 & 2.385e-05 & 4.143e-07 & 1.271  & 0.393 & 1.568  & 1.966 & 2.89e-07\\
      \hline
       Power & 0100 & 223 & 3.855e-05 & 2.348e-07 & 0.185 & 1.107  & 1.781 & 1.973 & 4.72e-07\\
             & 0200 & 229 & 2.787e-05 & 2.683e-07 & 0.142 & 1.151  & 1.536 & 1.293 & 6.79e-07\\
      \hline
       Beam size & 0010 & 231 & 2.101e-05 & 2.061e-07 & 0.102 & 0.668 & 1.891  & 1.445 & 7.73e-07\\
                 & 0020 & 220 & 3.333e-05 & 2.450e-07 & 0.133 & 0.750 & 1.460 & 1.708 & 9.14e-07  \\
      \hline
       Absorptivity & 0001 & 226 & 3.855e-05 & 2.348e-07 & 0.185 & 1.107  & 1.781 & 1.973 & 7.26e-07\\
                    & 0002 & 226 & 3.333e-05 & 2.450e-07 & 0.133 & 0.750 & 1.460 & 1.708 & 9.79e-07 \\
      \hline
    \end{tabular} 
    \vspace{-0.2cm}
    \caption{Optimum hyperparameters, with mean absolute error (MAE),
      for blocking along one dimension with two blocks. The block
      number has four digits in order of input variables. A value of 1
      indicates lower values and 2 indicates higher values of the
      input; 0 indicates no blocking. Thus block 0100 has lower
      values of the second input parameter, the laser power. Each
      block has roughly (462/2 = 231) sample points. The speedup
      compared to no blocking is 4.38 - 4.39.}
  \label{tab:opt_param_oned_twob}
  \end{center}
\end{table*}

\begin{table*}[htb]
  \begin{center}
    \begin{tabular}{| c | c |c| c| c | c | c | c | c | c |} \hline
      Block & Block & Time & $ \sigma_f$ & $\sigma_n $ & $l/w_1$ & $l/w_2$ & $l/w_3$ & $l/w_4$ & MAE \\
      variable & number & (s) & & & & & & & (opt) \\ 
      \hline
       Speed & 1000 & 27 & 5.300e-05 & 4.295e-07 & 0.282 & 0.612  & 1.248 & 1.902 & 9.51e-07 \\
             & 2000 & 28 & 4.771e-05 & 3.842e-07 & 1.867  & 0.789   & 1.974 & 1.803 & 3.46e-07\\
             & 3000 & 29 & 3.055e-05 & 4.341e-07 & 1.860  & 0.320 & 1.680 & 1.847 & 3.11e-07 \\
             & 4000 & 28 & 5.584e-05 & 4.068e-07 & 1.895  & 0.572 & 1.281 & 1.731 & 3.08e-07\\
      \hline
       Power & 0100 & 28 & 3.314e-05 & 2.908e-07 & 0.247 & 1.543  & 1.941 & 1.601 & 5.11e-07\\
             & 0200 & 27 & 3.855e-05 & 2.348e-07 & 0.185 & 1.107  & 1.781 & 1.973 & 5.54e-07 \\
             & 0300 & 29 & 2.787e-05 & 2.683e-07 & 0.142 & 1.151  & 1.536 & 1.293 & 1.47e-06 \\
             & 0400 & 27 & 2.795e-05 & 2.292e-07 & 0.171 & 1.247  & 1.115 & 1.701 & 1.30e-06\\
      \hline
       Beam size & 0010 & 27 & 3.855e-05 & 2.348e-07 & 0.185 & 1.107  & 1.781 & 1.973 & 1.49e-06\\
                 & 0020 & 29 & 3.152e-05 & 1.571e-07 & 0.253 & 1.221  & 1.449 & 1.961 & 1.10e-06\\
                 & 0030 & 24 & 3.855e-05 & 2.348e-07 & 0.185 & 1.107  & 1.781 & 1.973 & 2.23e-06\\
                 & 0040 & 30 & 3.385e-05 & 3.018e-07  & 0.156  & 0.749 & 1.167 & 1.969 & 1.21e-06\\
      \hline
       Absorptivity & 0001 & 27 & 3.806e-05 & 1.277e-07 & 0.261  & 0.823 & 1.888 & 1.617 & 1.40e-06\\
                    & 0002 & 28 & 3.855e-05 & 2.348e-07 & 0.185 & 1.107  & 1.781 & 1.973 & 1.37e-06\\
                    & 0003 & 26 & 3.855e-05 & 2.348e-07 & 0.185 & 1.107  & 1.781 & 1.973 & 1.31e-06\\
                    & 0004 & 29 & 3.333e-05 & 2.450e-07 & 0.133 & 0.750 & 1.460 & 1.708 & 1.61e-06\\
      \hline
    \end{tabular}
    \caption{Optimum hyperparameters, with mean absolute error (MAE),
      for blocking along one dimension with four blocks. The block
      number has four digits in order of input variables.  Values of
      1,2,3, and 4 indicate low to high input values; 0 indicates no
      blocking. Each block has roughly (462/4 = 115) sample
      points. The speedup compared to no blocking is 17.67 - 17.99.}
  \label{tab:opt_param_oned_fourb}
  \end{center}
\end{table*}

\begin{table*}[!htb]
  \begin{center}
    \begin{tabular}{| c | c |c| c| c | c | c | c | c | c |} \hline
      Block & Block & Time & $ \sigma_f$ & $\sigma_n $ & $l/w_1$ & $l/w_2$ & $l/w_3$ & $l/w_4$ & MAE \\
      variables & number & (s) & & & & & & & (opt) \\ 
      \hline
       Speed & 1100 & 27 & 3.855e-05 & 2.348e-07 & 0.185 & 1.107  & 1.781 & 1.973 & 6.82e-07\\
       and   & 1200 & 28 & 3.855e-05 & 2.348e-07 & 0.185 & 1.107  & 1.781 & 1.973 & 8.54e-07\\
       power & 2100 & 27 & 4.209e-05 & 5.392e-07 & 1.751 & 1.044  & 1.953 & 1.961 & 2.90e-07\\
             &2200 & 28 & 3.746e-05 & 5.059e-07 & 0.864 & 1.251  & 1.316 & 1.966 & 2.86e-07\\
      \hline
       Beam size    & 0011 & 27 & 3.855e-05 & 2.348e-07 & 0.185 & 1.107  & 1.781 & 1.973 & 1.88e-06\\
       and          & 0012 & 29 & 3.152e-05 & 1.571e-07 & 0.253 & 1.221  & 1.449 & 1.961 & 1.39e-06\\
       absorptivity & 0021 & 27 & 3.855e-05 & 2.348e-07 & 0.185 & 1.107  & 1.781 & 1.973 & 1.48e-06\\
                    & 0022 & 27 & 4.361e-05 & 3.459e-07 & 0.091 & 0.292  & 1.558 & 1.654 & 2.47e-06\\
      \hline
    \end{tabular}
  \vspace{-0.2cm}
  \caption{Blocking on two dimensions, four blocks: optimum parameters
    and mean absolute error (MAE) for each block. The four digits in
    each block number indicate the order of input variables. Values of
    1 indicate lower input values and 2 indicates higher values of the
    input; 0 indicates no blocking on that input. Each block has
    roughly (462/4 = 115) sample points. The speedup compared to no
    blocking is 17.99.}
  \label{tab:opt_param_twod_fourb}
  \end{center}
\end{table*}

\pagebreak

We next make several observations on these plots and tables:
\begin{itemize} 

\item {\bf Accuracy of predictions with independent-block GP:} Overall, we
  find that the accuracy of the predictions of melt-pool depth with
  blocking is very good, regardless of the number of blocks or the
  dimensions along which the blocks are created. As we move from two
  blocks to four blocks, we see increased scatter around the $y = x$
  line that denotes exact prediction. This behavior is also confirmed by
  the values of mean absolute error in the tables. A careful study
  indicated that there is pattern to this increased error.  We know
  that the depth values are large at very low values of speed and high
  values of power. If many, or all, of these large depth instances are in
  one block, the prediction accuracy is good, which is the case, for
  example, when we split the speed dimension into four blocks.  But,
  if the blocking distributes the instances with large depth across the
  blocks, there are too few such instances in each block and the
  prediction accuracy suffers. This suggests that we should block
  along the dimension(s) that allow us to keep instances with similar
  output values within a block.

\item {\bf Optimal hyperparameter values for each block:} In light of
  the preceding obervation, we would expect that the hyperparameters
  for the different blocks would likely be different from that of the
  full data set. The hyperparameter, $l/w_i$, indicates the
  importance of the $i$-th dimension, with a smaller value indicating
  greater importance. We also know from our previous
  work~\cite{kamath14:density,kamath16:statinf} that for the melt-pool
  depth for the full data set, the laser speed is a more important
  input than laser power, and both beam size and absorptivity are less
  important. This is confirmed by the hyperparameters in
  Figure~\ref{fig:gp_orig}. But, when we block on the laser speed
  input, its importance reduces for blocks with little variation in
  the speed, as seen in the second row of
  Table~\ref{tab:opt_param_oned_twob} and the second through fourth
  rows in Table~\ref{tab:opt_param_oned_fourb}. This would suggest
  that implementing blocking by dividing the data into separate GP
  models, as we have done, is a better option than creating a single
  model with a block diagonal covariance matrix, as it allows the
  hyperparameters to adapt to the characteristics of the data in each
  block, resulting in more accurate predictions.

\item {\bf Reduction in compute time:} We expect that blocking would
  reduce compute time for hyperparameter optimization by a factor of
  $B^2$.  In fact, the speedup values in the captions of the tables
  show that we obtain a better than $B^2$ speedup; this is the result
  of improved cache performance as blocking also reduces the size of
  the covariance matrices, leading to less communication within the
  memory system. Thus, for even for small values of $B$, such as 2 and
  4, we obtain a speedup of 4.4 and nearly 18,
  respectively. Further speedup can be obtained by observing that the
  $B$ blocks can be processed in parallel, without incurring any extra
  overhead. This would result in a greater than $B^3$ speedup overall.

\item {\bf Number of blocks to create:} While it is tempting to
  increase $B$ to reduce the computational time, this would reduce the
  size of each block. If the blocking is along one dimension, each block
  would be ``skinny'', covering a small range of values in that
  dimension, which would give lower accuracy as most instances would
  be near an internal block boundary. This would suggest blocking along
  more than one dimension if we want more blocks. 

\item {\bf Number of dimensions along which to block the data:} For
  our data set, the results for blocking along two dimensions were no
  worse than blocking along a single dimension. However, it is
  possible that for some data sets, blocking along two, or more,
  dimensions could result in lower prediction accuracy at instances
  near the corners of the blocks. This, and similar effects at
  instances near internal block boundaries, could be resolved by using
  overlapping blocks for building the models; the output at a new
  instance would then be predicted using the model for the block to
  which the instance would be assigned in the absence of overlap.

\end{itemize}

Overall, we conclude that the reduction in compute time makes
independent-block GP a very useful idea, especially if we can choose
the blocks to minimize the overall scatter across the output values in
the blocks.

%
\subsection{Evaluating iterative solvers}
\label{sec:results_iterative}
%

\begin{figure*}[!tb]
\centering
\begin{tabular}{cccc}
\includegraphics[width=0.23\textwidth]{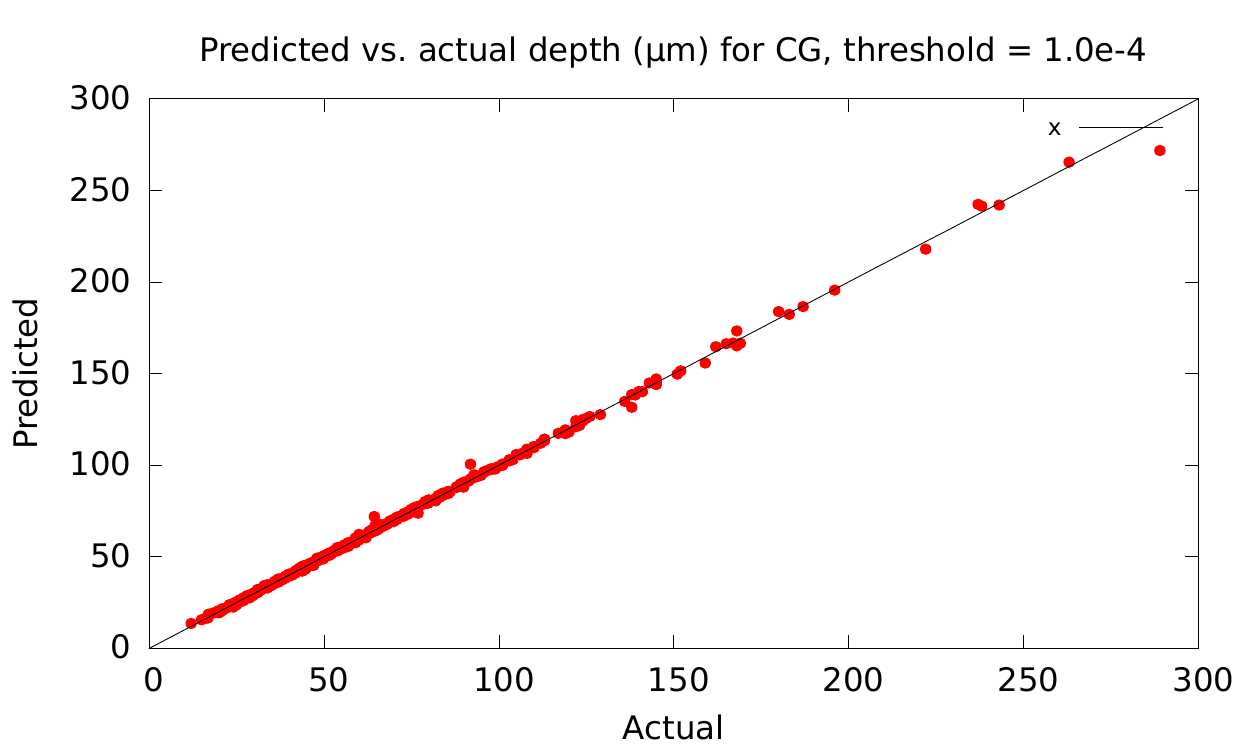} &
\includegraphics[width=0.23\textwidth]{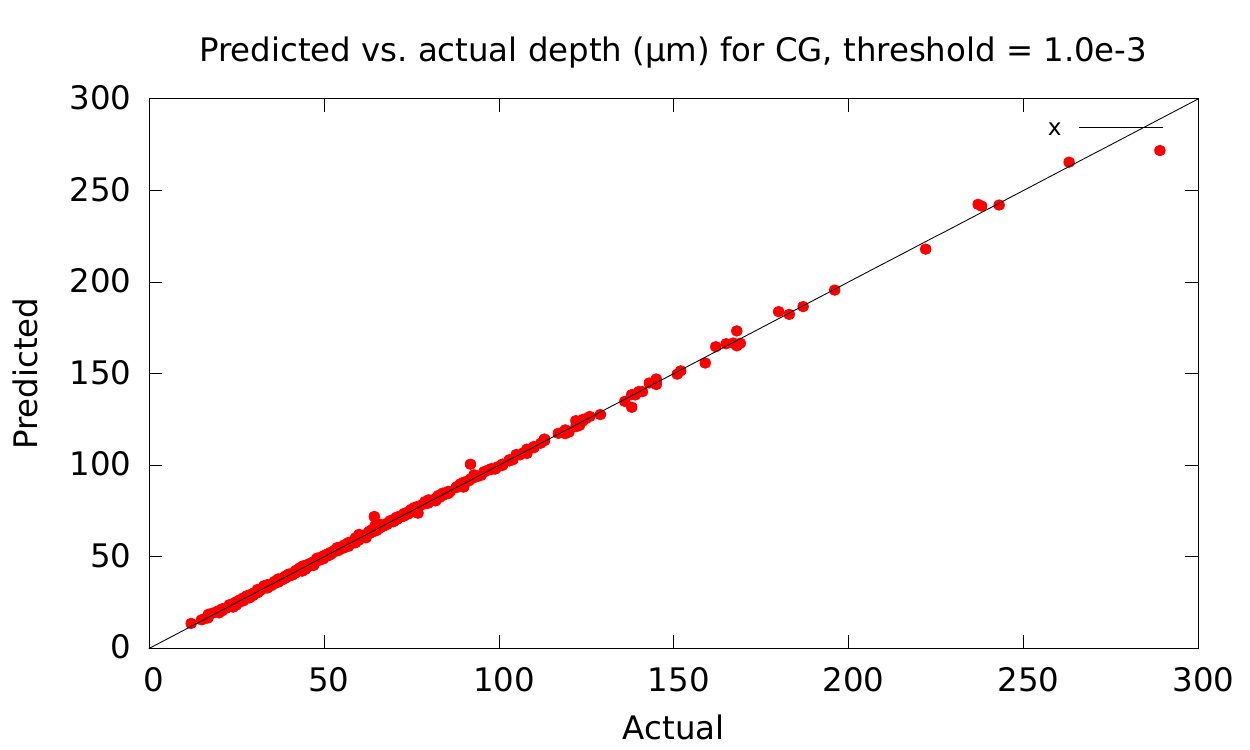} &
\includegraphics[width=0.23\textwidth]{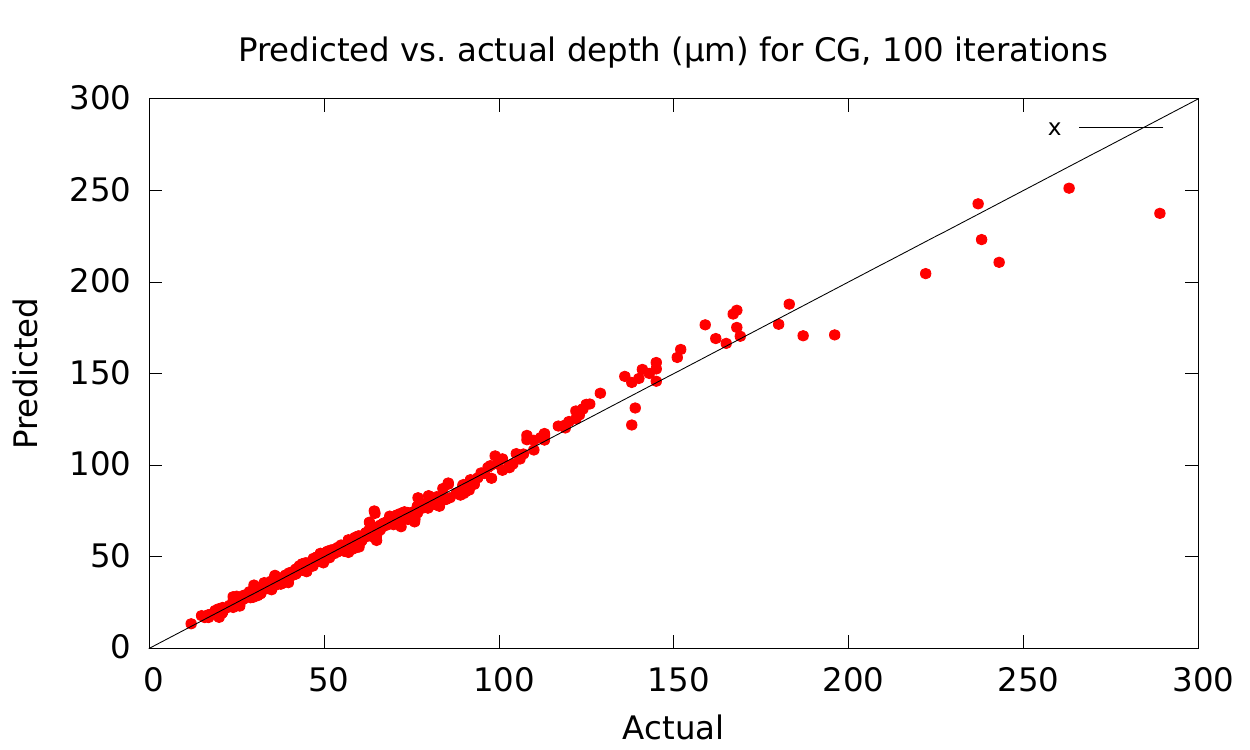} &
\includegraphics[width=0.23\textwidth]{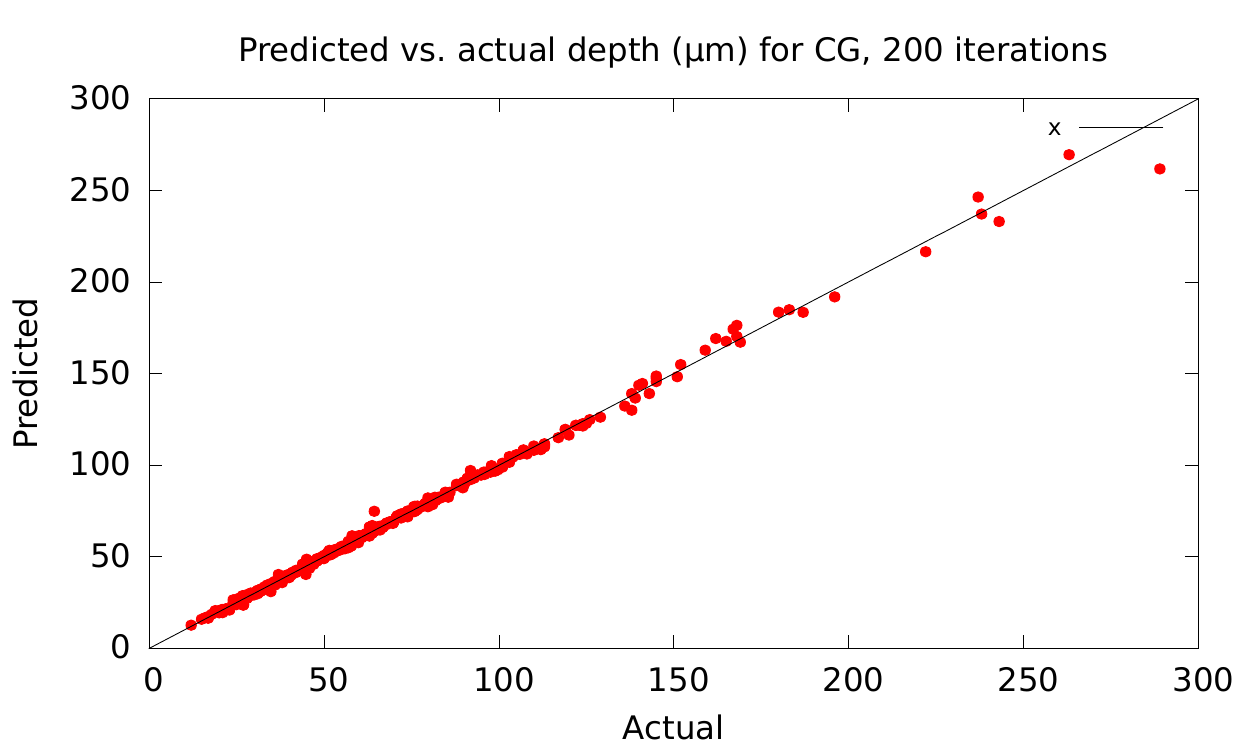} \\
(a) & (b) & (c) & (d) \\
\end{tabular} 
\vspace{-0.2cm}
\caption{GP predictions using an iterative solver. The results show
  the LOO predicted vs. actual values of the melt-pool depth obtained
  using the optimal hyperparameters. (a) residual threshold = 1.0e-4,
  maximum iterations = 462, time = 3570s. (b) residual threshold =
  1.0e-3, maximum iterations = 462, time = 3425s. (c) maximum
  iterations = 100, time = 1982s. (d) maximum iterations = 200, time =
  2496s.}
\label{fig:itsol_results}
\centering
\begin{tabular}{c}
\includegraphics[width=0.5\textwidth]{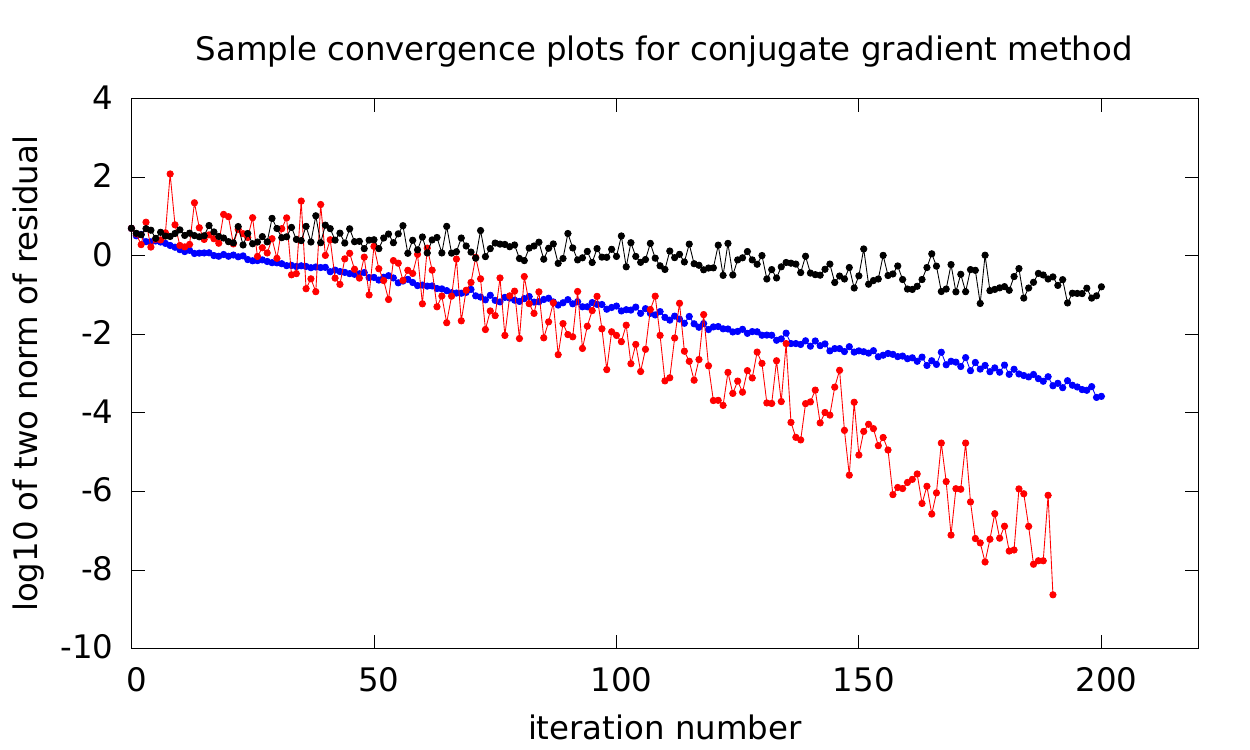} 
\end{tabular} 
\vspace{-0.2cm}
\caption{Convergence of the conjugate gradient solver for three
  different sets of hyperparameters, illustrating fast convergence in
  some cases, with the two norm of the residual reducing to 1.0e-8 in
  less than 200 iterations for a system of size 462, but quite slow
  and noisy in other cases. }
\label{fig:itsol_conv_results}
\end{figure*}

We conducted our experiments with iterative solvers by replacing all
calls to the direct solver in our code with a call to the conjugate
gradient solver described in Algorithm~\ref{algo:cg}. To obtain an
approximate solution for the full data set, we considered two
options. First, we set the maximum iterations to 462, which is the
size of the system, and experimented with two values --- 1.0e-3 and
1.0e-4 --- of the threshold on the two norm of the residual. The
iterations would stop when either of these two conditions was
satisfied. We present the results in Figure~\ref{fig:itsol_results}(a)
and (b) for the optimal hyperparameters. In evaluating the results, we
found that for some sets of hyperparameter values, the residual did
not meet the threshold criterion even after 462 iterations, suggesting
that the covariance matrix is ill-conditioned for these
hyperparameters.

Next, for greater control over the iterative process, we set the
maximum number of iterations to two values --- 100 and 200 --- with a
very low value of 1.0e-8 for the residual threshold. The results are
presented in Figure~\ref{fig:itsol_results}(c) and (d). Our
expectation was that for all 100 sets of hyperparameter values, the
constraint on the number of iterations would occur before the residual
reduced below the threshold. However, we found that this was not
always the case, suggesting that some parameter sets resulted in very
well-conditioned systems.  Figure~\ref{fig:itsol_conv_results} shows
the large disparity among convergence rates for three sample sets of
hyperparameters.  We also found that stopping the process after 100
iterations frequently meant that the iterations were far from
converging, and likely to identify a less-than-optimal set of
hyperparameters, resulting in the higher prediction error shown in
Figure~\ref{fig:itsol_results}(c).

Overall, a key observation in our experiments with iterative solvers
was that they were slow. Even the fastest case, with a limit of 100
iterations, at 1982s, took just as long as the direct solver on the
full data set at 1979s (Figure~\ref{fig:gp_orig}), but gave less
accurate results, as indicated by the scatter in
Figure~\ref{fig:itsol_results}(c).  Unlike the direct solvers, it is
also challenging to find appropriate stopping criteria for the
iterative solvers.

Finally, we observe that it is possible to improve the convergence of the
iterative solvers using preconditioning~\cite{saad03:book}. A possible
preconditioner would be a block diagonal preconditioner, but, our
experiences in Section~\ref{sec:results_taper} suggest that the
independent-block GP method would be computationally faster.

%
\subsection{Understanding the high-dimensional solution space}
\label{sec:results_high_dim}
%

We next explored the four-dimensional solution space in which
instances from the ET5000 data set with depth in the range $60 \pm
\Delta$, with $\Delta = 2 \ \text{and} \ 5$, lie using parallel
coordinate plots and Kohonen self organizing maps that were described
in Sections~\ref{sec:parplots} and \ref{sec:soms}, respectively.

\begin{figure*}[!tb]
\centering
\begin{tabular}{cc}
\includegraphics[width=0.46\textwidth]{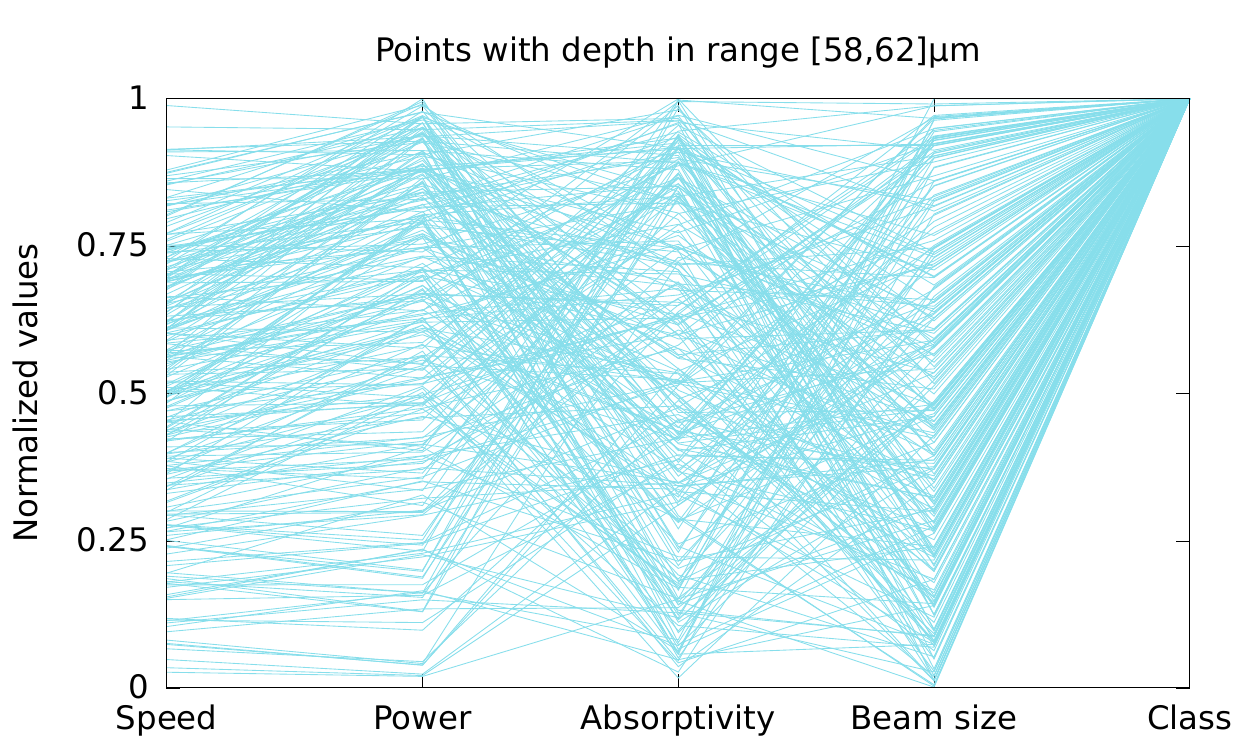} &
\includegraphics[width=0.46\textwidth]{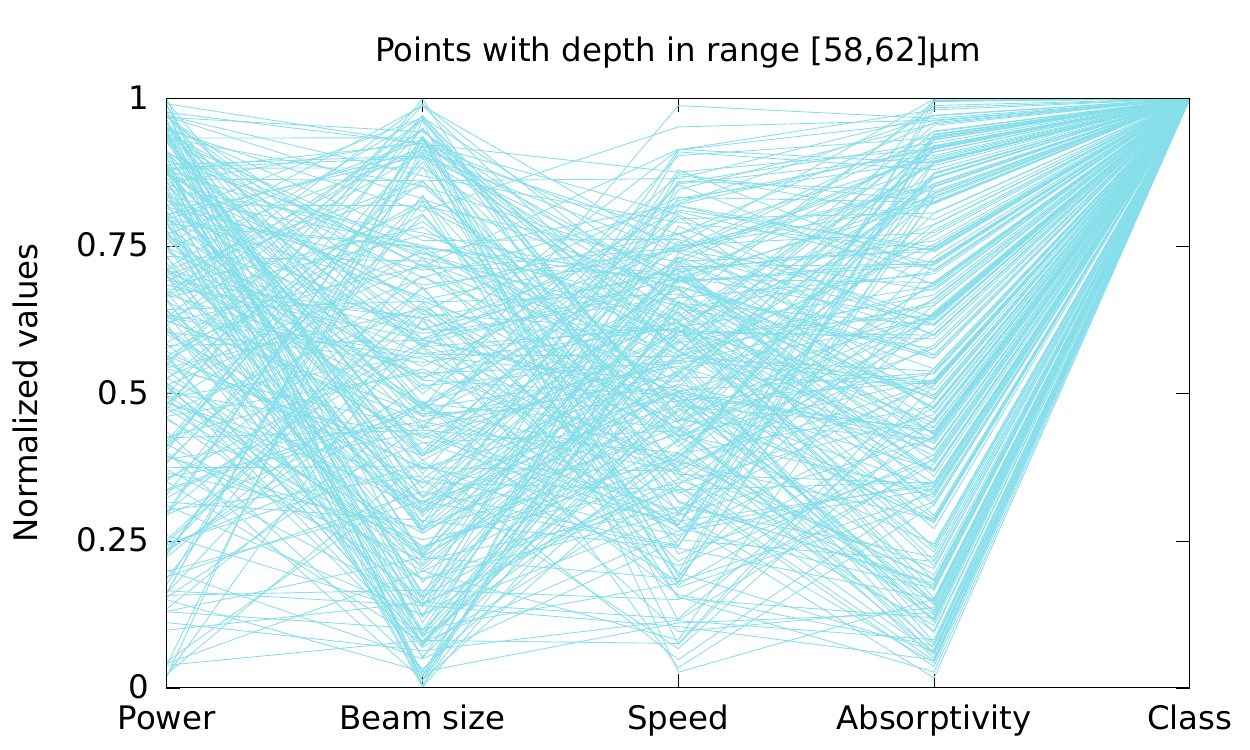} \\
\end{tabular}
\vspace{-0.2cm}
\caption{Parallel plots showing the points in the ET-5000 data set
  with melt-pool depth in the range [58,62]\micron. Two different
  ordering of the four input variables are shown, which allows each
  variable to be plot adjacent to each of the remaining three
  variables. The class value of 1 indicates that these are points that
  meet the criterion on the melt-pool depth. }
\label{fig:parallel_plots}
\end{figure*}

First, in Figure~\ref{fig:parallel_plots}, we show the solution
instances with depth $60 \pm 2$ in parallel coordinates. As this data
set has only 271 instances, the smaller number of points makes it
easier to identify patterns in the solution. We consider two different
orderings of the input dimensions to ensure each variable can be
adjacent to all remaining variables. We observe that for the solution
instances, speed and power have a nearly linear relationship as the
lines connecting the two are nearly, but not perfectly, horizontal.
The solution has few points at both low and high values of speed
(based on the sparsity of points at the extreme ends of the speed
axis), and has more instances at higher values of the power axis. This
agrees with the plot in the left panel of
Figure~\ref{fig:two_space_projection}. It is however difficult to draw
any additional conclusions regarding the solution to the inverse
problem from the parallel plots, though we found them invaluable in
understanding the full dataset itself~\cite{kamath14:density}.

The self organizing maps provide a very different insight into the
solution to the inverse problem. In our work, we considered two
options for the distance in Equation~\ref{eqn:ref_node} to find the
reference node - Euclidean distance and weighted Euclidean distance,
where we used as weights the inverse of the weighted length scales from
Figure~\ref{fig:gp_orig}, as a smaller length scale in GP indicates a more
important input.

\begin{figure*}[tb]
\centering
\begin{tabular}{cccc}
\includegraphics[trim = 2cm 0 2cm 0, clip, width=0.23\textwidth]{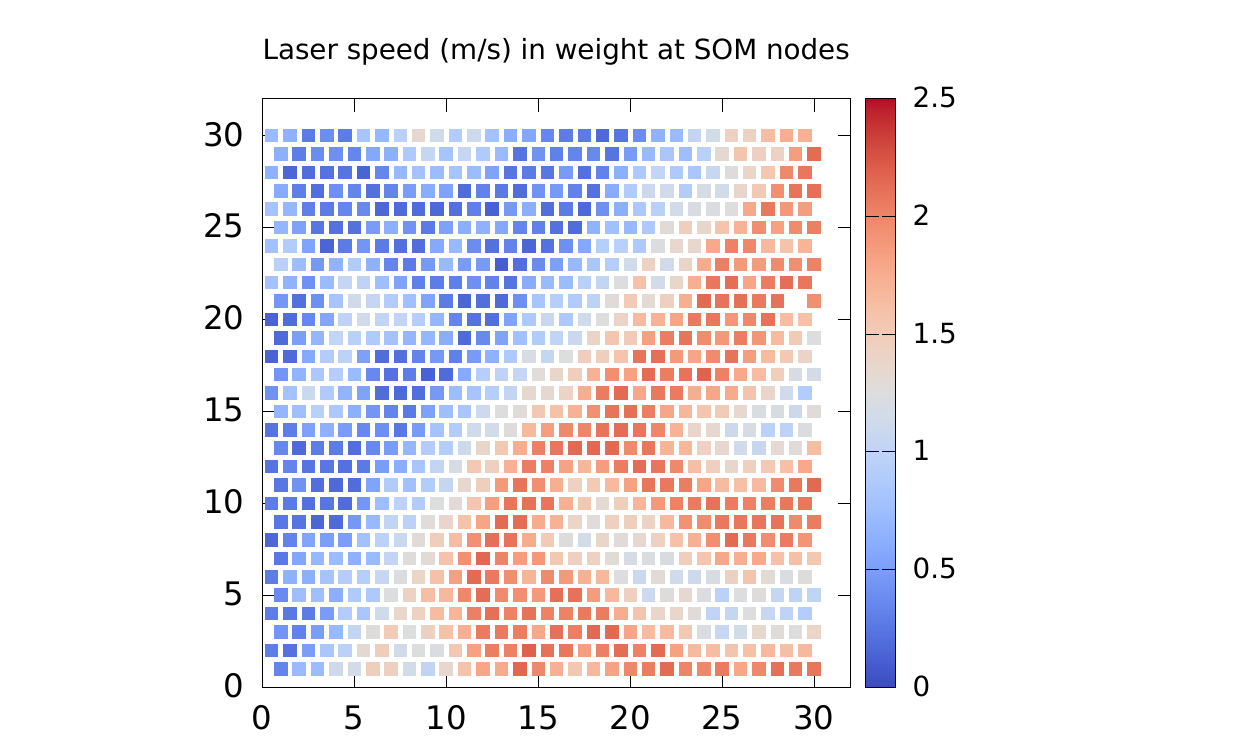} &
\includegraphics[trim = 2cm 0 2cm 0, clip, width=0.23\textwidth]{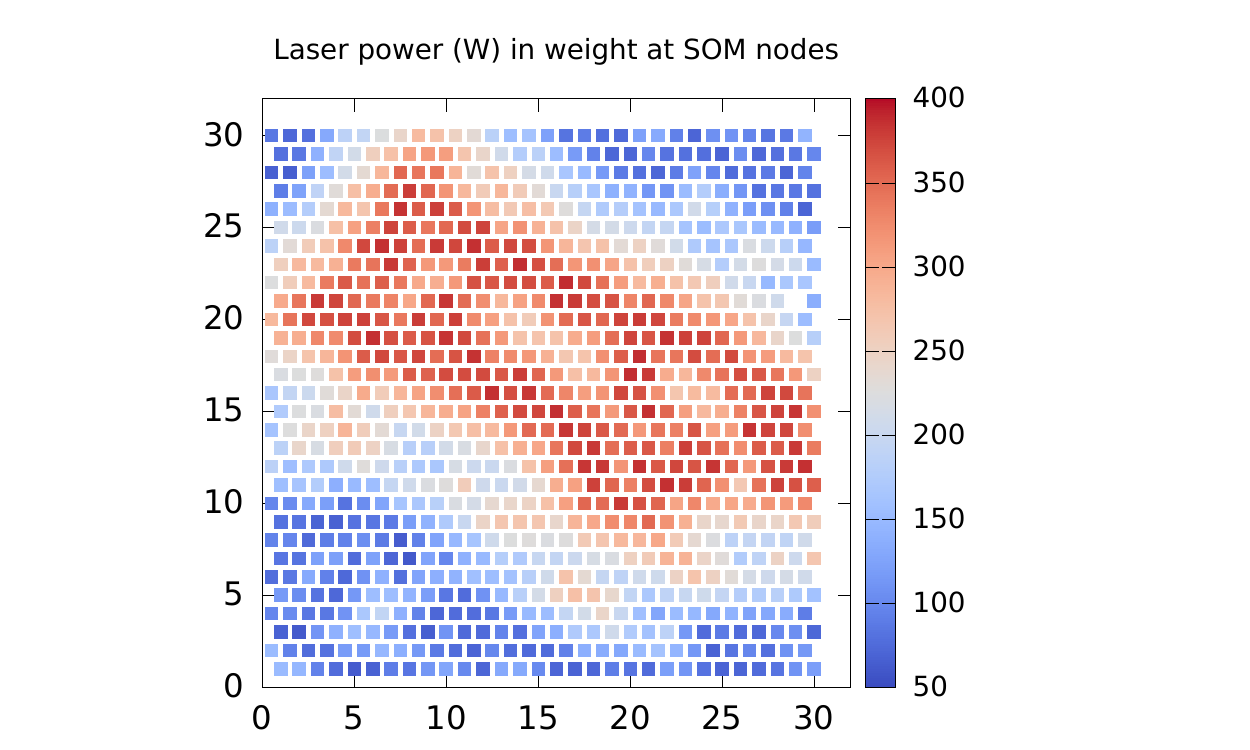} &
\includegraphics[trim = 2cm 0 2cm 0, clip, width=0.23\textwidth]{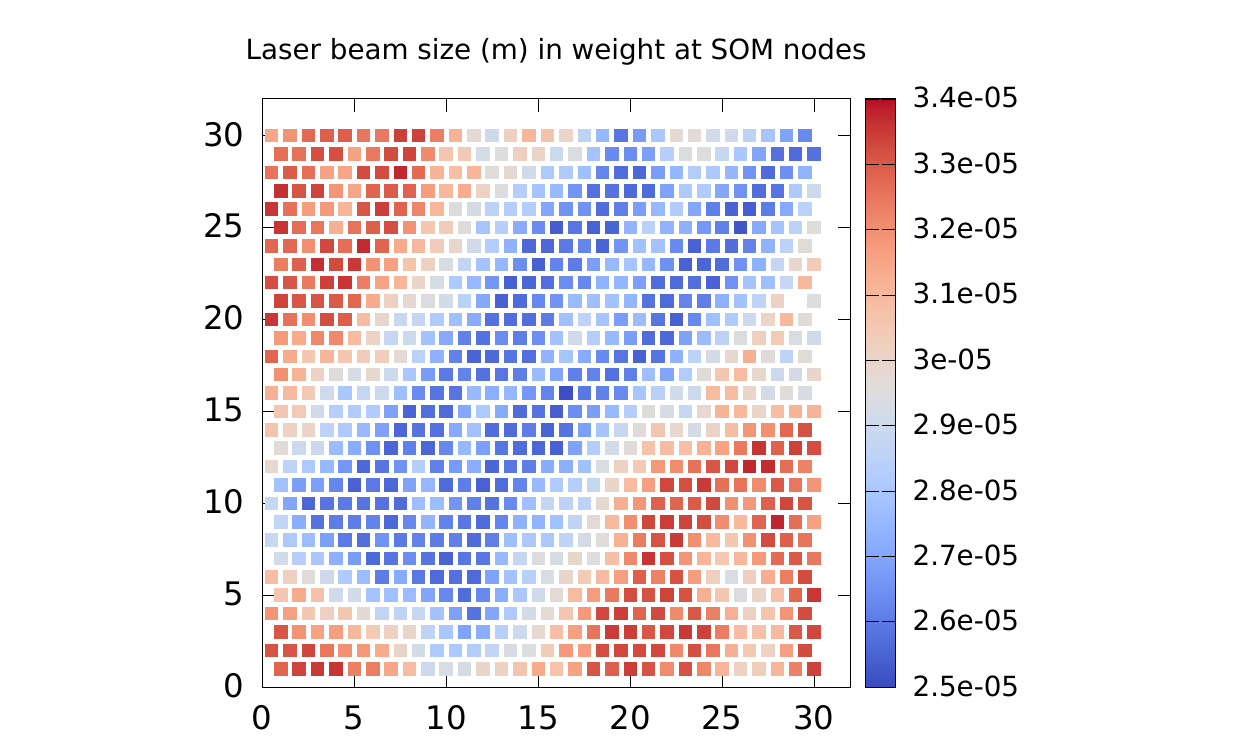} &
\includegraphics[trim = 2cm 0 2cm 0, clip, width=0.23\textwidth]{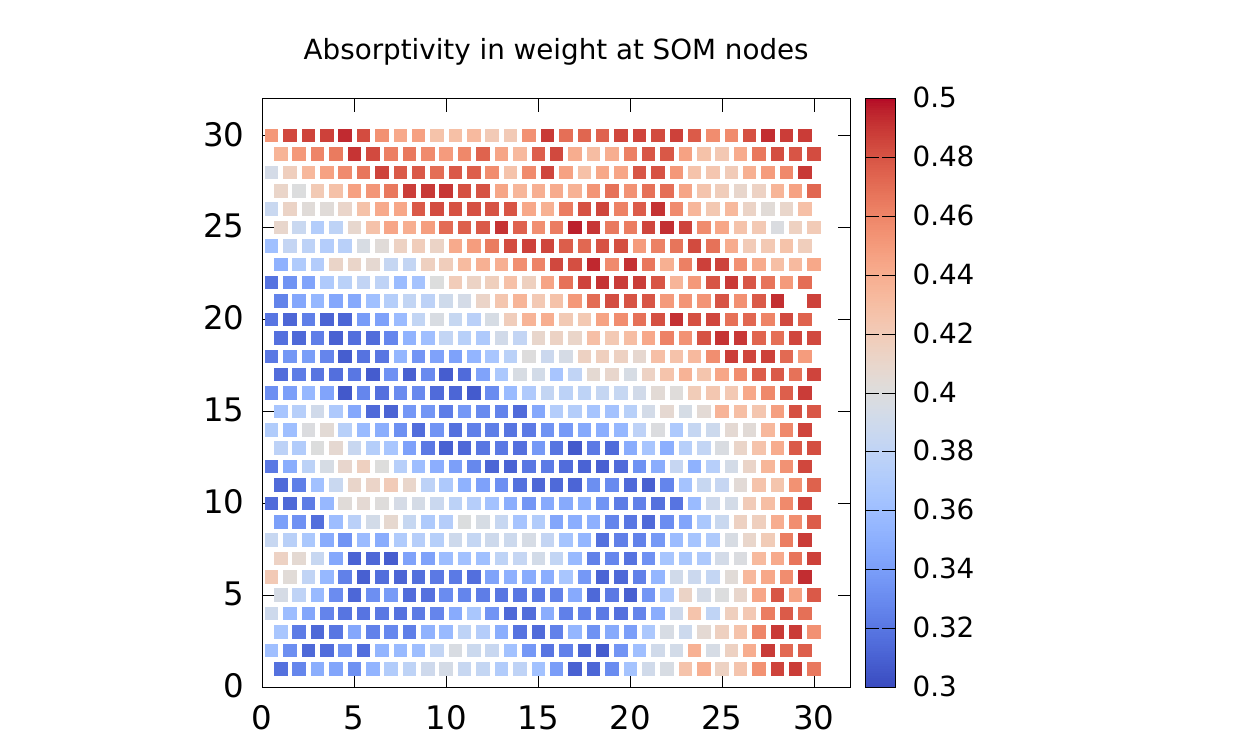} \\
(a) & (b) & (c) & (d) \\ 
\includegraphics[trim = 2cm 0 2cm 0, clip, width=0.23\textwidth]{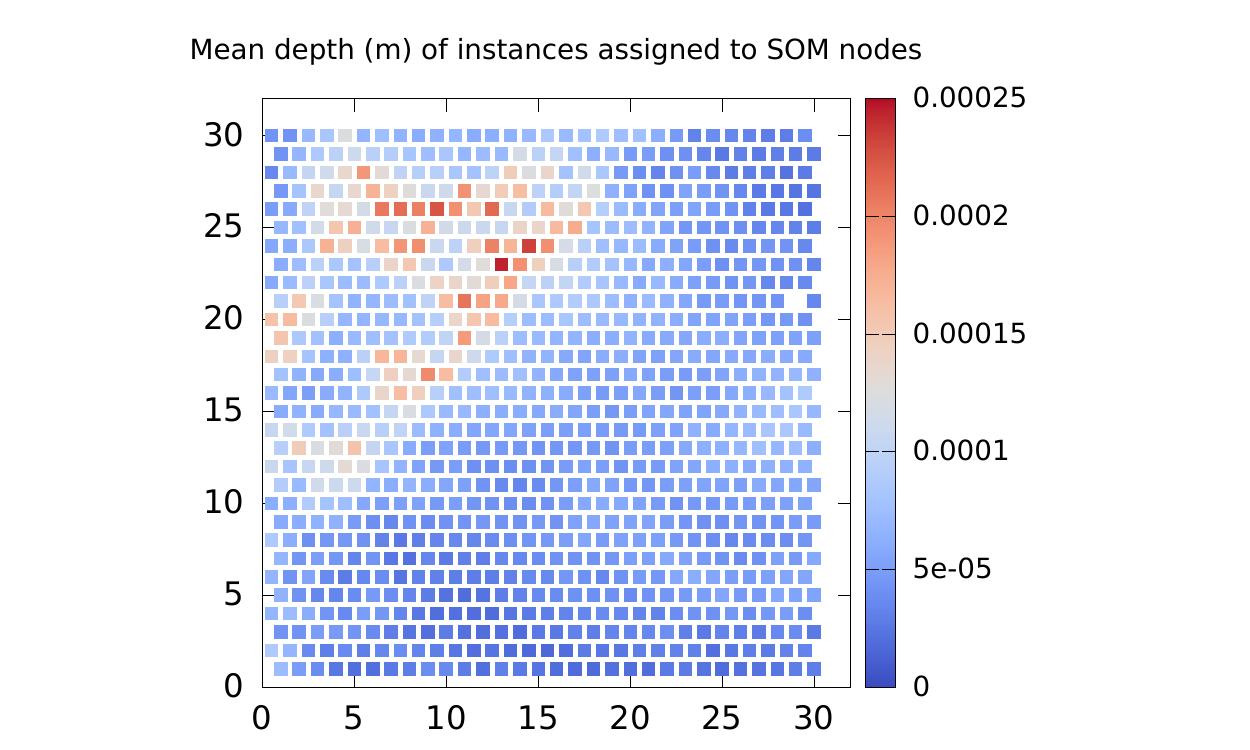} &
\includegraphics[trim = 2cm 0 2cm 0, clip, width=0.23\textwidth]{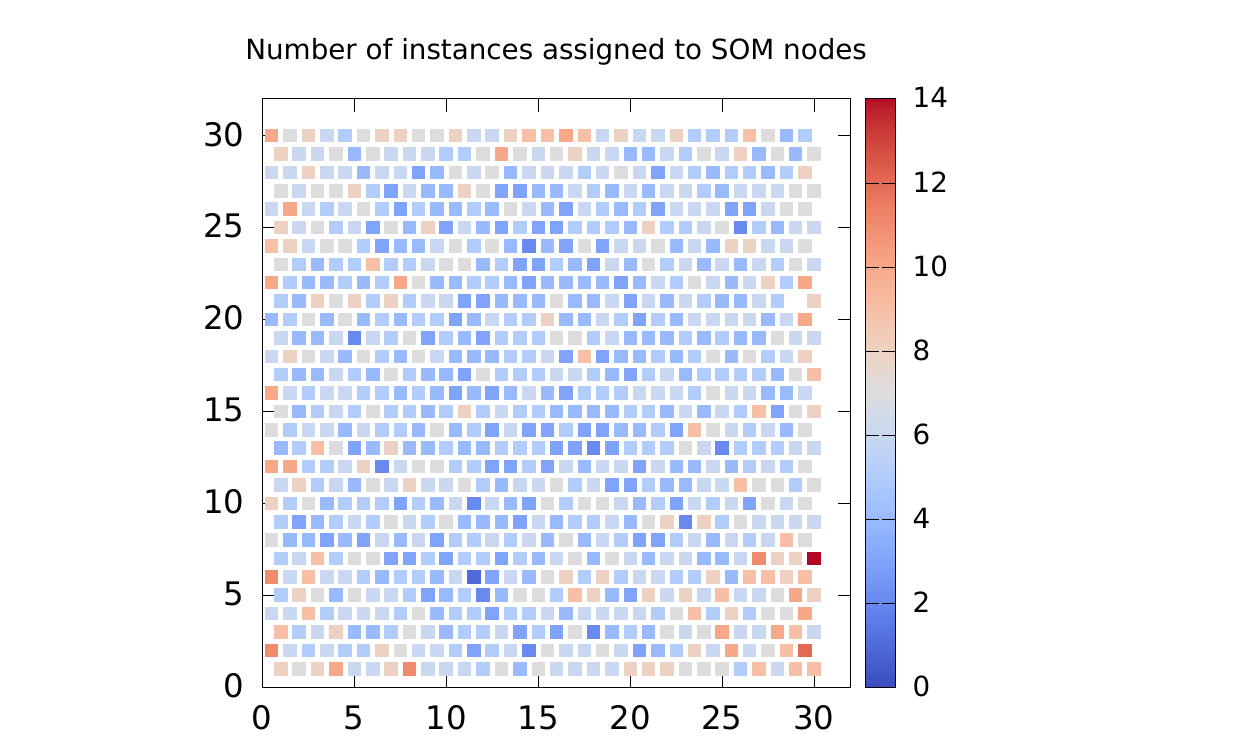} &
\includegraphics[trim = 2cm 0 2cm 0, clip, width=0.23\textwidth]{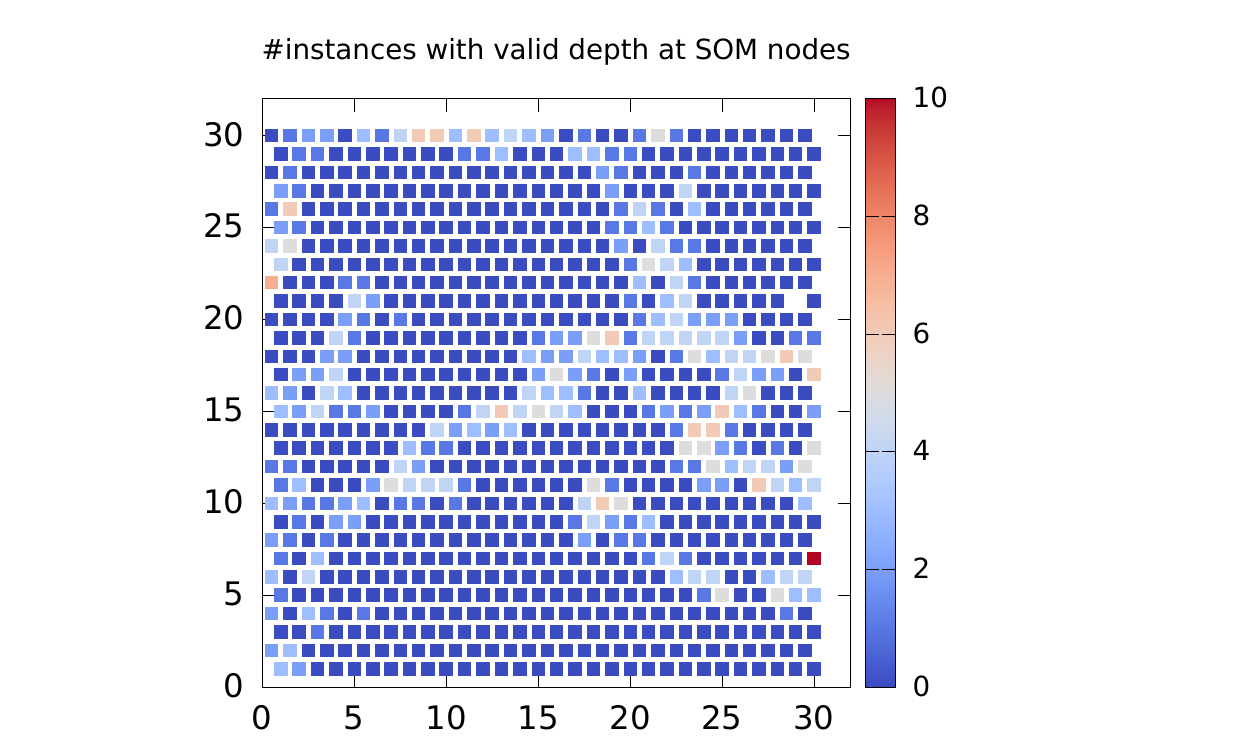} &
\includegraphics[trim = 2cm 0 2cm 0, clip, width=0.23\textwidth]{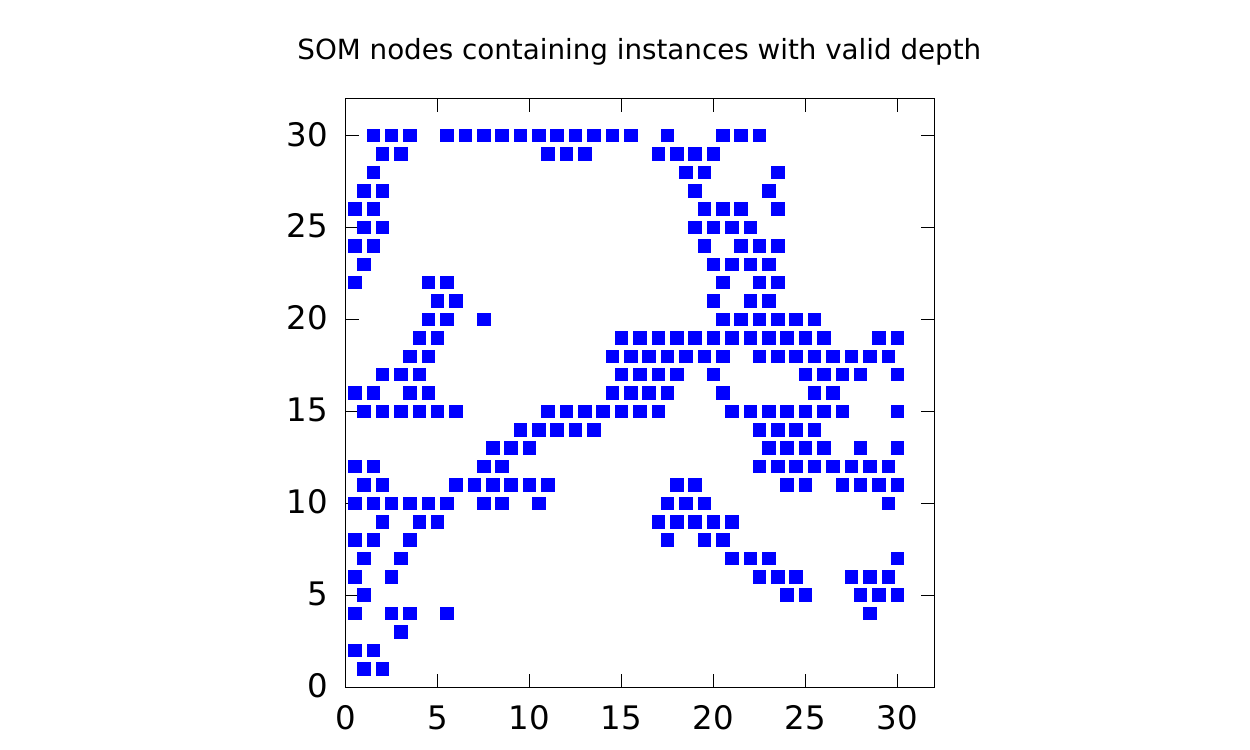} \\
(e) & (f) & (g) & (h) \\
\end{tabular}
\vspace{-0.2cm}
\caption{Results for the ET5000 data set using a $30\times30$ SOM and
  unweighted distances. (a)-(d) show the values of the weights at each
  SOM node in the four input dimensions --- speed, power, beam size,
  and absorptivity. (e) Mean value of the instances assigned to each
  SOM node. (f) Number of instances assigned to each SOM node. (g)
  Number of instances with melt-pool depth in range [55,65]\micron~ at
  each SOM node, with the SOM nodes with non-zero such instances
  hightlighted in (h). Note that there is a node with no instances
  assigned to it near the right edge.}
\label{fig:som_data_unwted}
\end{figure*}

\begin{figure*}[!htb]
\centering
\begin{tabular}{cccc}
\includegraphics[trim = 2cm 0 2cm 0, clip, width=0.23\textwidth]{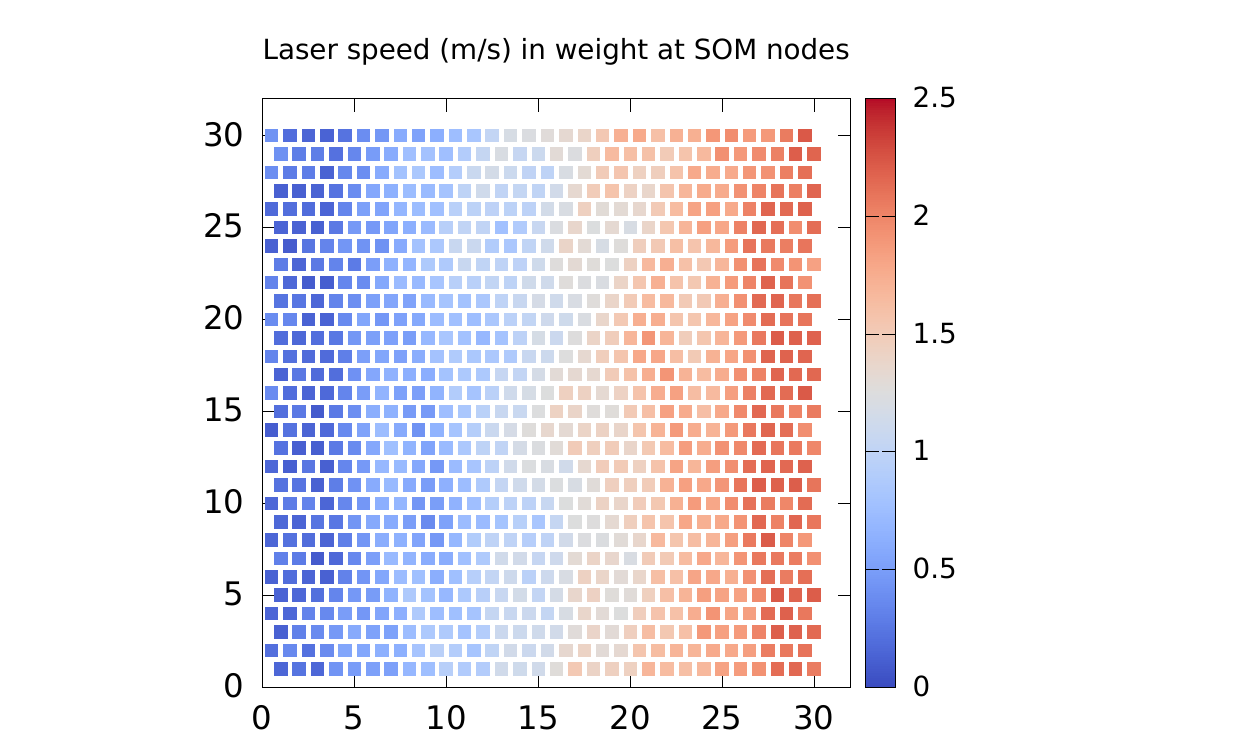} &
\includegraphics[trim = 2cm 0 2cm 0, clip, width=0.23\textwidth]{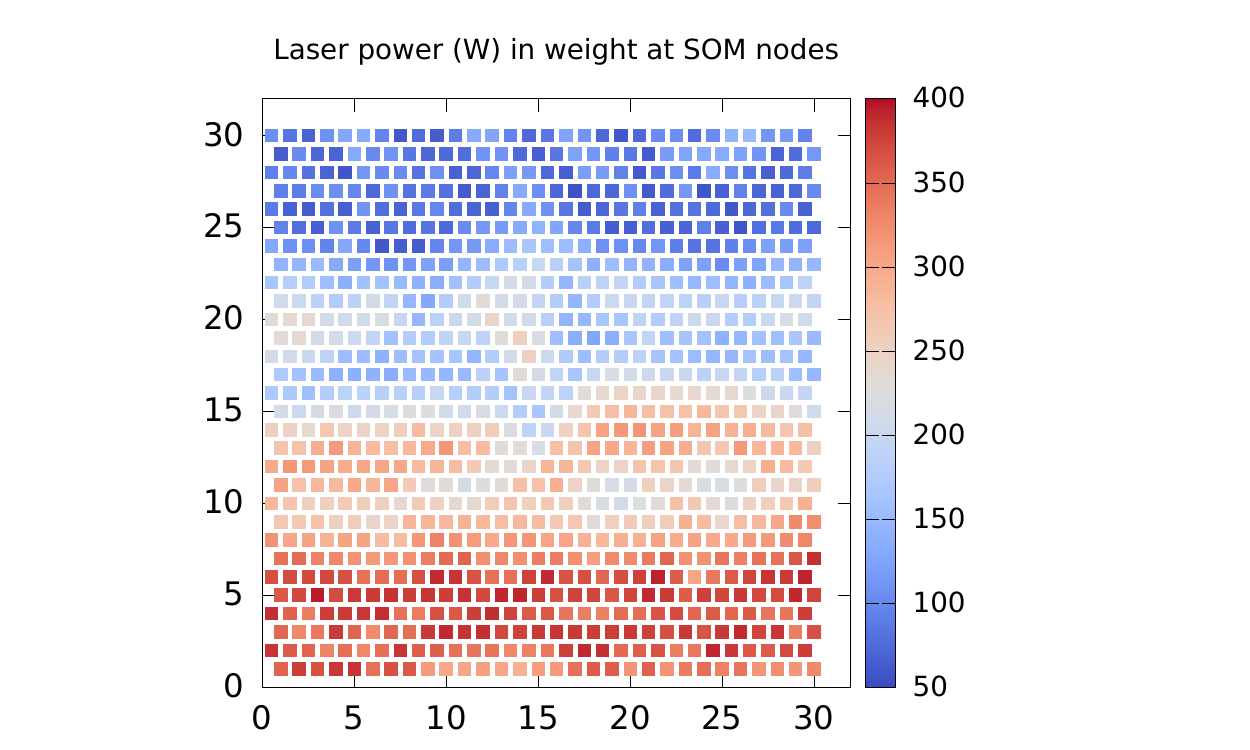} &
\includegraphics[trim = 2cm 0 2cm 0, clip, width=0.23\textwidth]{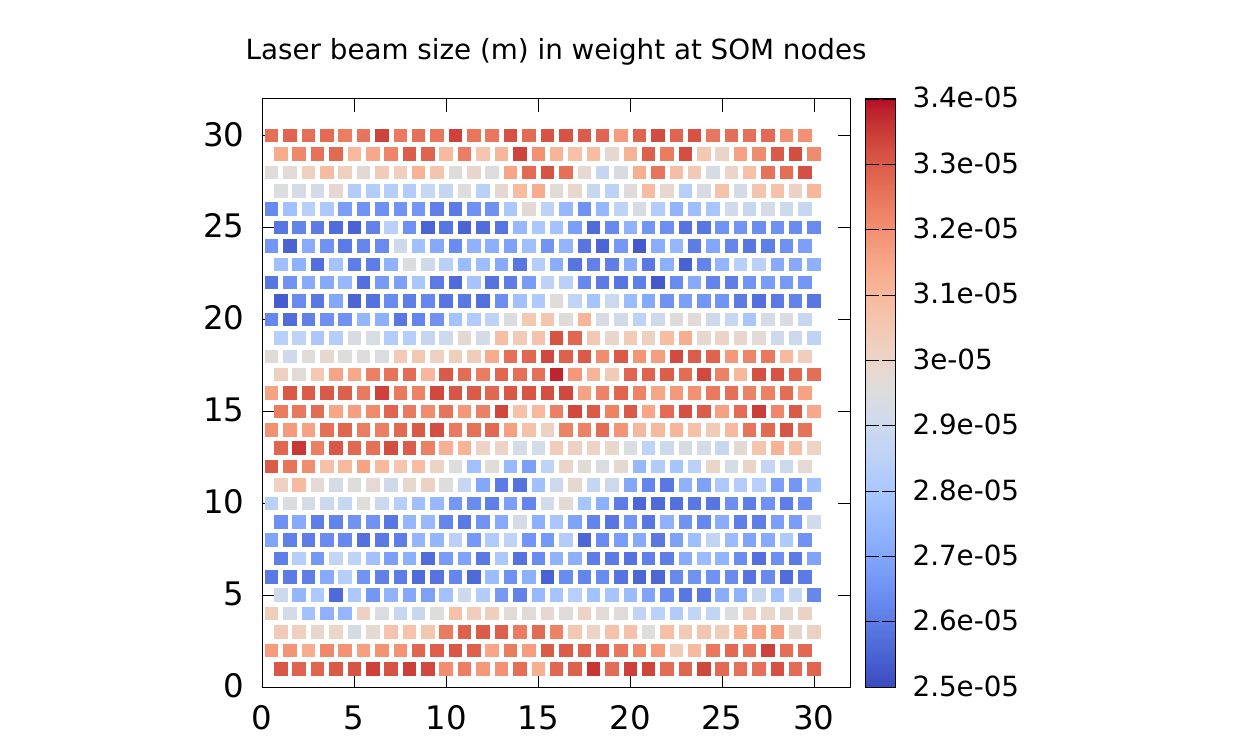} &
\includegraphics[trim = 2cm 0 2cm 0, clip, width=0.23\textwidth]{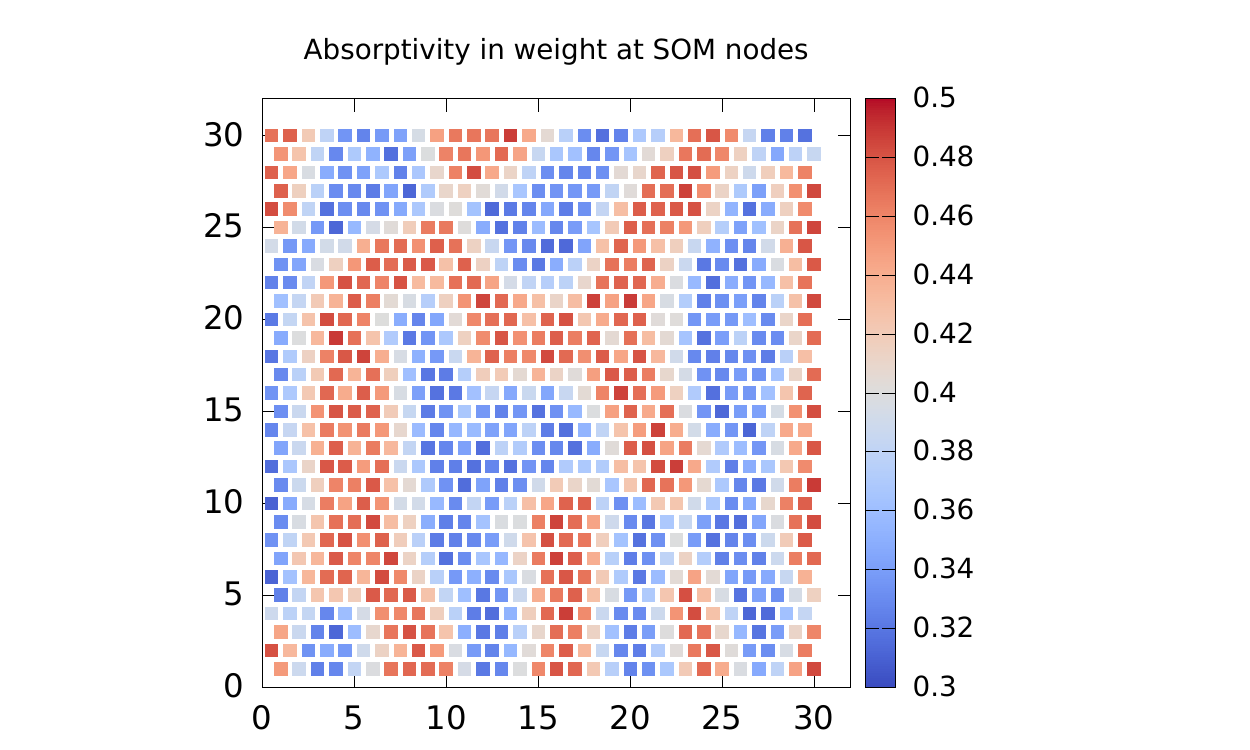} \\
(a) & (b) & (c) & (d) \\
\includegraphics[trim = 2cm 0 2cm 0, clip, width=0.23\textwidth]{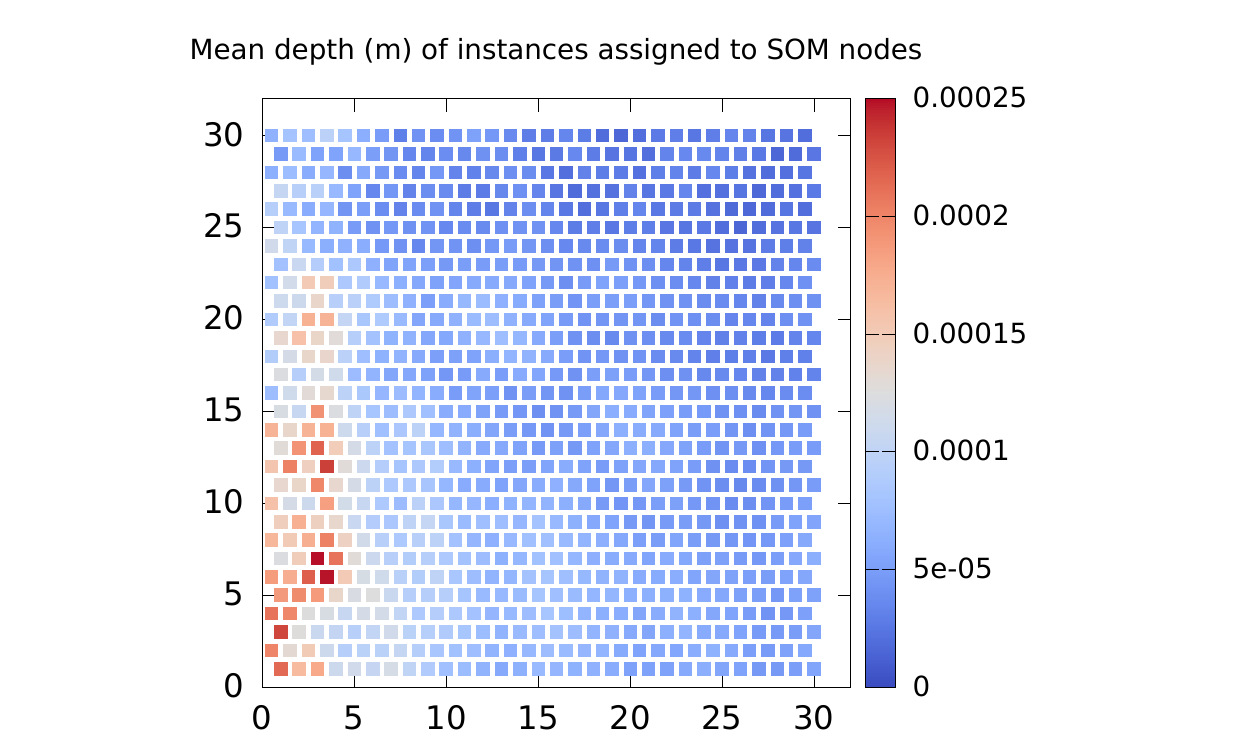} &
\includegraphics[trim = 2cm 0 2cm 0, clip, width=0.23\textwidth]{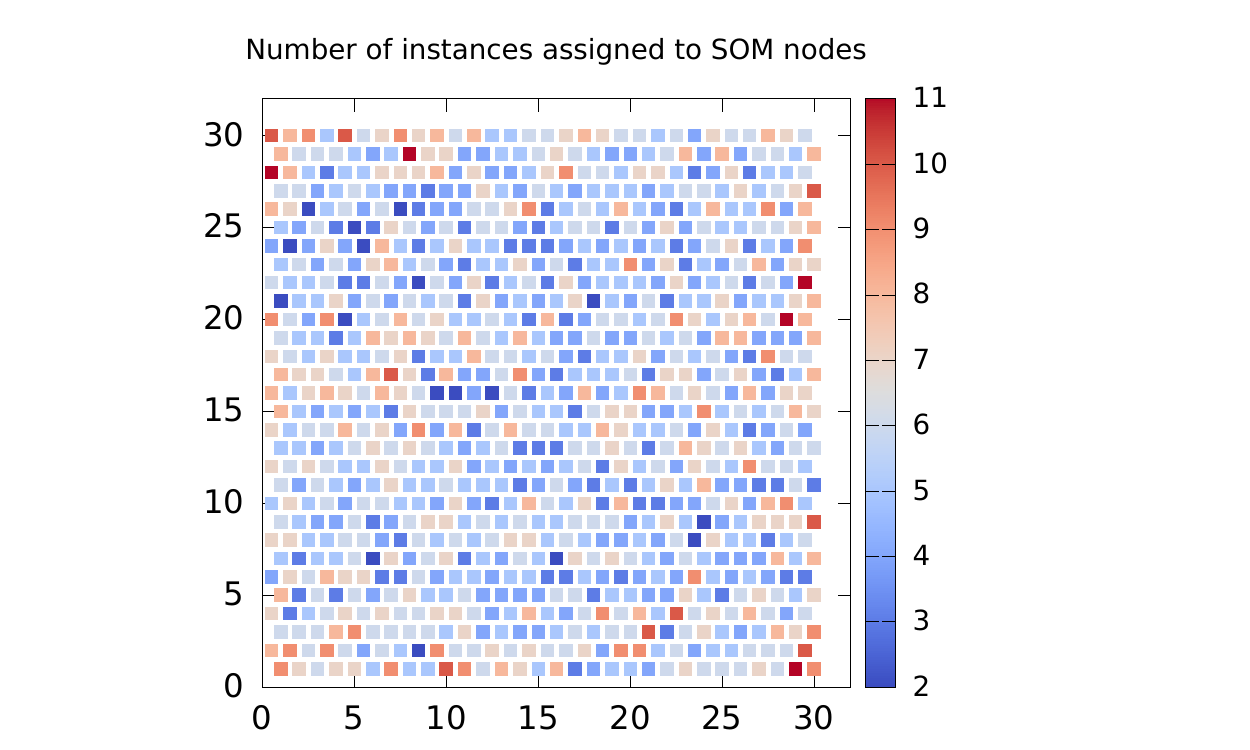} &
\includegraphics[trim = 2cm 0 2cm 0, clip, width=0.23\textwidth]{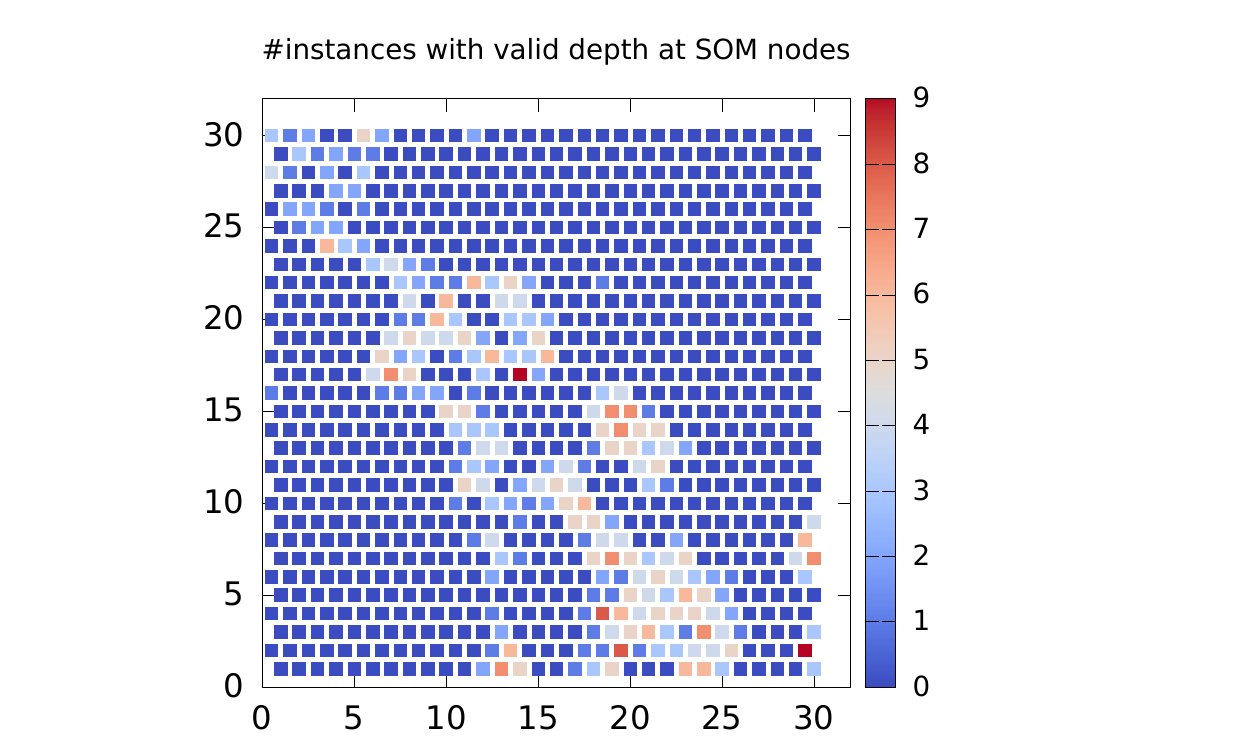} &
\includegraphics[trim = 2cm 0 2cm 0, clip, width=0.23\textwidth]{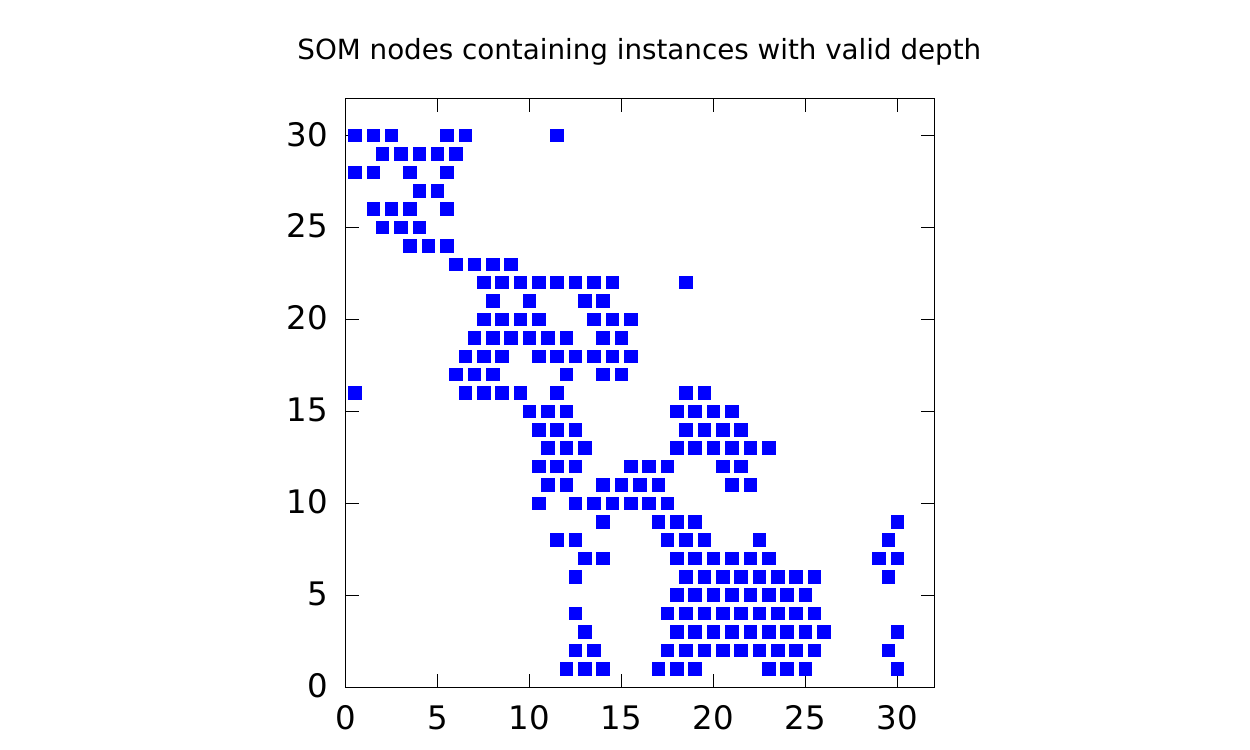} \\
(e) & (f) & (g) & (h) \\
\end{tabular}
\vspace{-0.2cm}
\caption{Results for the ET5000 data set using a $30\times30$ SOM and
  weighted distances. (a)-(d) show the values of the weights at each
  SOM node in the four input dimensions --- speed, power, beam size,
  and absorptivity. (e) Mean value of the instances assigned to each
  SOM node. (f) Number of instances assigned to each SOM node. (g)
  Number of instances with melt-pool depth in range [55,65]\micron~ at
  each SOM node, with the SOM nodes with non-zero such instances
  hightlighted in (h). }
\label{fig:som_data_wted}
\end{figure*}

First, in Figures~\ref{fig:som_data_unwted} and
\ref{fig:som_data_wted}, we show results for a $30 \times 30$ map for
the ET5000 data set using unweighted and weighted distances,
respectively. The four plots in the top row of each figure show the
value of each dimension in the weight vector at the SOM nodes,
followed in the second row by some statistics on the instances
assigned to each node. As the 5000 instances in the data set are
spread out uniformly in the input domain, we do not expect the SOM to
identify any topological structure in the location of the
instances in the data set.

We observe that weighted distances result in a better ``organized''
SOM as the transition from the high to low values of an input
variable, such as speed and power, is quite smooth. In contrast, with
unweighted distances, the regions with very low-valued points (in dark
blue) and the very high valued points (in dark red) both have
medium-valued regions (in lighter blue and red) in them. In comparing
panels (a) and (b) with panel (e) for both weighted and unweighted
distances, we observe that deep melt pools occur at low values of
speed and higher values of power, with the correlation being clearer
in plots using weighted distances. Figures~\ref{fig:som_data_unwted}
and \ref{fig:som_data_wted} also show the number of instances assigned
to each SOM node, aas well as the SOM nodes that have instances with
depth in the range [55,65]\micron. The latter indicate that the
solution to the inverse problem is spread out in the four-dimensional
input space in the form of ``veins'', rather than clustered in a
bunch, in which case they would have appeared in a few nodes that are
close to each other.

\begin{figure*}[htb]
\centering
\begin{tabular}{cc}
\includegraphics[width=0.39\textwidth]{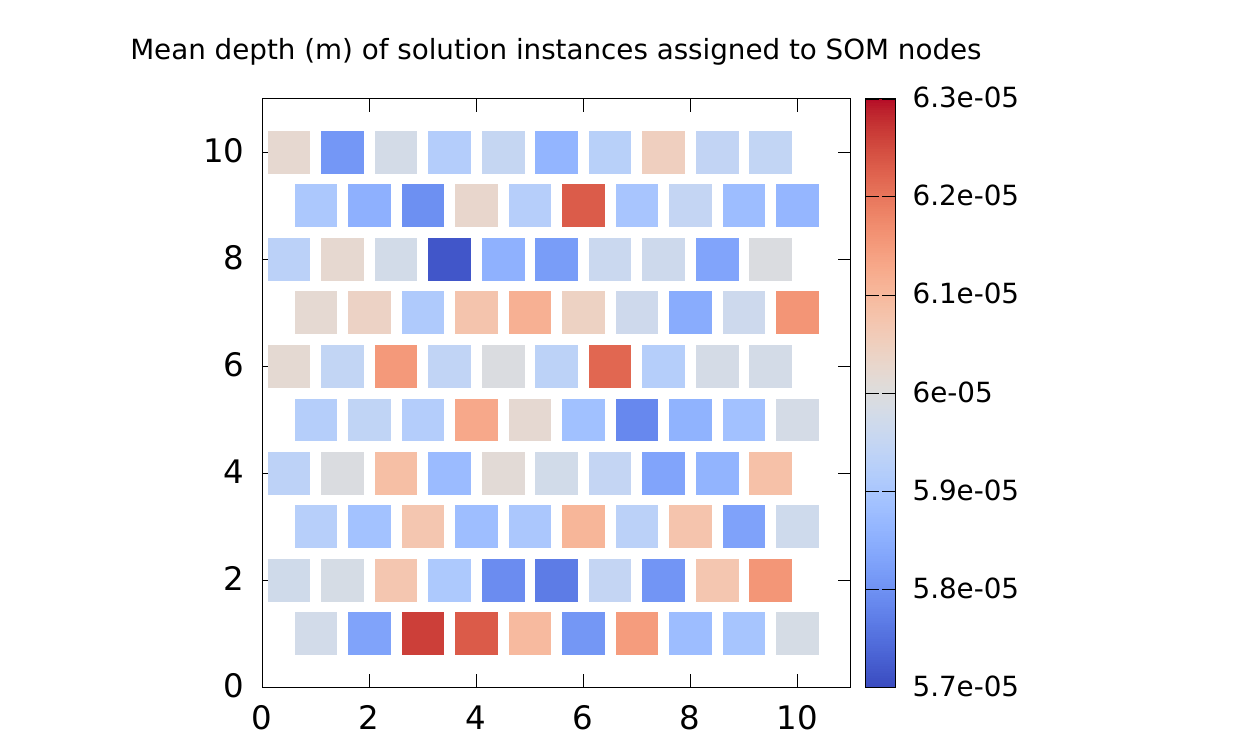} &
\includegraphics[width=0.39\textwidth]{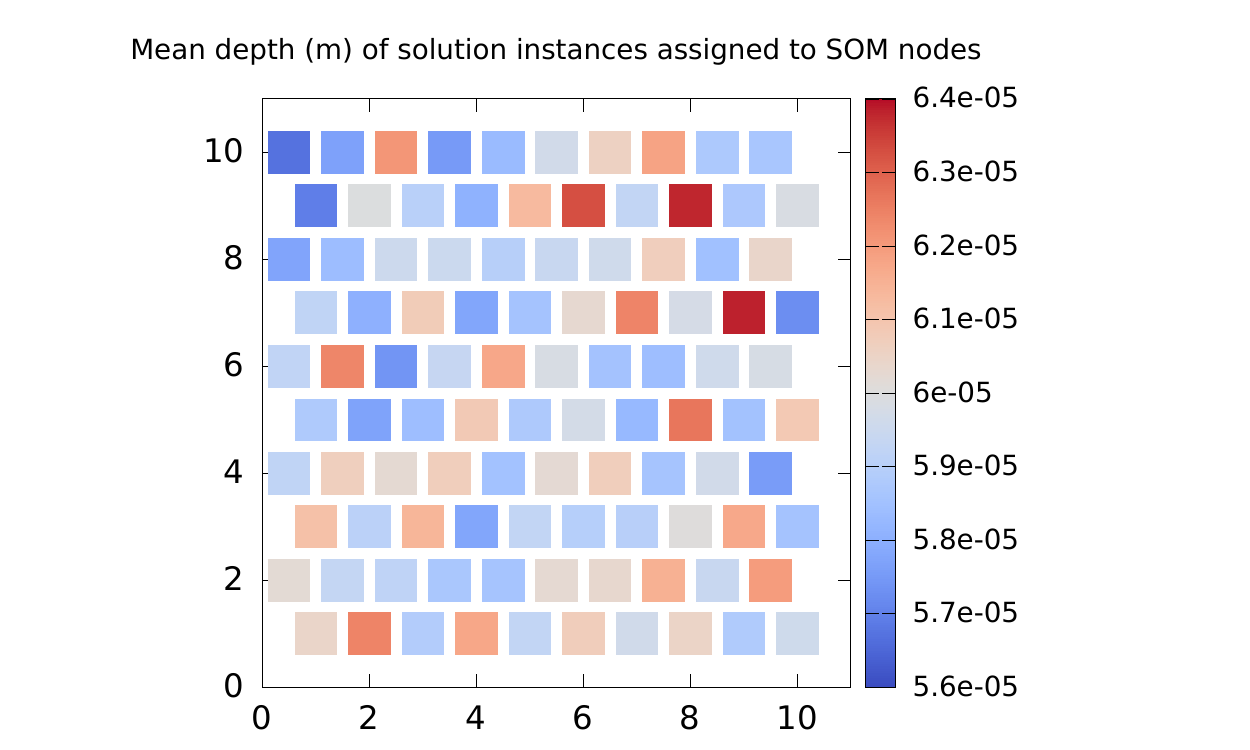} \\
\end{tabular}
\vspace{-0.2cm}
\caption{Solution instances, with melt-pool depth in range
  [55,65]\micron, after processing using a $10\times10$ SOM. The plots
  show the mean value of the melt-pool depth for the instances
  assigned to each node using unweighted (left) and weighted (right)
  distances. }
\label{fig:som_soln_1}
\end{figure*}

\begin{figure*}[htb]
\centering
\begin{tabular}{cccc}
\includegraphics[trim = 1.5cm 0 2cm 0, clip, width=0.23\textwidth]{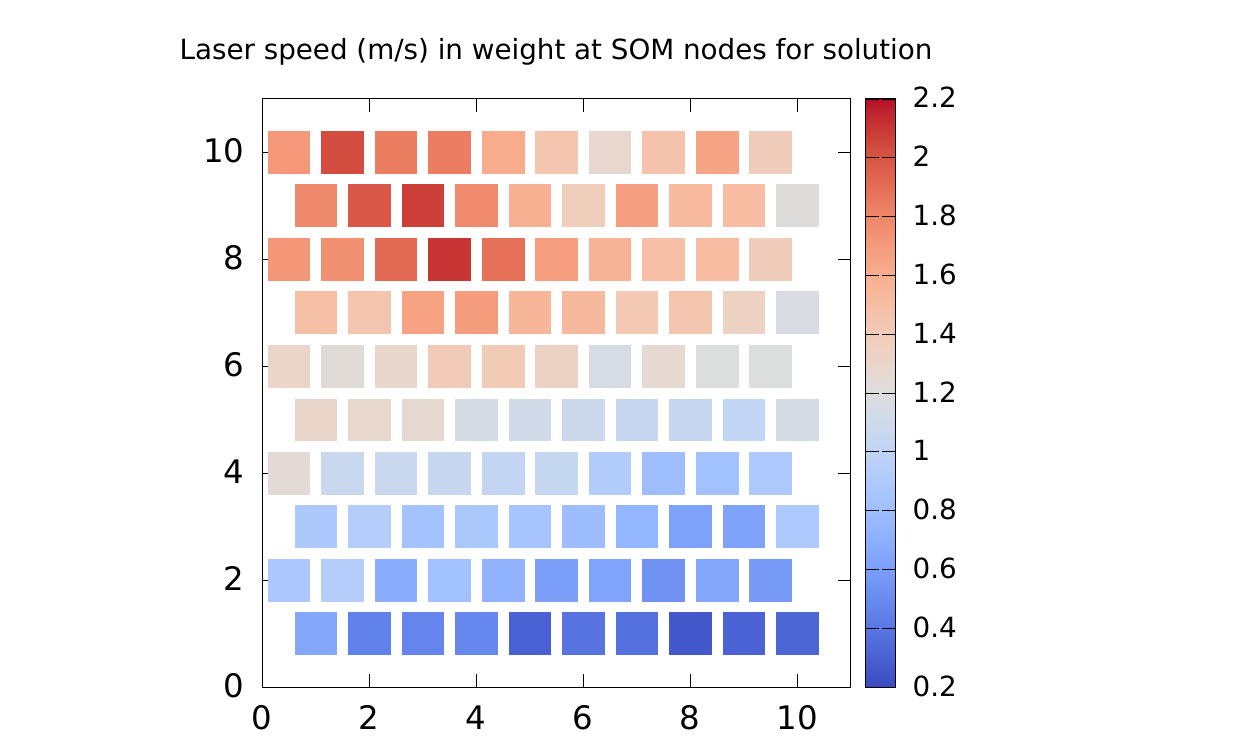} &
\includegraphics[trim = 1.5cm 0 2cm 0, clip, width=0.23\textwidth]{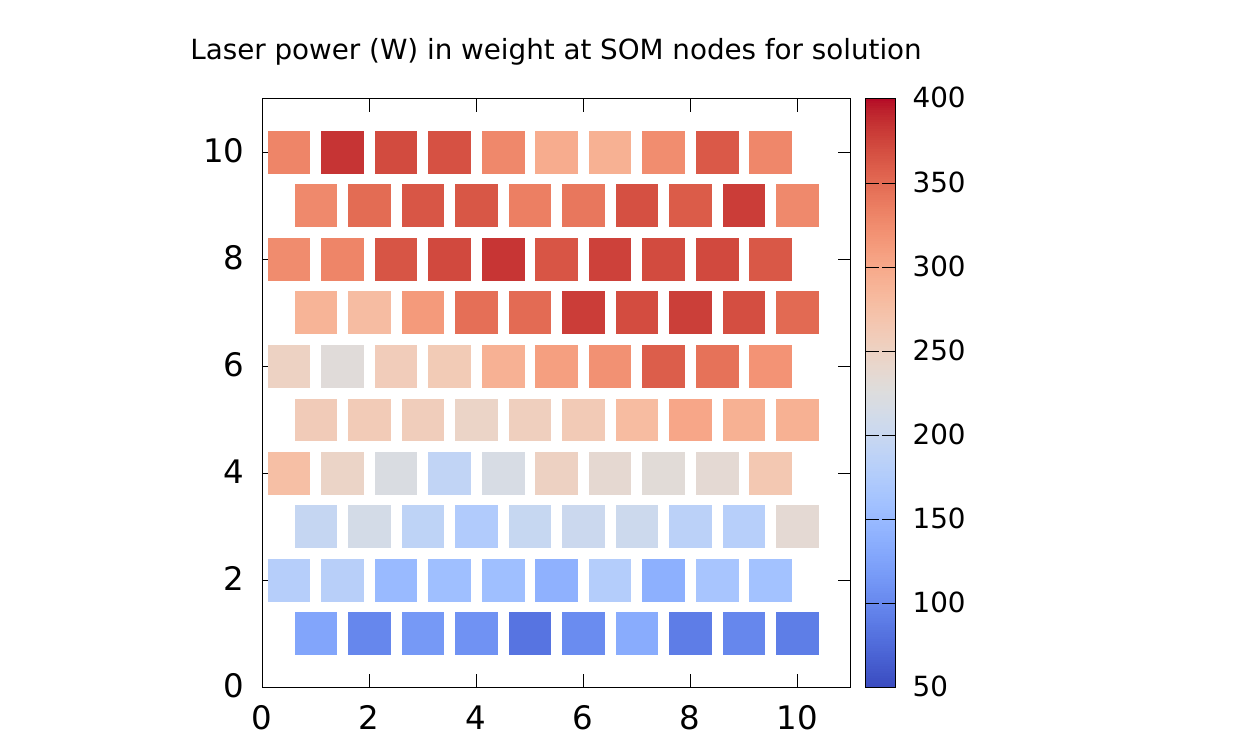} &
\includegraphics[trim = 1.5cm 0 2cm 0, clip, width=0.23\textwidth]{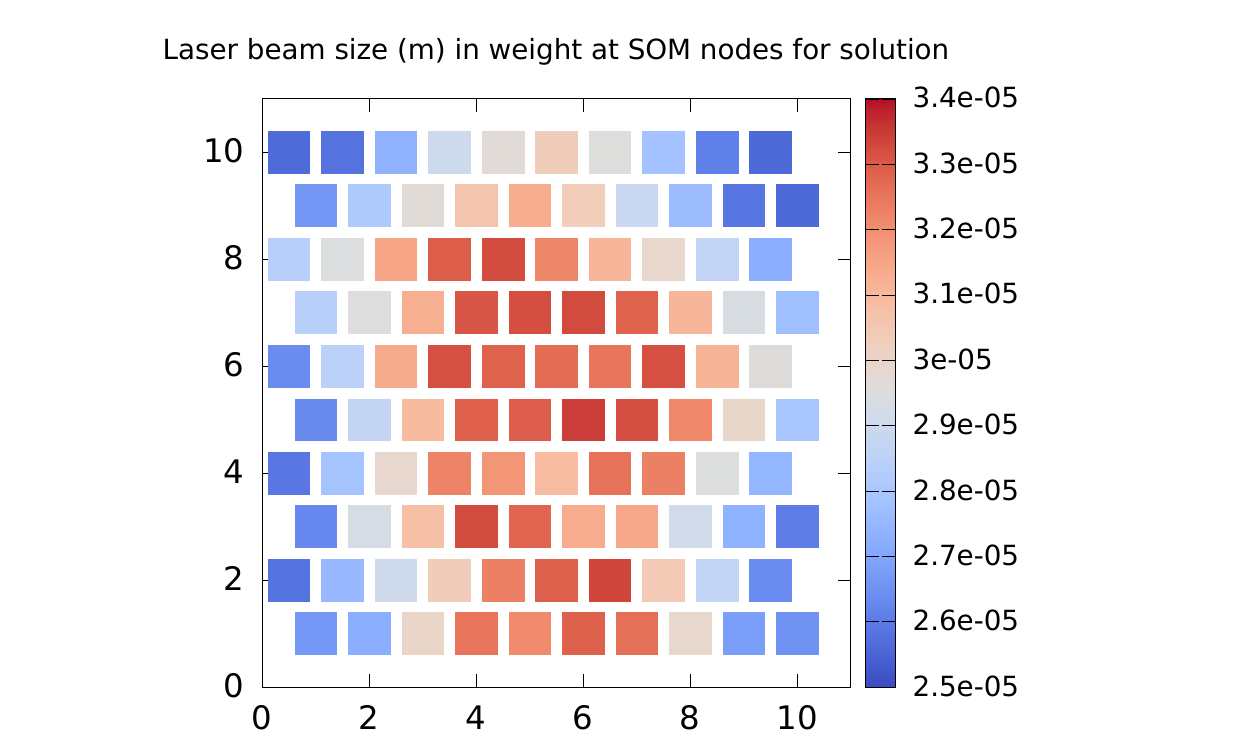} &
\includegraphics[trim = 1.5cm 0 2cm 0, clip, width=0.23\textwidth]{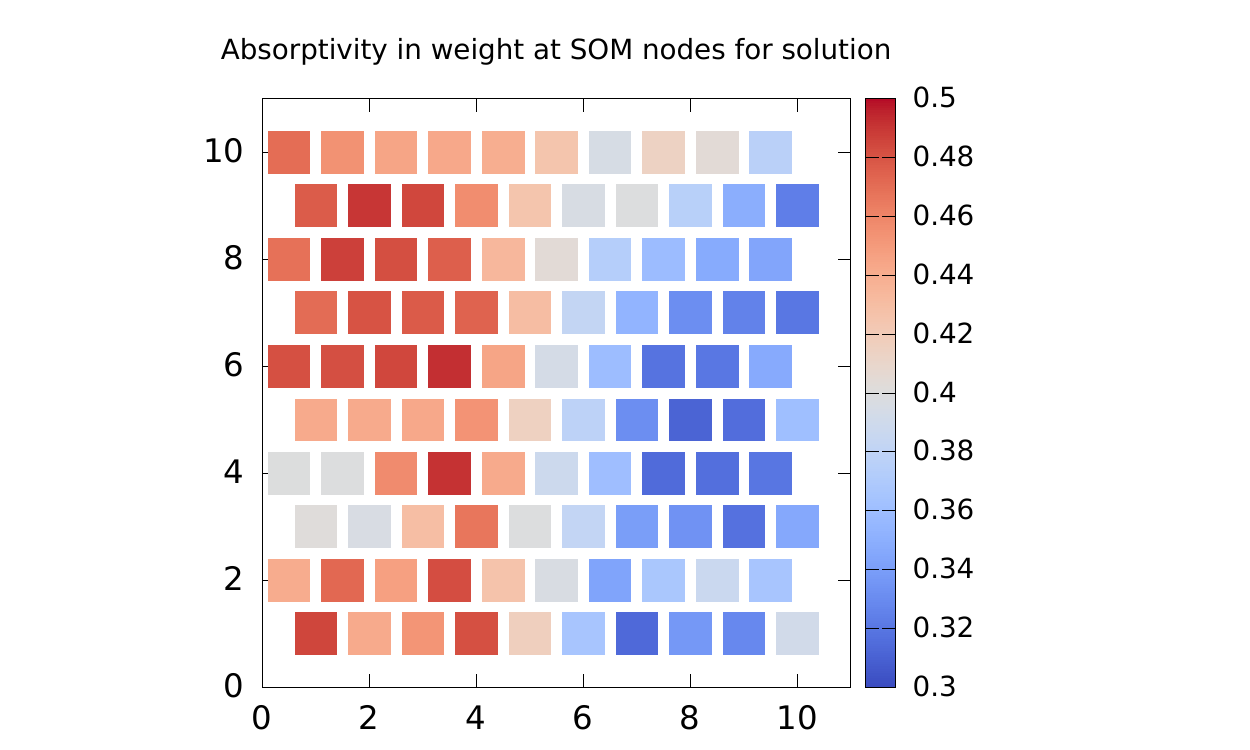} \\
(a) & (b) & (c) & (d) \\
\includegraphics[trim = 1.5cm 0 2cm 0, clip, width=0.23\textwidth]{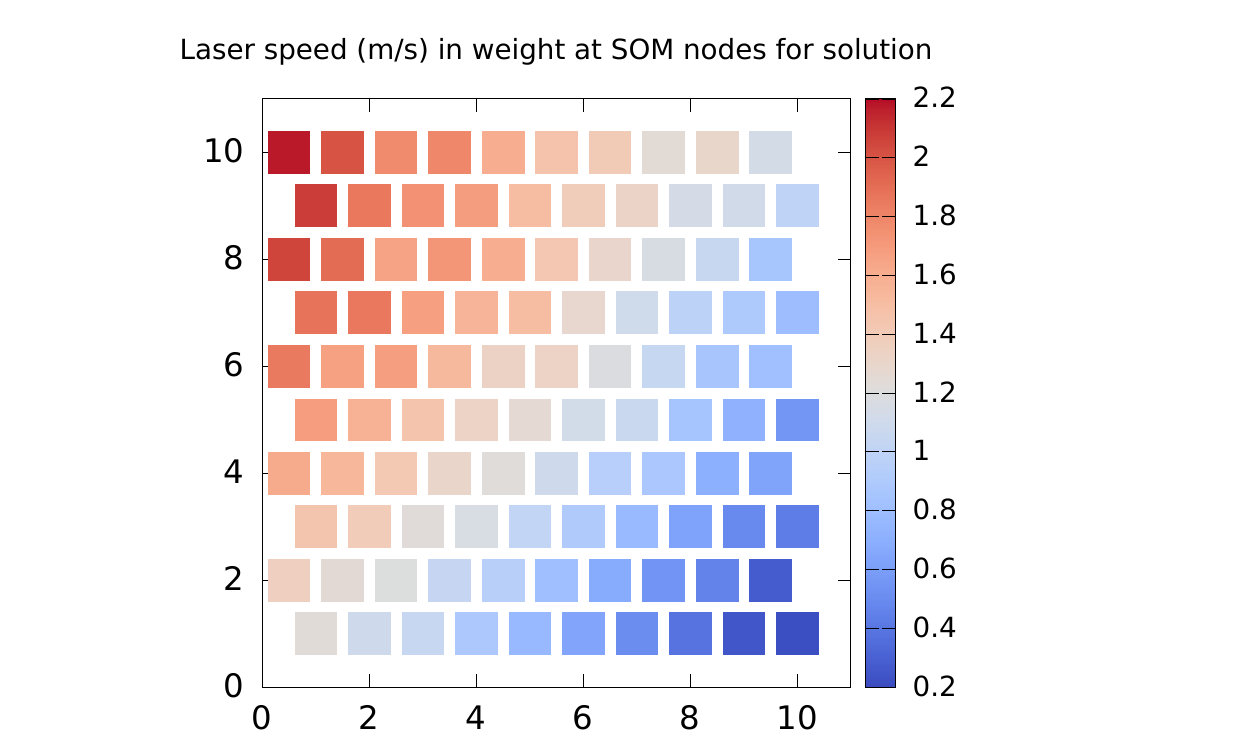} &
\includegraphics[trim = 1.5cm 0 2cm 0, clip, width=0.23\textwidth]{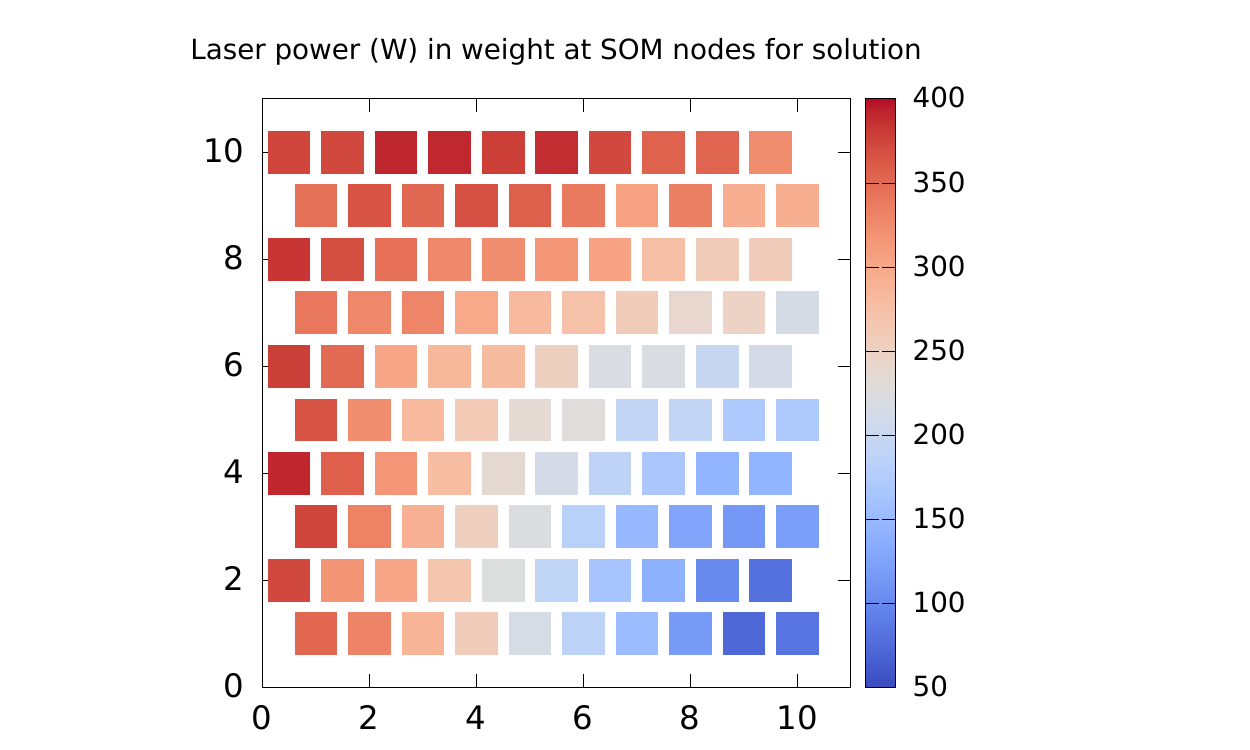} &
\includegraphics[trim = 1.5cm 0 2cm 0, clip, width=0.23\textwidth]{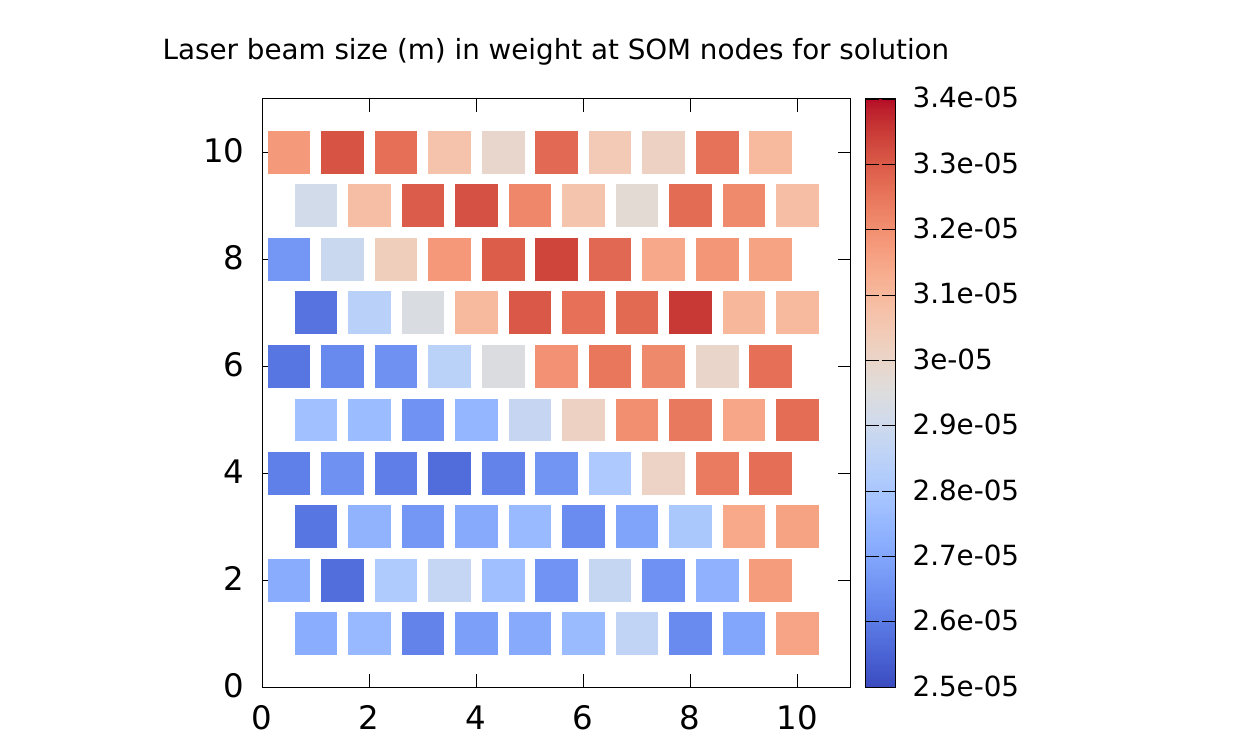} &
\includegraphics[trim = 1.5cm 0 2cm 0, clip, width=0.23\textwidth]{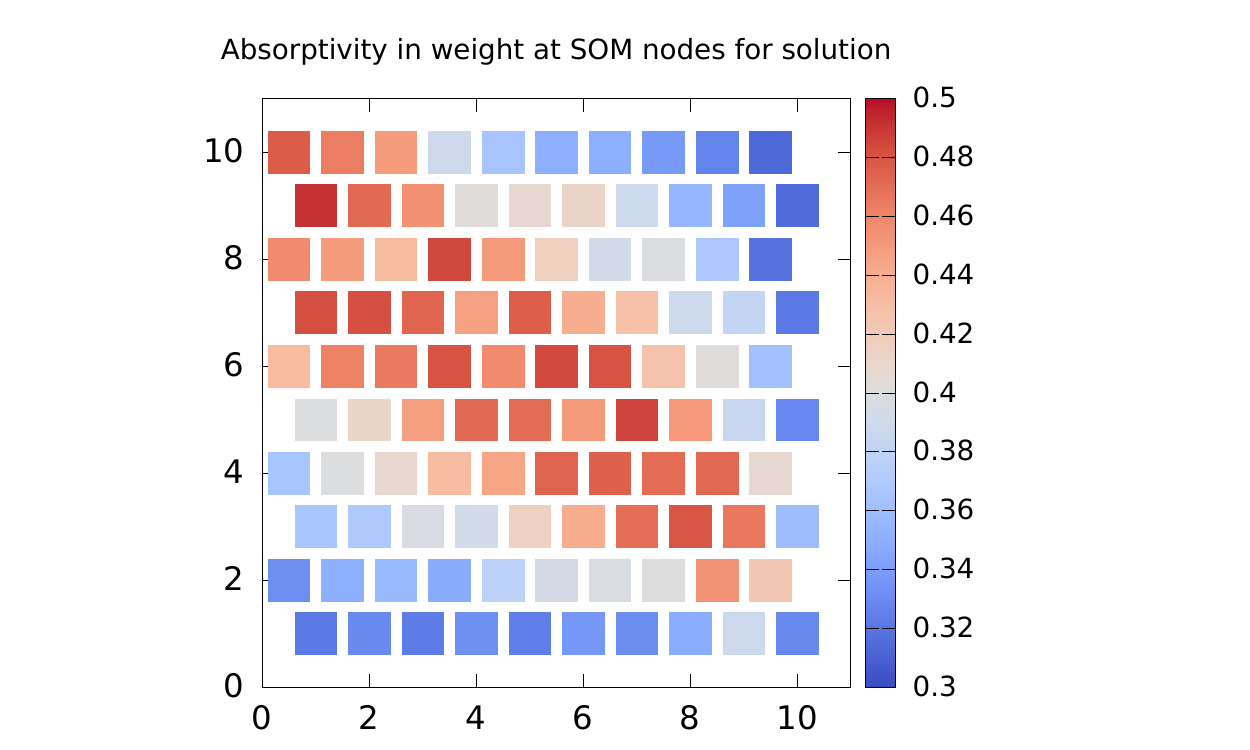} \\
(e) & (f) &(g) & (h) \\
\end{tabular}
\vspace{-0.3cm}
\caption{Solution instances, with melt-pool depth in range [55,65]\micron, after
  processing using a $10\times10$ SOM.  The plots 
  show the weight at each SOM node for the four inputs from left
  to right --- (a) speed, (b) power, (c) beam size, and (d)
  absorptivity --- using unweighted (top row, a-d) and weighted (bottom, e-h) distances.}
\label{fig:som_soln_2}
\end{figure*}

Next, we construct a $10 \times 10$ SOM using just the 700 instances
with depth in the range [55,65]\micron; a smaller number of nodes is
used as the data set is smaller. Figure~\ref{fig:som_soln_1} shows the
mean value of the output at each SOM node; as the range of outputs is
quite small, the plots are unremarkable. However,
Figure~\ref{fig:som_soln_2} that shows the values of the weight
vectors in each dimension using unweighted (top) and weighted (bottom)
distances is a lot more interesting. First, as with the full ET5000
data set, we see that using a weighted distance results in a better
``organized'' SOM as the variation of weights across the nodes,
especially in the speed and power dimensions, is much smoother than
with unweighted distances.  We also note that the solution spans the
full range in all four input dimensions, furthur confirming that the
solution is not a tight cluster.

\begin{figure*}[htb]
\centering
\begin{tabular}{cc}
\includegraphics[width=0.42\textwidth]{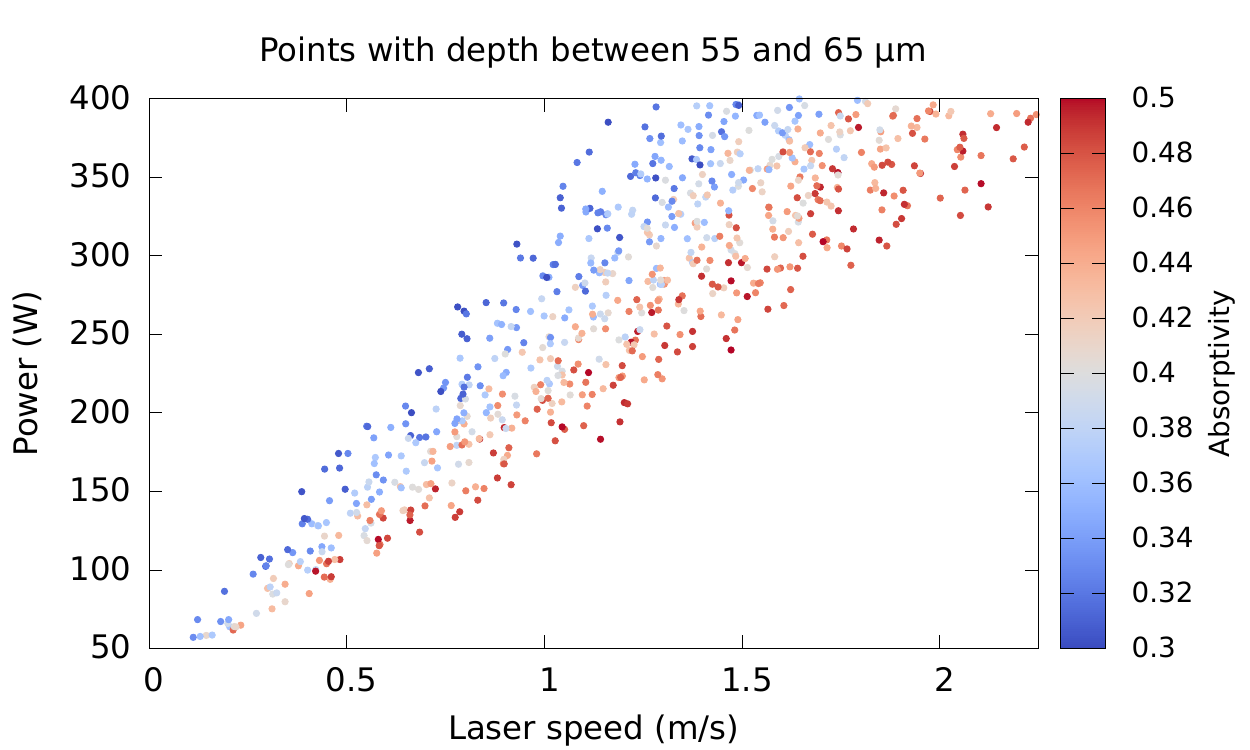} &
\includegraphics[width=0.42\textwidth]{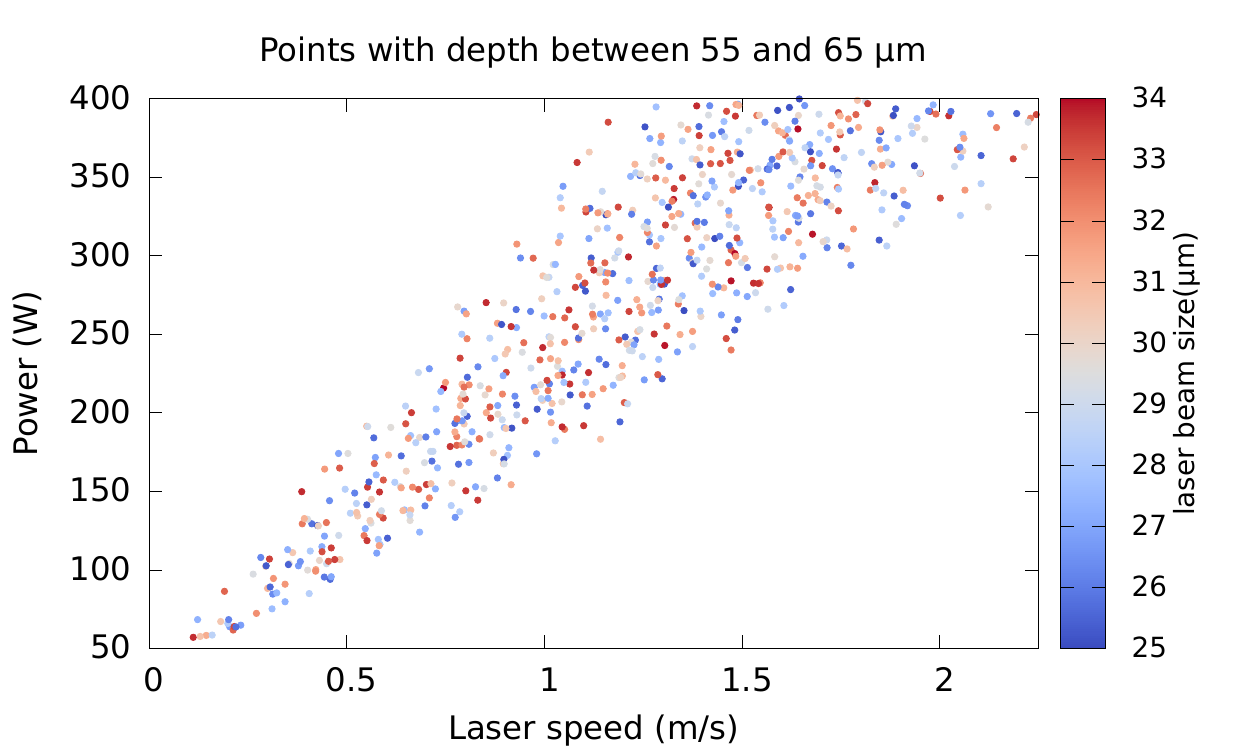} \\
\end{tabular}
\vspace{-0.2cm}
\caption{Solution instances, with melt-pool depth in range
  [55,65]\micron, in the speed and power space, with points colored by
  absorptivity (left) and laser beam size (right). }
\label{fig:som_soln_3}
\end{figure*}

We next focus on the results with weighted distances
(Figure~\ref{fig:som_soln_2}, panels (e)-(h)), to interpret the values
in the corresponding nodes of the SOM across the four dimensions and
understand where the solution lies in the four-dimensional input
space.  The plot for speed (panel (e)) has few nodes with very high or
very low speed, indicating few such points in the solution, while the
plot for power (panel (f)) has more nodes at higher power (in red)
than at lower power (in blue). Looking at corresponding nodes in these
two plots, we notice that instances at high power (along left and top
edges of the SOM in panel (f)) have speed ranges that are medium to
high in panel (e).  However, instances at low power in the lower right
corner of panel (f) have, in the corresponding nodes of panel (e),
speed values that are also low. These observations are also
corroborated by the plot in the left panel of
Figure~\ref{fig:two_space_projection} that indicates that at low
power, the solution exists across a narrow range of speed values, but
at high power, the solution spans a broader range of speed values.
Also, as expected, to keep the melt-pool depth roughly constant, when
the power is high (low), the speed is also high (low).

The SOM results in Figure~\ref{fig:som_soln_2}, bottom row, also allow
us to interpret the values in the corresponding nodes in the other two
dimensions. We note that the node values for the beam size vary less
smoothly than the values for absorptivity, suggesting that beam size
plays a less important role in determining the melt pool depth. The
variation in beam size (high at top right reducing to low values at
bottom left), is orthogonal to the variation in the power and speed,
which is high at top left and low at bottom right. So, at any value of
the power or speed, the solution spans nearly the full range of beam
sizes. This is especially clear for nodes that have high values of
power, along the top and left edges of the SOM.

We also observe that for these nodes with a high value of power, the
nodes with a high value of speed (panel (e), top left) coincide with a high value
of absorptivity. As the speed reduces from left to right and top to
bottom, the absorptivity also reduces. The same behavior also holds at
moderate values of power. This suggests that if we create a plot with
the solution points in the speed and power space, colored by the
absorptivity, then, for a given power value, we should see the
absorptivity being high at high speed and low at low speed. This is
confirmed by Figure~\ref{fig:som_soln_3}, left panel. On hindsight,
this finding is not unexpected --- for near-constant depth at a given
power, if we increase the speed, we should increase absorptivity to
keep the energy input near constant. In our problem, we cannot
explicitly control the absorptivity of the material; we used a range
of values simply to account for the uncertainty in the value. But it
is interesting none-the-less that the SOMs were able to identify the
importance of absorptivity, especially as it is non-trivial, even with
just four inputs, to find such a pattern simply by projecting the
solution on subsets of the input dimensions.  In
Figure~\ref{fig:som_soln_3}, right panel, we include a similar plot
using beam size, which, as expected, shows that beam size does not
play an important role in the solution.

%
\section{Conclusions}
\label{sec:conclusions}
%

In this paper, we considered two issues that make it challenging to
solve inverse problems --- the slow speed of code surrogates, such as
Gaussian processes, and our inability to understand and interpret the
solution in a high dimensional input space. We evaluated our proposed
solutions to these issues using a problem from additive manufacturing.
We showed how the simple idea of independent-block GP can be used to speed
up Gaussian processes substantially, without a loss in prediction accuracy. By
dividing the data into $B$ blocks, we could obtain a speed up of at
least $B^2$, or $B^3$ if we parallelized the algorithm across the
blocks. We also discussed how we should identify the dimensions on
which to block the data and the number of blocks to create. To address
the second issue, we found that using parallel coordinate plots did
not offer much insight into the solution to our inverse problem in the
four-dimensional input space. However, Kohonen maps, especially when
used with weighted distances, were a simple and robust technique that
allowed us to interpret the solution to the inverse problem.

%
\section{Acknowledgment}
\label{sec:ack}
%

LLNL-TR-818722 This work performed under the auspices of the U.S. Department of
Energy by Lawrence Livermore National Laboratory under Contract
DE-AC52-07NA27344. This work is part of the IDEALS project
funded by the ASCR Program (late Dr. Lucille Nowell, Program
Manager) at the Office of Science, US Department of Energy.

The early work on tapering was performed by Juliette Franzman,
undergraduate at UC Berkeley, during a summer internship in 2019. The
early work on Kohonen maps was performed by Ravi Ponmalai,
undergraduate at UC Irvine, during a summer internship in 2019.

\bibliographystyle{acm}
\bibliography{ms_arxiv}

\begin{thebibliography}{10}

\bibitem{anderson99:lapack}
{\sc Anderson, E., Bai, Z., Bischof, C., Blackford, S., Dongarra, J. D.~J.,
  Croz, J.~D., Greenbaum, A., Hammarling, S., McKenney, A., and Sorensen, D.}
\newblock {\em LAPACK Users' Guide}, third~ed.
\newblock SIAM, Philadelphia, Pennsylvania, USA, 1999.

\bibitem{atkeson97:regression}
{\sc Atkeson, C., Schaal, S.~A., and Moore, A.~W.}
\newblock Locally weighted learning.
\newblock {\em AI Review 11\/} (1997), 75--133.

\bibitem{bergstra12:hyperparam}
{\sc Bergstra, J., and Bengio, Y.}
\newblock Random search for hyper-parameter optimization.
\newblock {\em Journal of Machine Learning Research 13\/} (2012), 281--305.

\bibitem{beuth13:processmap}
{\sc Beuth, J., et~al.}
\newblock Process mapping for qualification across multiple direct metal
  additive manufacturing processes.
\newblock In {\em International Solid Freeform Fabrication Symposium, An
  Additive Manufacturing Conference, D. Bourell (Ed.), University of Texas at
  Austin, Austin, Texas\/} (2013), pp.~655--665.

\bibitem{bolin2013:taper}
{\sc Bolin, D., and Lindgren, F.}
\newblock A comparison between markov approximations and other methods for
  large spatial data sets.
\newblock {\em Computational Statistics \& Data Analysis 61\/} (2013), 7 -- 21.

\bibitem{bolin2015:taper}
{\sc Bolin, D., and Wallin, J.}
\newblock Spatially adaptive covariance tapering.
\newblock {\em Spatial Statistics 18\/} (06 2015).

\bibitem{eagar83:temp}
{\sc Eagar, T., and Tsai, N.}
\newblock Temperature-fields produced by traveling distributed heat-sources.
\newblock {\em Welding Journal 62\/} (1983), S346--S355.

\bibitem{ebden08:gp}
{\sc Ebden, M.}
\newblock Gaussian process for regression and classification: A quick
  introduction.
\newblock Available at http://arxiv.org/abs/1505.02965, submitted May 2015,
  2008.

\bibitem{franzman2019:taper}
{\sc Franzman, J., and Kamath, C.}
\newblock {Understanding the Effects of Tapering on Gaussian Process
  Regression}.
\newblock Tech. Rep. LLNL-TR-787826, Lawrence Livermore National Laboratory,
  Livermore, CA, August 2019.
\newblock Available at https://www.osti.gov/servlets/purl/1558874.

\bibitem{furrer2006:taper}
{\sc Furrer, R., Genton, M.~G., and Nychka, D.}
\newblock Covariance tapering for interpolation of large spatial datasets.
\newblock {\em Journal of Computational and Graphical Statistics 15}, 3 (2006),
  502--523.

\bibitem{inselberg09:book}
{\sc Inselberg, A.}
\newblock {\em Parallel Coordinates: Visual Multidimensional Geometry and Its
  Applications}.
\newblock Springer, New York, NY, 2009.

\bibitem{kamath16:statinf}
{\sc Kamath, C.}
\newblock Data mining and statistical inference in selective laser melting.
\newblock {\em International Journal of Advanced Manufacturing Technology 86\/}
  (September 2016), 1659--1677.

\bibitem{kamath14:density}
{\sc Kamath, C., El-dasher, B., Gallegos, G.~F., King, W.~E., and Sisto, A.}
\newblock {Density of additively-manufactured, 316L SS parts using laser
  powder-bed fus ion at powers up to 400 W}.
\newblock {\em International Journal of Advanced Manufacturing Technology 74\/}
  (2014), 65--78.

\bibitem{kamath2018:small}
{\sc Kamath, C., and Fan, Y.-J.}
\newblock Regression with small data sets: a case study using code surrogates
  in additive manufacturing.
\newblock {\em Knowledge and Information Systems 57\/} (2018), 475--493.

\bibitem{king2014:keyhole}
{\sc King, W.~E., et~al.}
\newblock Observation of keyhole-mode laser melting in laser powder-bed fusion
  additive manufacturing.
\newblock {\em Journal of Materials Processing Technology 214}, 12 (2014), 2915
  -- 2925.

\bibitem{kohonen93:things}
{\sc {Kohonen}, T.}
\newblock Things you haven't heard about the self-organizing map.
\newblock In {\em IEEE International Conference on Neural Networks\/} (1993),
  vol.~3, pp.~1147--1156 vol.3.

\bibitem{kohonen00:book}
{\sc Kohonen, T.}
\newblock {\em Self-Organizing Maps}, 3rd~ed.
\newblock Springer-Verlag, Berlin, Heidelberg, 2000.

\bibitem{kohonen2013:essentials}
{\sc Kohonen, T.}
\newblock Essentials of the self-organizing map.
\newblock {\em Neural Networks 37\/} (2013), 52 -- 65.
\newblock Twenty-fifth Anniversay Commemorative Issue.

\bibitem{li2018:hyperband}
{\sc Li, L., Jamieson, K., DeSalvo, G., Rostamizadeh, A., and Talwalkar, A.}
\newblock Hyperband: A novel bandit-based approach to hyperparameter
  optimization.
\newblock {\em Journal of Machine Learning Research 18}, 185 (2018), 1--52.

\bibitem{mitchell91:sampling}
{\sc Mitchell, D.~P.}
\newblock Spectrally optimal sampling for distribution ray tracing.
\newblock {\em Computer Graphics 25}, 4 (1991), 157--164.

\bibitem{ponmalai2019:som}
{\sc Ponmalai, R., and Kamath, C.}
\newblock {Self-Organizing Maps and Their Applications to Data Analysis}.
\newblock Tech. Rep. LLNL-TR-791165, Lawrence Livermore National Laboratory,
  Livermore, CA, September 2019.
\newblock Available at https://www.osti.gov/servlets/purl/1566795.

\bibitem{rasmussen06:book}
{\sc Rasmussen, C.~E., and Williams, C. K.~I.}
\newblock {\em Gaussian Processes for Machine Learning}.
\newblock MIT Press, Cambridge, MA, 2006.

\bibitem{saad03:book}
{\sc Saad, Y.}
\newblock {\em Iterative Methods for Sparse Linear Systems}, 2nd~ed.
\newblock Society for Industrial and Applied Mathematics, USA, 2003.

\bibitem{stein13:taper}
{\sc Stein, M.~L.}
\newblock Statistical properties of covariance tapers.
\newblock {\em Journal of Computational and Graphical Statistics 22}, 4 (2013),
  866--885.

\end{thebibliography}

\end{document}